\documentclass[format=acmsmall,review=false,screen=true, nonacm =True, standardpagestyle=True]{acmart}
\usepackage{longtable}
\usepackage{adjustbox,lipsum}

\ifxetex
\DeclareTextFontCommand{\texturdu}{\beginR\urdufont\aftergroup\endR}
\fi
\ifluatex
\DeclareTextFontCommand{\texturdu}{\textdir TRT\urdufont}
\fi
\usepackage{graphicx,subfigure}
\usepackage{hyperref,xspace}
\usepackage[autostyle]{csquotes}

\begin{document}
	\title{Neural Machine Translation for Low-Resource Languages: A Survey}
	
	\author{Surangika Ranathunga}
	\email{surangika@cse.mrt.ac.lk}
	\orcid{https://orcid.org/0000-0003-0701-0204 }
	\affiliation{%
		\institution{University of Moratuwa}
		\streetaddress{Bandaranayaka Mawatha}
		\city{Katubedda}
		\country{Sri Lanka}
		\postcode{10400}
	}
	\author{En-Shiun Annie Lee}
	\email{annie.lee@cs.toronto.edu}
	\affiliation{%
		\institution{University of Toronto}
		\streetaddress{9th Floor - 700 University Avenue}
		\city{Toronto}
		\country{Canada}
		\postcode{M5G 1Z5}
	}
	
	\author{Marjana Prifti Skenduli}
		\email{marjanaprifti@unyt.edu.al}
	\orcid{https://orcid.org/0000-0002-2707-1621}
	\affiliation{%
		\institution{University of New York Tirana}
		\city{Tirana}
		\country{Albania}
		\postcode{1000}
		}
	\author{Ravi Shekhar}
		\email{r.shekgar@qmul.ac.uk}
	\orcid{}
	\affiliation{%
		\institution{Queen Mary University}
		\country{London}
		\postcode{}
		}
		\author{Mehreen Alam}
	\email{}
	\orcid{}
	\affiliation{%
		\institution{National University of Computer and Emerging Sciences}
		\country{Pakistan}
		\postcode{}
		}
	
	\author{Rishemjit Kaur}
		\email{rishemjit.kaur@csio.res.in}
		\affiliation{%
		\institution{CSIR-Central Scientific Instruments Organisation}
		\streetaddress{Sector-30C}
		\city{Chandigarh}
		\country{India}
		\postcode{160030}
	}
	\orcid{}	

	\begin{abstract}

Neural Machine Translation (NMT) has seen a tremendous
spurt of growth in less than ten years, and has already entered a mature phase. While considered as the most widely used solution for Machine Translation, its performance on low-resource language pairs still remains sub-optimal compared to the high-resource counterparts, due to the unavailability of large parallel corpora. Therefore, the implementation of NMT techniques for low-resource language pairs has been receiving the spotlight in the recent NMT research arena, thus leading to a substantial amount of research reported on this topic. This paper presents a detailed survey of research advancements in low-resource language NMT (LRL-NMT), along with a quantitative analysis aimed at identifying the most popular solutions. Based on our findings from reviewing previous work, this survey paper provides a set of guidelines to select the possible NMT technique for a given LRL data setting. It also presents a holistic view of the LRL-NMT research landscape and provides a list of recommendations to further enhance the research efforts on LRL-NMT.  
	\end{abstract}

	
	
	\maketitle

\section{Introduction}

Since the beginning of time, language and communication has been central to human interactions. Therefore, translating between different languages has been pivotal in societal and cultural advancements.  

Machine Translation (MT) was one of the first applications conceived to be solvable by computers; this vision was birthed by the ``translation memorandum'' presented by Warren Weaver, and the word-for-word translation system by IBM in 1954~\cite{hutchins1997first}.  

Consequently, different techniques were developed to address the problem of Machine Translation, with a prominent being Statistical Machine Translation (SMT).  
Because the performance of the SMT system is directly impacted by the number of parallel sentence pairs available for training, a heavy emphasis has been placed on creating parallel datasets (also known as bitext) in addition to research on new MT techniques.\\
In 2013, the introduction of end-to-end neural encoder-decoder based MT systems saw a breakthrough with promising results, which soon got popularized as Neural Machine Translation (NMT). Currently NMT is the dominant technique in the community. 
However, it was quickly realized that these initial NMT systems required huge volumes of parallel data to achieve comparable results to that of SMT~\cite{koehn2017six}.

High resource language pairs (such as English and French) do not have dataset size concerns because researchers have created ample amounts of parallel corpora over the years\footnote{The English-French corpus~\citet{cho-etal-2014-properties} used contained 348 Million parallel sentences.}. 
However, the requirement of having large amounts of parallel data is not a realistic assumption for many of the 7000+ languages currently in use around the world and therefore is considered a major challenge for low-resource languages (LRLs)~\cite{koehn2017six}. 
Due to economic and social reasons, it is useful to automatically translate between most of these LRLs, particularly for countries that have multiple official languages. Therefore, in recent years, there has been a noticeable increase in NMT research (both by academia and industry) that specifically focused on LRL pairs.

Despite this emphasis, we are not aware of any literature review that systematically examines the NMT techniques tailored for LRL pairs.  Although there exists some work that discusses the challenges of using NMT in the context of LRL pairs~\cite{zhang2020neurala} and the application of specific techniques for LRL pairs~\cite{dabre2020survey}, 
 none of them gives a comprehensive view of the available NMT techniques for LRL pairs. This makes it difficult for new researchers in the field to identify the best NMT technique for a given dataset specification. In addition, none of these surveys presents a holistic view of the NMT landscape for LRL pairs to derive insights on research efforts and current practices. 

This survey aims to address the above shortcomings in the NMT research landscape for LRLs. More specifically, it provides researchers working on LRLs a catalogue of methods and approaches for NMT and identifies factors that positively influence NMT research on LRL pairs. To achieve these aims, we answer the following research questions: 
\begin{enumerate}
    \item \textbf{NMT Techniques:}  What are the major NMT techniques that can be applied to LRL pairs, and what are the current trends?
    \item \textbf{Technique Selection:}  How to select the most suitable NMT technique for a given language?
    \item \textbf{Future Directions:} How to increase research efforts and what are the future directives for NMT on LRL pairs?
\end{enumerate}
To answer the above questions, we first conducted a systematic analysis of the NMT techniques that have been applied for LRL pairs, and their progress (Section~\ref{section:label-NMT-Techniques-for-Low-Resource-Languages}). 
Secondly, we critically analysed the applicability of these techniques for LRL pairs in practical terms. Based on our observations, we provide a set of guidelines for those who want to use NMT for LRL pairs to select the most suitable NMT technique by considering the size and type of the datasets, as well as the available computing resources (Section~\ref{technique_selection}). Lastly, we conducted a comprehensive analysis of the amount of NMT research conducted for LRLs in the world (Section~\ref{section:trend_analysis}).  Here, we note a strong correlation between the amount of NMT research per language and the amount of publicly available parallel corpora for that language. We also note the recent rise of regional level research communities that contributed to parallel dataset creation, and thus NMT for LRL pairs in turn.



Therefore, our recommendations to advance the area of NMT for LRL pairs are to 1) create LRL resources (datasets and tools), 2) make computational resources and trained models publicly available, and 3) involve research communities  at a regional-level.  Based on our analysis of the existing NMT techniques, we also recommend possible improvements to existing NMT techniques that would elevate the development of NMT techniques that work well LRL pairs.

\section{Background}
\label{defs_scope}
\subsection{Low-Resource Languages (LRLs)}
\label{LR-Definition}
For Natural Language Processing (NLP), a low-resource problem can arise mainly due to the considered languages being low-resourced, or the considered domains being low-resourced~\cite{hedderich2020survey}. In this paper, we focus on LRLs only. \\
Researchers have attempted to define LRLs by exploring various criteria such as the number of mother-tongue speakers and the number of available datasets. According to \citet{besacier2014automatic}, an LRL\footnote{An LRL is also known as under resourced, low-density, resource-poor, low data, or less-resourced language \cite{besacier2014automatic}} is a language that lacks a unique writing system, lacks (or has) a limited presence on the World Wide Web, lacks linguistic expertise specific to that language, and/or lacks electronic resources such as corpora (monolingual and parallel), vocabulary lists, etc. NLP researchers have used the availability of data (in either labelled, unlabelled or auxiliary data), and the NLP tools and resources as the criteria to define LRLs~\cite{hedderich2020survey}.

Over the years, there have been many initiatives to categorise languages according to the aforementioned different criteria~\cite{harmon1995status,joshi2020state}. Given that the category of a language may change with time, we rely on the language categorization  recently proposed by \citet{joshi2020state} to identify LRLs. As shown in Table~\ref{tab:language_categories},~\citet{joshi2020state} categorised 2485 languages into six groups based on the amount of publicly available data.

\begin{table}[htp]
\centering
{\small %
    \begin{tabular}{p{.04\textwidth}p{.7\textwidth}p{.2\textwidth}}

    \hline
    Class & Description & Language Examples \\ \hline \hline
    0 & Have exceptionally limited resources, and have rarely been considered in language technologies. & Slovene, Sinhala  \\ \hline
    1 & Have some unlabelled data; however, collecting labelled data is challenging. & Nepali, Telugu\\ \hline
    2 &  A small set of labeled datasets has been collected, and language support communities are there to support the language. & Zulu, Irish  \\ \hline 
    3 & Has a strong web presence, and a cultural community that backs it. Have been highly benefited by unsupervised pre-training. & Afrikaans, Urdu \\ \hline
    4 & Have a large amount of unlabeled data, and lesser, but still a significant amount of labelled data. have  dedicated NLP communities researching these languages. & Russian, Hindi \\ \hline
    5 &  Have a dominant online presence. There have been massive investments in the development of resources and technologies. & English, Japanese\\ \hline
   \end{tabular}
   }
  \caption{Language Categories identified by~\citet{joshi2020state}}~\label{tab:language_categories}
\end{table}
Unlike other NLP tasks, MT take place between two languages. Thus, in MT the resourcefulness of a language pair is determined by the available amount of parallel corpora between the considered languages. The terms `high-resource', `low-resource', as well as `extremely low-resource' have been commonly used  when referring to the parallel corpora at hand. However, there is no  minimum requirement in the size of the parallel corpora to categorise a language pair as high, low, or extremely low-resource. Some early research considered even $1$ million parallel sentences as LR~\cite{zoph2016transfer}. More recent research seems to consider a language pair as LR or extremely LR if the available parallel corpora for the considered pair for NMT experiments is below $0.5$ Million, and below $0.1$ Million, respectively~\cite{liu2020multilingual,zaremoodi-etal-2018-adaptive,Platanios2018ContextualPG,lakew2019adapting,qi-etal-2018-pre}; however, these are not absolute values for the size of the corpora.\\
Even if a particular language has a large number of monolingual corpora while still having a small parallel corpus  with another language, this language pair is  considered as LR for the NMT task. We assume that languages that have been labelled as LR by~\citet{joshi2020state} have very small parallel corpora with other languages, or have no parallel corpora at all.

\subsection{Related Work} \label{label-Related-Work}
Some of the previous survey papers discussed different
NMT architectures~\cite{popescu2019context,stahlberg2020neural,yang2020survey,zhang2020neurala, vazquez2020systematic}. They did not contain any reference of LRL-NMT, except for~\citet{zhang2020neurala}, which briefly identified Multilingual NMT (multi-NMT), unsupervised, and semi-supervised LRL-NMT techniques.
Another set of papers surveyed only one possible NMT methodology, for example multi-NMT~\cite{dabre2020survey}, leveraging monolingual data for
NMT~\cite{gibadullin2019survey}, use of pre-trained embeddings for NMT~\cite{qi-etal-2018-pre}, or domain adaptation techniques for NMT ~\cite{chu-wang-2018-survey}. Out of these surveys,~\citet{gibadullin2019survey} specifically focused on LR settings, may be because monolingual data is more useful in that scenario.
We also observed that some surveys focused on the broader MT, both SMT and NMT, in problem domains such as document-level MT~\cite{maruf2019survey}, while others focused on MT for a selected set of languages~\cite{alsohybe2017machine}. On a different front, we found surveys that discussed LRL scenarios in the general context of NLP, but did not have a noticeable focus on NMT or even MT~\cite{hedderich2020survey}.
   
Table~\ref{tab:survey_papers} categorises the survey papers discussed above. In conclusion, although there are surveys that dedicated a brief section on LRL-NMT and others that explicitly focus on LRLs for a selected NMT technique, there is no comprehensive survey on leveraging NMT for LRLs.\\

  \begin{table} [htp]
\centering
{\small %
    \begin{tabular}{p{.3\textwidth}p{.7\textwidth}}

    \hline
    Type of survey & Examples\\ \hline \hline
    NMT Architectures & \citet{zhang2020neurala}, \citet{yang2020survey}, \citet{stahlberg2020neural}, \citet{popescu2019context}, ~\citet{vazquez2020systematic}\\ \hline
    Specific NMT Methodologies & \citet{dabre2020survey}, \citet{chu-wang-2018-survey}, \citet{gibadullin2019survey}, \citet{qi-etal-2018-pre}\\ \hline
    Specific MT Problem Domain & \citet{maruf2019survey} \\ \hline
    Specific Language & \citet{alsohybe2017machine}\\ \hline
    LRL NLP & \citet{hedderich2020survey}\\ \hline
   \end{tabular}}
  \caption{Type of Survey papers}~\label{tab:survey_papers}
\end{table}

\subsection{Scope of the Survey}
Most of the NMT techniques discussed in this paper can be used in the context of LRL translation as well as LR domain translation. However, an LR domain, such as medical or finance, can exist for a high-resource language, such as English as well~\cite{li2019metamta, hedderich2020survey}. In that case, additional language resources (e.g. WordNet, Named Entity Recognisers) can be utilised in developing the solution. However, such resources might not be available for LRL pairs. Thus, solutions that only apply for LR domains are considered out of scope for this paper. In this paper, we use the phrase low-resource language NMT (LRL-NMT) to refer to NMT techniques that are applicable for translation between LRL pairs.

Similarly, we omit techniques that focused on NMT in general, without any specific focus on the translation of LRL pairs. We also omit techniques that focus on speech translation only, and multimodal translation (which is typically between images and text), as such research is not common in the context of LRL pairs. Some techniques (e.g. data augmentation (Section~\ref{sec:dataAug}) and pivoting(Section~\ref{sec:zero_shot})) have been used in the context of SMT as well, which is not discussed in this review.

NMT solutions for zero-shot translation  (no parallel data to train an MT model) are included because of its relationship to the task of transation of LRL pairs, with an overlap between the techniques used.
    
    
\section{NMT Techniques for Low-Resource Languages}
\label{section:label-NMT-Techniques-for-Low-Resource-Languages}
\subsection{Overview}
NMT methodologies fall broadly into supervised, semi-supervised, and unsupervised.  Supervised NMT is the default architecture that relies on large-scale parallel datasets. Recurrent neural architecture with attention~\cite{38ed090f8de94fb3b0b46b86f9133623}, as well as the recently introduced transformer architecture~\cite{vaswani2017attention}, are commonly used. However, due to space limitations, we do not detail out these techniques, and interested readers can refer to the aforementioned references.
Both these neural architectures rely on large parallel corpora, an advantage not available to LRLs. A solution is to synthetically generate data, which is called \textbf{data augmentation} (Section~\ref{sec:dataAug}). These techniques can be applied irrespective of the NMT architecture used.    
In the extreme case where no parallel data is available, \textbf{unsupervised} NMT techniques (Section~\ref{unsupervised_nmt}) can be employed. Even if the available parallel corpora is small, it is possible to combine them with the monolingual data of the languages concerned, in a \textbf{semi-supervised} manner (Section~\ref{semi-supervised}). 

Even if parallel data is available, building (bilingual) NMT models between each pair of languages is not practical. As a solution, \textbf{multi-NMT} models (Section~\ref{sec:multiNMT}) were introduced, which  facilitate the translation between more than one language pair using a single model. Most of the multi-NMT models are based on supervised NMT, while some research is available on the applicability of semi-supervised, and unsupervised NMT in a multilingual setting. Although multi-NMT models were initially introduced to avoid the need to build individual bilingual translation models, their capability in the translation of LRL pairs is shown to be promising. 

\textbf{Transfer learning} (Section~\ref{transfer_learning_nmt}) is a technique that is commonly used in low-resource NLP, including NMT. Here, an NMT model trained on a high-resource language pair is used to initialize a child model, which reduces the amount of time taken to train the latter, while guaranteeing better performances over training the child model from scratch. In particular, transfer learning on multi-NMT models have shown very good performance for LRL pairs. This is a very promising development, as it is time-consuming to train a multi-NMT model every time a dataset for a new language pair comes up.

\textbf{Zero-shot} NMT (Section~\ref{sec:zero_shot}) is a problem related to LRL-NMT. In zero-shot, there is no parallel data, and the model is trained with no parallel data for the considered language pair. While some researchers~\cite{blackwood-etal-2018-multilingual, murthy2018addressing} consider zero-shot to be synonymous to extremely LR case, others ~\cite{johnson2017google, kim2019pivot} disagree.   Promising solutions for zero-shot translation that have been presented include pivoting, multi-NMT, unsupervised NMT, and transfer learning. Zero-shot translation is extremely useful because it eliminates the requirement of the existence of parallel data between every pair of languages. 

\begin{figure}
    \centering
    \includegraphics[scale = 0.45]{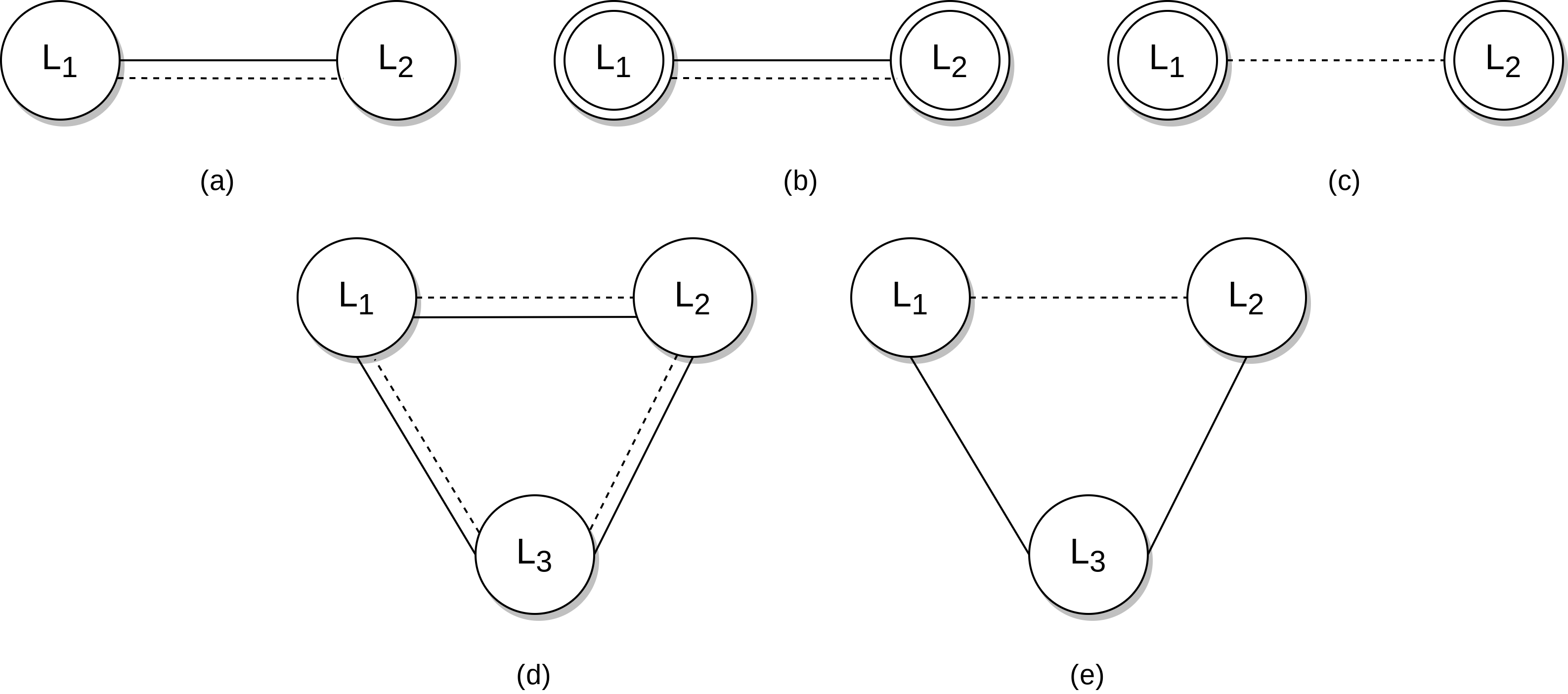}
    \caption{NMT techniques applicable for the translation of LRL pairs. $L_1 - L_3$ refer to languages.  Dashed lines indicate translation task, and solid lines indicate the availability of parallel corpora. Double circles in (a) and (b) indicate the languages have monolingual data. (a) Bilingual supervised NMT, (b) Bilingual semi-supervised NMT, (c) Bilingual unsupervised NMT, (d) Multi-NMT, (e) Pivoting. }
    \label{fig:NMT_techniques}
\end{figure}

Figure~\ref{fig:NMT_techniques} gives an overview of these techniques. Note that it does not cover all the possible scenarios. For example, semi-supervised NMT techniques can work with monolingual data available either at the source or target side, and multi-NMT works with more than three languages.\\
The following sub-sections discuss the aforementioned techniques at length. At the end of each sub-section, we discuss how the technique has been employed with respect to LRLs. 
\subsection{Data Augmentation Techniques}
\label{sec:dataAug}
Data augmentation (DA) is a set of techniques that is used to create additional data either by modifying existing data or adding data from different sources, to be used in training Machine Learning models. For the problem of MT, data augmentation is used to generate \emph{synthetic} parallel sentence pairs to train data-centric MT models, such as SMT and NMT. In contrast to the other techniques discussed in this section, data augmentation techniques usually do not alter the NMT architecture but generate data to train these neural architectures. Data augmentation techniques for NMT could be divided into 3 categories: i) word or phrase replacement based augmentation, ii) back-translation based augmentation, and iii) parallel corpus mining.
\\

\textbf{1. Word or phrase replacement based augmentation:} In this technique, a subset of sentences from an existing parallel or monolingual corpus is selected, and new synthetic sentences are generated by replacing words or phrases in that selected set of sentences. One solution is to use a bilingual dictionary and replace all the words~\cite{nag2020incorporating} or rare words~\cite{peng2020dictionary} in the selected  sentences of a monolingual corpus, with the words in the other language in order to generate its translation. Another solution is to replace frequent words in the target sentences with rare words in the target vocabulary and then modifying the aligned source words accordingly~\cite{fadaee2017data}. The main problem with such synthetic data is a lack of fluency. There have been subsequent attempts to select the best set of words considering fluency ~\cite{wang-etal-2018-switchout, gao-etal-2019-soft}. Alternatively, instead of replacing words, phrases can be replaced, which preserves the context and in turn, improves the fluency of resulting sentences~\cite{liu2021counterfactual}. To further improve fluency, syntactic rules (e.g.~morphological, POS, or dependency rules) have been imposed during word replacement~\cite{tennage2017neural, duan2020syntax}.
   
\textbf{2. Back-Translation based Data Augmentation: } Back-Translation is the process of translating a monolingual corpus in the target language by pre-existing MT system, in the reverse translation direction, into the source language. Then the obtained synthetic source language sentences along with their respective target language sentences are used to construct a synthetic parallel corpus~\cite{sennrich-etal-2016-improving}. Usually, target-side sentences are selected to be back-translated, because monolingual target data helps improve the fluency of the translation model.~\citet{fadaee-monz-2018-back}  empirically showed that starting with the source side has a lesser success. \\
   Synthetic data generated using BT tends to be noisier than the original parallel data, especially if the MT system used to generate the synthetic data is suboptimal. This is particularly the case with MT systems trained with very small amounts of data. Thus, subsequent research improved BT using  data selection, data filtering, distinguishing between synthetic and original data, sampling and iterative BT.  These improvements are further discussed below.   \\
     \textbf{Iterative back-translation}: In one form of iterative back-translation, source and target monolingual data are back-translated using source to target and target to source NMT models, respectively. This procedure is continued iteratively, where the same set of sentences is back-translated several times until no improvement is observed in both translation direction~\cite{hoang2018iterative, artetxe2020all}. Another option is to improve the forward and backward translators in an iterative manner~\cite{cotterell2018explaining, he2016dual}. However, in these systems, the two translators are trained independently. As a solution,~\citet{Zheng2020Mirror-Generative} jointly trained the two translators. 
   \\
   \textbf{Monolingual data selection}: In BT, simply back-translating all the available monolingual data would not guarantee optimal results. One factor that determines the performance of back-translation is the original-synthetic data ratio~\cite{edunov-etal-2020-evaluation}. Thus, the synthetic to original parallel data ratio has to be selected carefully. The purpose of data selection is to select the best subset from the available monolingual corpora to be back-translated~\cite{fadaee-monz-2018-back, dou2020dynamic}.
   \\
    \textbf{Synthetic parallel data filtering}: Even if a subset of monolingual data is selected to be back-translated, the resulting synthetic data could contain noise. Data filtering refers to the process of selecting a subset of the generated synthetic parallel sentences (the highest quality ones) to be used alongside the original data to train the NMT system~\cite{imankulova2017improving, imankulova2019filtered}. 
    \\
   \textbf{Distinguishing between original and back-translated data}: Even after monolingual and parallel data filtering discussed above, it is highly likely that this data would be of lesser quality, compared to the original parallel data. It has been shown that adding a tag to the back-translated data to distinguish them from original data gives better results~\cite{caswell2019tagged, marie2020tagged, khatri2020filtering}. An alternative is to distinguish these two types of data by assigning a weight according to the quality of sentences\cite{dou2020dynamic, wang2019improving, khatri2020filtering}.
  \\
 \textbf{Sampling}: In sampling, multiple source sentences are generated per target sentence in an attempt to average out errors in synthetic sentences~\cite{imamura2018enhancement, gracca2019generalizing}. Note that this technique as well as the previously discussed two techniques can be applied with other forms of DA techniques as well. However, we find experiments reported only in the context of BT.  \\

   
{\textbf{3. Parallel Data Mining (bitext mining) from comparable corpora:}}
Comparable corpora refer to text on the same topic that is not direct translations of each other but may contain fragments that are translation equivalents (e.g.~Wikipedia or news articles reporting the same facts in different languages).  Parallel sentences extracted from comparable corpora have been long identified as a good source of synthetic data for MT. 

Because recently introduced multilingual sentence embeddings have become the common technique for generating parallel data to train NMT models, we only discuss those techniques\footnote{We only discuss multilingual sentence embedding techniques that have been evaluated on NMT tasks. Some techniques have been evaluated on some other NLP tasks s.a. Natural Language Inference.}. In these techniques, a multilingual sentence embedding representation is first learnt between two or more languages. Then, during the sentence ranking step, for each given sentence in one language, a set of nearest neighbours is identified as parallel sentences from the other language, using a sentence similarity measurement technique.

\textbf{Multilingual Embedding generation: }Early research used NMT inspired encoder-decoder architectures to generate multilingual sentence embeddings. These include bilingual dual encoder architectures~\cite{guo2018effective, yang2019improving}, and shared multilingual encoder-decoder architectures~\cite{schwenk2018filtering}.~\citet{artetxe-schwenk-2019-margin} leveraged the shared encoder-decoder model across 93 languages, which is publicly released under the LASER toolkit. This toolkit was the base for subsequent massive-scale parallel corpus extraction projects~ \cite{schwenk2019wikimatrix, banon2020paracrawl,schwenk2019ccmatrix}.

The above-discussed multilingual embedding generation techniques require large parallel corpora during the training process. As a result, unsupervised~\cite{lai2020unsupervised}, as well as self-learning~\cite{ruiter2020self} NMT architectures have been used to generate multilingual embeddings.  
A notable development is the use of pre-trained multilingual embedding models such as multilingual BERT (mBERT) or XLM-R~\cite{zhang2020parallel, keung2020unsupervised, acarcicek2020filtering, sun2021parallel}.  \\
\textbf{Sentence Ranking: }The choice of the sentence similarity measurement technique has been largely unsupervised (cosine similarity was the simplest one employed). However, this simple method is suboptimal, and improved cosine similarity measurements are available~\cite{guo2018effective, hangya2019unsupervised, artetxe-schwenk-2019-margin}. In addition, supervised sentence measurement techniques have been employed to a lesser extent~\cite{acarcicek2020filtering, sun2021parallel}. 

  
\subsubsection{Data Augmentation for Low-Resource Language NMT}
The above three data augmentation techniques have shown promising results for translation between LRL pairs. However, each technique has its practical limitations when applied to the translation of LRL pairs.  BT assumes that an MT system is already available between the given language pair. Moreover, as evidenced by many empirical studies, the success of BT depends on many factors such as the original-synthetic parallel data ratio, and the domain relatedness of the parallel and monolingual data~\cite{fadaee-monz-2018-back, edunov-etal-2018-understanding, xu2019analysis}. In addition, there have been attempts to leverage high-resource language data for LRL with BT; however, its success depends on the relatedness of the high-resource language and LRL~\cite{karakanta2018neural} or the availability of bilingual lexicons~\cite{xia-etal-2019-generalized}.  Word or phrase replacement based augmentation techniques rely on language-specific resources (e.g.~bilingual dictionaries, Part of Speech (POS) taggers, dependency parsers) that many LRLs would not have. One possibility to explore the use of neural language models trained on monolingual data (e.g. BERT and its variants) to increase the fluency of the synthetic data.  For parallel data mining, the applicability of pre-trained multilingual models such as LASER or mBERT is restricted to the languages already included in the pre-trained model. Thus, it is worthwhile to investigate the heuristic and statistical-based parallel corpus mining techniques employed in early SMT research, in the context of LRLs.
\subsection{Unsupervised NMT}
\label{unsupervised_nmt} 
The generation of parallel corpora for language translation is expensive and resource-intensive. In contrast, monolingual corpora are often easier to obtain. As a result, unsupervised NMT using monolingual corpora or cross-lingual word embeddings (i.e., cross-lingual representations of words in a joint embedding space) is  less data-intensive for LRL-NMT \footnote{The input corpora used for training data for unsupervised NMT is assumed to be monolingual corpora, while testing uses parallel corpora to evaluate the true translation}. 
Generally, the architecture for unsupervised NMT makes use of Generative Adversarial Networks (GANs) and contains the following three steps~\citep{yang2017improving}: i) initialization, ii) back-translation, and iii) discriminative classifier. 

\begin{figure}
    \centering
    \includegraphics[scale = 0.45]{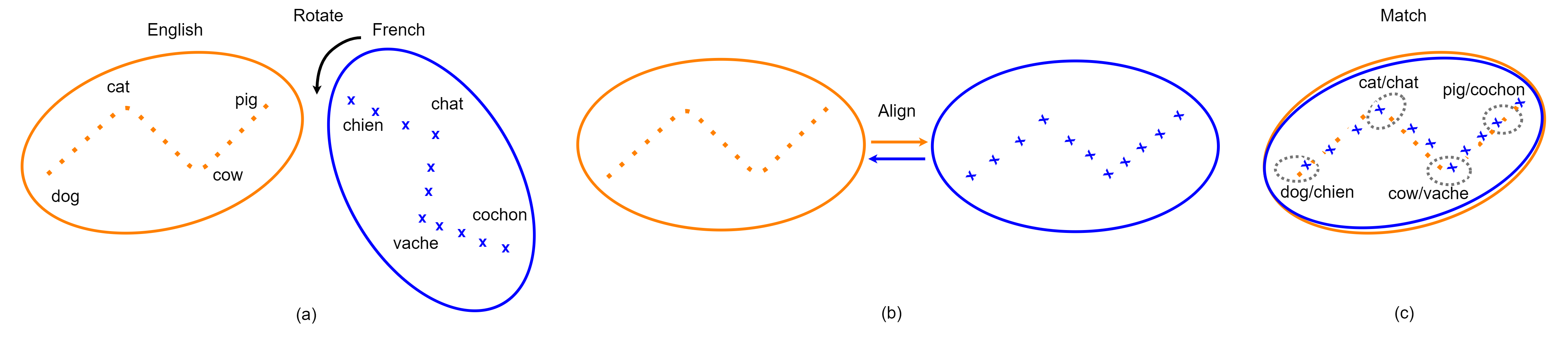}
    \caption{The initialization step of unsupervised NMT, where two languages are mapped to a common space.  The input is a trained embedding for each language and a dictionary of mapped words.  The dictionary pairs help guide the two embeddings by a) rotation and  b) alignment.  The resulting embedding space of the dictionary pairs is matched. }
    \label{fig:unsupervised}
\end{figure}

\textbf{Initialization}:
The underlying goal of the initialization step is to bridge the gap between the input representations of the different languages in an unsupervised manner. 
As shown in Figure~\ref{fig:unsupervised}, the unsupervised NMT model is initialized by learning a mapping between two or more languages. 
The intuition is that human beings live a shared common experience in the same physical world. Thus, the embedding of different languages should have a shared mathematical context. Researchers have experimented with various linguistic resources  and neural input representations for initialisation. \\
The traditional lexical resources include bilingual dictionaries \cite{artetxe2018unsupervised,lample2018unsupervised,lample-etal-2018-phrase, duan2020bilingual} or word-by-word gloss \cite{pourdamghani-etal-2019-translating} (inferred by aligning words \cite{conneau2017word}, phrases \cite{lample-etal-2018-phrase, artetxe-etal-2019-effective}, or sub-words \cite{artetxe-etal-2019-effective}).\\
Neural representations include cross-lingual n-grams, word embeddings, language models (LMs) or dependency embeddings \cite{lample2019cross,chronopoulou-etal-2020-reusing, ren-etal-2019-explicit, edman-etal-2020-low}. 
These neural input representations are either jointly trained by concatenating the source and target monolingual corpora or by learning the  transformation between the separate monolingual embeddings to map them into the shared space. One such technique is to leverage the available bilingual dictionaries by initialising the model with bilingual word embedding \cite{duan2020bilingual, lample2018unsupervised,  artetxe2018unsupervised}. Instead of words, using sub-word representations such as Byte Pair Encoding (BPE) has shown more promise~\cite{lample-etal-2018-phrase}.
    


    
In recent works, it has been shown  that the cross-lingual masked LMs could be more effective in initialising the models \cite{lample2019cross}. During training, the LM tries to predict the percentage of tokens that are randomly masked in the input. \citet{ren-etal-2019-explicit} further extended on the same lines by using n-grams instead of BPE tokens and inferred the cross-lingual n-gram translation tables. 



\textbf{Back-Translation}:  
Next, the generative step uses back-translation (discussed previously in Section~\ref{sec:dataAug})  by a denoising autoencoder that combines forward and backward translation from source to target, then target back to the source.  The loss function compares the original source text against the doubly translated text.  Again, the intuition is that there exists a common latent space between two languages so that the model can reconstruct the sentence in a given noisy version and then reconstruct the source sentence given noisy translation.\\
Recent back-translation based on multi-NMT (Section~\ref{sec:multiNMT}) models have shown much better results compared to the bilingual counterpart     ~\cite{sun-etal-2020-knowledge, sen2019multilingual, garcia2020multilingual, li2020reference, wang2020crosslingual}.

\textbf{Adversarial Architecture}:   
Finally, a discriminative step uses a binary classifier to differentiate the source language from the target language by distinguishing translated target text from original target text. An adversarial loss function trades-off between the reconstruction loss from the back-translation against the discrimination loss from the classifier.  The result of this step is a high-quality translation that is more fluent for LRLs.\\
Existing methods in the unsupervised NMT literature modifies the adversarial framework by incorporating additional adversarial steps or additional loss functions into the optimization step. These include dual cycle-GAN architecture \cite{artetxe-etal-2019-effective} and local-global GAN \cite{yang-etal-2018-unsupervised}. On the other hand, those methods that add a loss function include embedding agreement  \cite{sun-etal-2019-unsupervised}, edit and extract \cite{wu-etal-2019-extract}, and comparative translations \cite{sun-etal-2019-unsupervised}.

\subsubsection{Unsupervised NMT for Low-Resource Languages}

The majority of the early unsupervised techniques focused on high-resource languages that have monolingual data in abundance \cite{ren2020retrieve, sen2019multilingual, sogaard-etal-2018-limitations, wu-etal-2019-extract, yang2018unsupervised}; except for English-Russian and English-Romanian \cite{duan2020bilingual, ren-etal-2019-explicit, yang2017improving}. 
However, having the required input representation for the LRLs is a limitation  because some LRLs do not have bilingual dictionaries or proper word alignments.  On the other hand, to build a neural representation, large monolingual corpora are needed.  How these resources perform in extreme LRLs has not been properly studied. More recent work that explored the conditions for using unsupervised NMT for LRLs having less monolingual data is a promising development~\cite{chronopoulou-etal-2020-reusing, edman-etal-2020-low,kim-etal-2020-unsupervised, DBLP:conf/wmt/MarchisioDK20}.   \\ 
Various researchers have found a reduced translation quality for LRL pairs that are not from similar linguistic families or similar domains \cite{DBLP:conf/wmt/MarchisioDK20, edman-etal-2020-low, kim-etal-2020-unsupervised}. 
Four areas of concerns are: i) different script and dissimilar language, ii) imperfect domain alignment or domain mismatch, iii) diverse datasets, and iv) extremely LRLs \cite{DBLP:conf/wmt/MarchisioDK20}.
To resolve issues i and ii, \citet{kim-etal-2020-unsupervised} proposed robust training by using language models that are agnostic to language similarity and domain similarity, while \citet{chronopoulou-etal-2020-reusing} resolved issue i by combining transfer learning from HRL to LRL with unsupervised NMT to improve translations on a low-resource supervised setup.

\subsection{Semi-Supervised NMT}
\label{semi-supervised}

In contrast to unsupervised techniques, semi-supervised techniques assume the availability of some amount of parallel corpora alongside monolingual corpora. Semi-supervised techniques can be categorised according to the way the monolingual data is utilised: 

\textbf{Using monolingual data to generate synthetic parallel data:} The simplest strategy is to create the synthetic parallel corpus (this is essentially data augmentation) either by 1) copying monolingual data of one language as the translated text~\cite{currey-etal-2017-copied}, or 2) creating the source side with a null token~\cite{sennrich-etal-2016-improving}. A better way to generate synthetic data is through back-translation, as discussed in Section~\ref{sec:dataAug}.

\textbf{Using monolingual data to generate a language model (LM):} A LM can be integrated into the target-side of the NMT to improve the fluency of the generated text. This is named LM fusion, which can be broadly categorised as shallow fusion and deep fusion~\cite{gulcehre2015using}. In shallow fusion, the LM is used to score the candidate words generated by the decoder of the NMT system either during inference time~\cite{gulcehre2015using, skorokhodov2018semi}, or training time~\cite{stahlberg2018simple}. In deep fusion, the NMT architecture is modified to concatenate the LM with the decoder. Deep fusion provides better performance. However, LM model fusion has few limitations: 1) The NMT model and LM are trained independently and are not fine-tuned, 2) LM is only used at the decoder, 3) in deep fusion, only the final layers of the LM are integrated, disregarding the low-level LM features, and 4) the NMT architecture has to be changed to integrate the LM~\cite{baziotis2020language, ramachandran2016unsupervised}. 
    
Instead of LM fusion,~\citet{baziotis2020language} used the trained LM model as a weakly informative prior, which drives the output distributions of the NMT model to be probable under the distributions of the LM. This does not require any change to the NMT architecture.  
Another alternative is to use LMs to initialize the NMT model.~\citet{ramachandran2016unsupervised} initialized both encoder and decoder with the LMs of respective languages, while~\citet{abdou2017variable} used source embeddings to initialize the encoder. Following this line of research, recently there have been initiatives to incorporate BERT fine-tuning for NMT~\cite{zhu2019incorporating}. 
A promising extension is the use of pre-training multilingual LMs, such as mBART, in the form of an autoregressive sequence-to-sequence model~\cite{liu2020multilingual}. 
    
\textbf{Changing the NMT training objective to incorporate monolingual data: }
\citet{cheng2016semi} appended a reconstruction term to the training objective, which reconstructs the observed monolingual corpora using an autoencoder. This method assumes both source and target monolingual corpora are available. They jointly train source-to-target and target-to-source translation models, which serve as the encoder and decoder (respectively) for the autoencoder. \citet{zhang2018joint} also made use of both source and target monolingual data and employed source-to-target and target-to-source translation models. They introduced a new training objective by adding a joint Expectation Maximization (EM) estimation over the monolingual data to the Maximum Likelihood Estimation (MLE) over parallel data. 

\textbf{Multi-task learning:} Here, separate NMT models are used. \citet{zhang-zong-2016-exploiting} trained one model on the aligned sentence pairs to predict the target sentence from the source sentence, while the other is trained on the source monolingual data to predict the reordered source sentence from original source sentences. In essence, they strengthened the encoder using source-side monolingual data.~\citet{domhan2017using} followed a similar approach, however, they strengthened the decoder using target-side monolingual data. A similar technique is joint learning, where the source-to-target and target-to-source translation models, as well as language models, are aligned through a shared latent semantic space~\cite{zheng2019mirror}.

\textbf{Dual Learning:} Dual learning is based on the concept of Reinforcement Learning (RL) and requires monolingual data on both sides. Parallel data is used to build two weak source-to-target and target-to-source translation models. Then, monolingual data of both sides undergo a two-hop translation. For example, source side data is first translated using the source-to-target model, the output of which is again translated by the target-to-source model. This final output is evaluated against the original monolingual sentence and is used as a reward signal to improve the translation models~\cite{he2016dual}. This process is carried out iteratively and shows some resemblance to iterative BT~\cite{zheng2019mirror}. However, RL based techniques are known to be very inefficient~\cite{wu2018study, wang2018dual}.~\citet{wu2018study} also argued that the above RL based technique does not properly exploit the monolingual data, and suggested several improvements.~\citet{wang2018dual} transferred the knowledge learned in this dual translation task into the primary source-to-target translation task.      
\subsubsection{Semi-supervised NMT for Low-Resource Languages}
Although semi-supervised techniques have been presented as a solution to the scarcity of parallel data, we note the following concerns with respect to their applicability in the context of LRLs: 1) A LRL translation scenario has been simulated by taking small amounts of parallel data from high-resource languages such as English, French, German, and Chinese. 2) Some research has employed very large monolingual corpora. Although many LRLs have monolingual corpora with sizes larger than parallel corpora, it is difficult to assume they would have such large amounts of monolingual corpora, 3) Lack of comparative evaluations across the different semi-supervised techniques. Except for a few research~\cite{wang2018dual, zheng2019mirror, baziotis2020language}, most of the others compared with back-translation only. Interestingly some reported results less than BT~\cite{domhan2017using} and iterative BT~\cite{xu2020dual}, while some reported only marginal gains over BT~\cite{wu2018study, wang2018dual, zheng2019mirror}. This makes one doubt the actual benefit of these sophisticated techniques. Thus to establish the usefulness of these techniques for true LRLs, experiments should be carried out concerning a wide array of languages and different monolingual dataset sizes.\\
Although integrating massive language models such as BERT have shown promising results, these techniques have also been tested with high-resource languages. How these models would work with models built with rather small amounts of monolingual data should be investigated\footnote{Note that there is a long list of very recent publications in this line. However, due to this reason, above we cited only one.}. However, multilingual models such as mBART are indeed very promising for the translation of LRL pairs, which has been already proven~\cite{fan2020englishcentric} as further discussed in Section~\ref{sec:multiNMT}.   

\subsection{Multilingual NMT}
\label{sec:multiNMT}
Multilingual NMT (multi-NMT) systems are those handling translation between more than one language pair~\cite{ha2016toward,dabre2020survey}. Recent research has shown multilingual models outperform their bilingual counterpart, in particular when the number of languages in the system is small and those languages are related \cite{lakew-etal-2018-comparison, tan2018multilingual}. Particularly, in English-centric datasets, multi-NMT models trained with roughly $50$ languages have shown clear performance gains over bilingual models for LRLs~\cite{arivazhagan2019massively}. This is mainly due to the capability of the model to learn an \textit{interlingua} (shared semantic representation between languages)~\cite{johnson2017google}. Training multi-NMT models is a more practical solution as opposed to building separate bilingual models in a real-world setting~\cite{arivazhagan2019massively}.

Despite these benefits, multi-NMT faces challenging problems such as i) Inclusion of a large number of languages that have varying differences among them, ii) noise especially in the parallel data used, iii) data imbalance (some languages just having a fraction of parallel sentences compared to high-resource languages), and iv) other discrepancies concerning factors such as writing style and topic~\cite{arivazhagan2019massively}.  

\begin{figure}
    \centering
    \includegraphics[scale = 0.5]{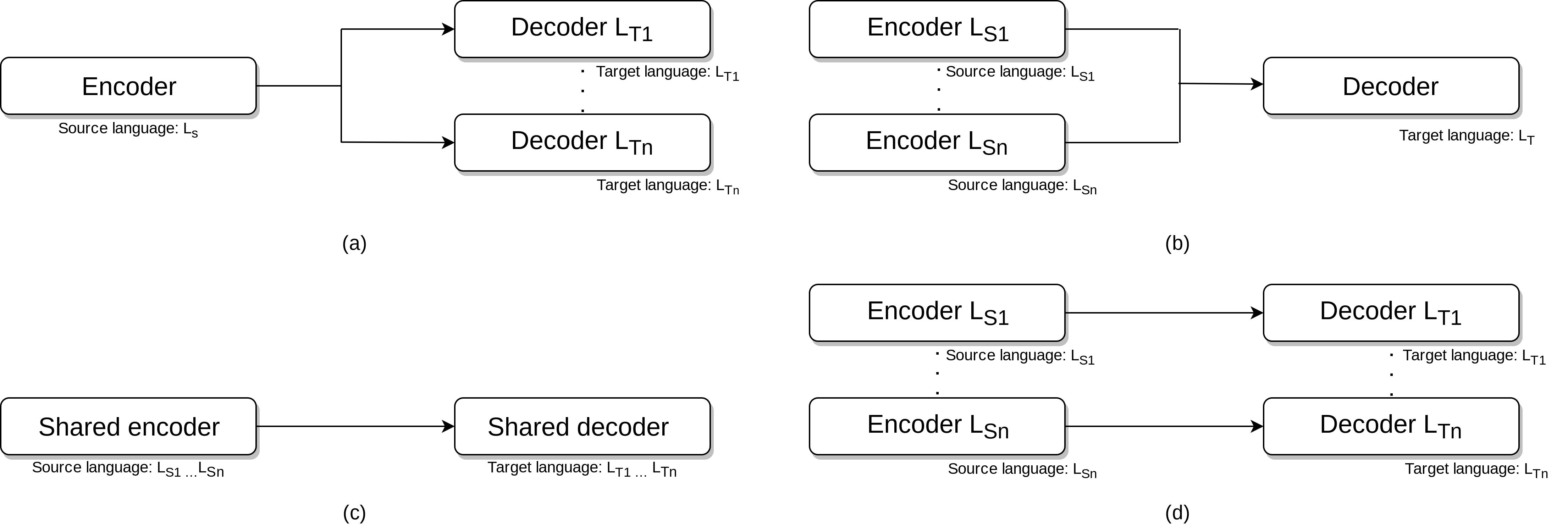}
    \caption{Supervised multi-NMT architectures.}
    \label{fig:multiNMT}
\end{figure}

With respect to the translation task, multi-NMT can be categorised into three types (Figure~\ref{fig:multiNMT}): 
\begin{enumerate}
    \item Translating from one source language to multiple target languages, (one-to-many) (Figure~\ref{fig:multiNMT} (a)): This is essentially a multi-task problem, where each target becomes a new task \cite{dong2015multi, sachan-neubig-2018-parameter}.
    \item Translating from multiple source languages to a single target language, (many-to-one) (Figure~\ref{fig:multiNMT} (b)): This can be considered as a multi-source problem, considered relatively easier than the multi-task problem \cite{zoph-knight-2016-multi, garmash2016ensemble}.
    \item Translating from multiple languages to multiple languages,(many-to-many) (Figure~\ref{fig:multiNMT} (c) and (d)). This is the multi-source, multi-target problem, and the most difficult scenario \cite{firat-etal-2016-multi, johnson2017google}.
\end{enumerate}

Supervised multi-NMT architectures introduced to tackle the aforementioned translation tasks can be broadly categorised into three paradigms: i) single encoder-decoder for all the languages (all source sentences are fed into the encoder irrespective of the language, and the decoder can generate any of the target languages (Figure~\ref{fig:multiNMT}. (c)); ii) per-language encoder-decoder (each source language has its own encoder, and each target language has its own decoder (Figure~\ref{fig:multiNMT}. (d)); and iii) shared (a single) encoder/decoder at one side, with per-language decoder/encoder at the other side (Figure~\ref{fig:multiNMT}. (a) and (b)). The main objective of these different architectures is to maximize the common information shared across languages while retaining language-specific information to distinguish between different languages. This mainly depends on how parameters are shared between individual encoders and decoders.  All of these architectures are based on either the recurrent model with attention \cite{johnson2017google}, or the transformer-based model~\cite{vaswani2018tensor2tensor}. Comparison of the recurrent model against the transformer model under the same settings has shown that the latter is better \cite{lakew-etal-2018-comparison, sachan-neubig-2018-parameter, lakew2019multilingual}. Almost all the recent multi-NMT architectures are based on the transformer model.

   \textbf{Single encoder-decoder for all the languages:}
For large-scale multi-NMT implementations, this is currently the state-of-the-art, especially in real-world industry-level systems~\cite{DBLP:journals/corr/JohnsonSLKWCTVW16, aharoni-etal-2019-massively, arivazhagan2019massively}. Since all the source languages share the same encoder and all the target languages share the same decoder, while simultaneously supporting one-to-many, many-to-one, and many-to-many cases, this model is commonly known as the `\textit{universal NMT model}'. The main advantage of a universal model is lower model complexity compared to per language encoder-decoder models (discussed next) because it has a lower parameter count.
Moreover, as demonstrated by \citet{johnson2017google}, this universal model is capable of learning a form of interlingua, which is crucial in facilitating zero-shot translation (see Section~\ref{sec:zero_shot}). 
A major challenge in using this architecture is enabling the decoder to distinguish the target language. The common practice is to add a  language identification tag to the source sentence~\cite{johnson2017google, wang-etal-2018-three, ha2016toward, blackwood-etal-2018-multilingual}. An alternative is to add the language name as an input feature~\cite{ha2017effective}.
More recent work has used language-dependent positional embeddings representations ~\cite{wang-etal-2018-three, wang2019compact, zhu2020language}.

\textbf{Per-language encoder-decoder:} 
In this architecture, there is a separate encoder per source language, as well as a separate decoder per each target language\cite{firat-etal-2016-multi}. As opposed to the universal NMT models described above, the requirement to capture language-independent features can be easily achieved by this setup. However, sharing common information across languages is a challenge. The commonly applied solution is the use of shared parameters in the model, employing shared attention~\cite{firat-etal-2016-multi, lu2018neural, raganato2019evaluation, vazquez2018multilingual}.
\citet{Platanios2018ContextualPG} extended this idea even further, and introduced a contextual parameter generator, which enables the model to learn language-specific parameters, while sharing information between similar languages. 

\textbf{Single encoder with per-language decoder / per-language encoder with single decoder:} 
The single encoder per language with multiple decoder architecture supports the one-to-many scenario, where a single source language gets translated to multiple languages via multiple decoders (multi-task). The multiple encoders, single decoder architecture supports the many-to-one scenario, where multiple encoders process separate source languages while using a single decoder to translate into one target language (multi-source).

In the one-to-many scenario, each decoder has its attention mechanism, so no parameter sharing takes place at the decoder side~\cite{dong2015multi}. A more effective model is to allow partial sharing of parameters across decoders~\cite{sachan-neubig-2018-parameter, wang2019synchronously}. 
When there are multiple encoders alongside a single decoder, encoder output has to be combined to be sent to the decoder. Initial solutions assumed the used corpus is multi-way parallel~\cite{zoph-knight-2016-multi}. Ways to relax this restriction are either to mark a corresponding sentence as null if a particular language does not have that sentence~\cite{nishimura2018multi} or to generate the missing sentences with a pre-trained model~ \cite{nishimura2018multia}. 


Many research has evaluated the pros and cons of these different architectures. For example, \citet{hokamp2019evaluating} showed that a unique decoder for each target language or unique decoder attention parameters for each target language outperform models with fully shared decoder parameters.~\citet{sachan-neubig-2018-parameter} obtained better results with partial parameter sharing in the transformer model, over the full-parameter sharing recurrent model of \citet{johnson2017google}. However, the best model selection would depend on the nature of the task at hand. For example, if the model is expected to deal with hundreds of languages, it is desirable to have maximum parameter sharing like in \citet{johnson2017google}, to reduce the model complexity \cite{arivazhagan2019massively}.

\subsubsection{Multi-NMT for Low-Resource Languages}
\label{Low-Resource_Languages}

All of the three supervised multi-NMT techniques discussed above have been leveraged for LRL translation. In addition, the multilingual version of unsupervised and semi-supervised NMT models, as well as transfer learning on multi-NMT parent models have been used for LRL translation. 

\textbf{1. Supervised multi-NMT architectures:} In the available multilingual datasets, the LRL pairs are heavily under-represented. Thus, the results of supervised multi-NMT models for the LRL pairs are far below the results for high-resource languages, even though they use the same multi-NMT model~\cite{aharoni-etal-2019-massively}. The simplest strategy to alleviate this problem would be to over-sample the parallel corpus related to the LRL pair. There are different sampling strategies such as simple over-sampling and temperature-based sampling~\cite{firat-etal-2016-multi, arivazhagan2019massively, wang-neubig-2019-target, wang2020balancing}. Sampling data into mini-batches also has to be given careful consideration, to avoid any form of bias. Some strategies include scheduling (cycle through the bilingual language pairs)~\cite{firat-etal-2016-multi}, using mini-batches that consist of different languages~\cite{johnson2017google, sachan-neubig-2018-parameter}, or using mini-batches that contain data from the same target~\cite{blackwood-etal-2018-multilingual}. Data augmentation~\cite{Platanios2018ContextualPG} (Section~\ref{sec:dataAug}) and pivoting~\cite{firat-etal-2016-zero} (Section~\ref{sec:zero_shot}) can also be used to generate parallel data for LRL pairs.

    
\textbf{2. Unsupervised multi-NMT:} When a multi-NMT model is built entirely upon monolingual data, it is referred to as unsupervised multi-NMT, which are trained following a similar process to that of bilingual unsupervised models (see Section~\ref{unsupervised_nmt}). The difference between bilingual and multilingual NMT comes from the way the input representation is constructed. In the English-centric unsupervised model proposed by \citet{sen2019multilingual}, first, the embeddings of non-English languages are mapped into the latent space of English embeddings.~ \citet{sun-etal-2020-knowledge} constructed a multilingual masked language model using only a single encoder. Better results over the pure unsupervised model can be obtained if at least one language has parallel data with some other language~\cite{garcia2020multilingual, li2020reference, wang2020crosslingual}.

\textbf{3. Semi-supervised multi-NMT} In semi-supervised multi-NMT, monolingual data is used to create an additional training objective on top of the supervised translation training objective. While~\citet{siddhant-etal-2020-leveraging} used the MASS objective~\cite{song2019mass} for this purpose,~\citet{wang2020multi} employed two monolingual auxiliary tasks: masked language modelling (MLM) for the source-side, and denoising autoencoding for the target side. Semi-supervised NMT is further discussed in Section~\ref{semi-supervised}.
    
\textbf{4. Transfer Learning on a pre-trained multi-NMT model:} Transfer Learning is discussed in Section~\ref{transfer_learning_nmt}. Here we note that transfer learning using a multilingual parent has been identified as a promising approach for LRL-NMT~\cite{dabre-etal-2019-exploiting, gu2018meta, neubig2018rapid, gu2018universal, lakew2018transfer}. In particular, some LRL data may not be available during multi-NMT training time and very large multilingual models cannot be re-trained every time parallel data for a new language pair becomes available  \cite{lakew2018transfer,nguyen-2017-transfer}.

\textbf{Input Representation:} Despite the multi-NMT methodology selected, a major factor that decides the success of multi-NMT for LRLs is the input representation. The input representation determines the ability to group semantically similar words from different languages in the embedding space. Input representation can be broadly broken down into two categories: surface form (word-level) representation, and embedding-based representation.

When the surface form representation is used, a semantic grouping of words is achieved by adding a language token to each word~\cite{ha2016toward}, or by using additional information s.a POS of words~\cite{espana2017going}. However, using word-level input results in large vocabulary sizes that are difficult to scale \cite{Platanios2018ContextualPG}. Even if there are linguistically similar languages that share a common script, the amount of vocabulary overlap is minimal~\cite{dabre2020survey}. LRLs are severely affected by this~\cite{gu2018universal}.

As a solution, sub-word level encoding (BPE~\cite{johnson2017google, sachan-neubig-2018-parameter}, sentence piece representation~\cite{arivazhagan2019massively}, or transliteration~\cite{maimaiti2019multi, goyal-etal-2020-efficient} was used. 
\citet{gu2018universal} noted that even sub-word level encoding does not create enough overlap for extremely LRLs, since it still uses the surface form of the word. Further, as \citet{wang2018multilingual} pointed out, with such sub-word based techniques, semantically similar and similarly spelt words could get split into different sub-words for different languages. 

An alternative is to use input representations based on cross-lingual embeddings. \citet{qi-etal-2018-pre} argued that, when the input languages are in the same semantic space, the encoder has to learn a relatively simple transform of the input. Moreover, in such shared spaces, LRLs get enriched with more semantic information with the help of high-resource languages. Such universal embedding representations have shown very promising results for LR as well as extremely LRL pairs~\cite{gu2018universal, wang2018multilingual}.
A very interesting development is the use of multilingual denoising models pre-trained on monolingual data of a large number of languages (e.g. mBART~\cite{liu2020multilingual}), which can be fine-tuned on multilingual translation tasks. This has shown very promising results for LRL pairs~\cite{fan2020englishcentric}.  

In summary, we see research efforts on multiple fronts to leverage multi-NMT for LRLs. While earlier research experimented with datasets sub-sampled from high-resource language pairs\cite{firat-etal-2016-multi, dong2015multi, ha2016toward}, later research has experimented with actual LRLs~\cite{maimaiti2019multi}. Moreover, it is encouraging to see very promising results from transfer learning over multi-NMT models, as training large multi-NMT models is time-consuming. However, most of this research has been carried out holistically, without focusing on individual language, or language-family characteristics. How these characteristics can be exploited to better leverage NMT for LRLs would result in multi-NMT models that are more focused on a required set of languages.

\subsection{Transfer Learning in NMT}
\label{sec:transfer_learning}

\label{transfer_learning_nmt}
\begin{figure}
    \centering
    \includegraphics[scale = 0.5]{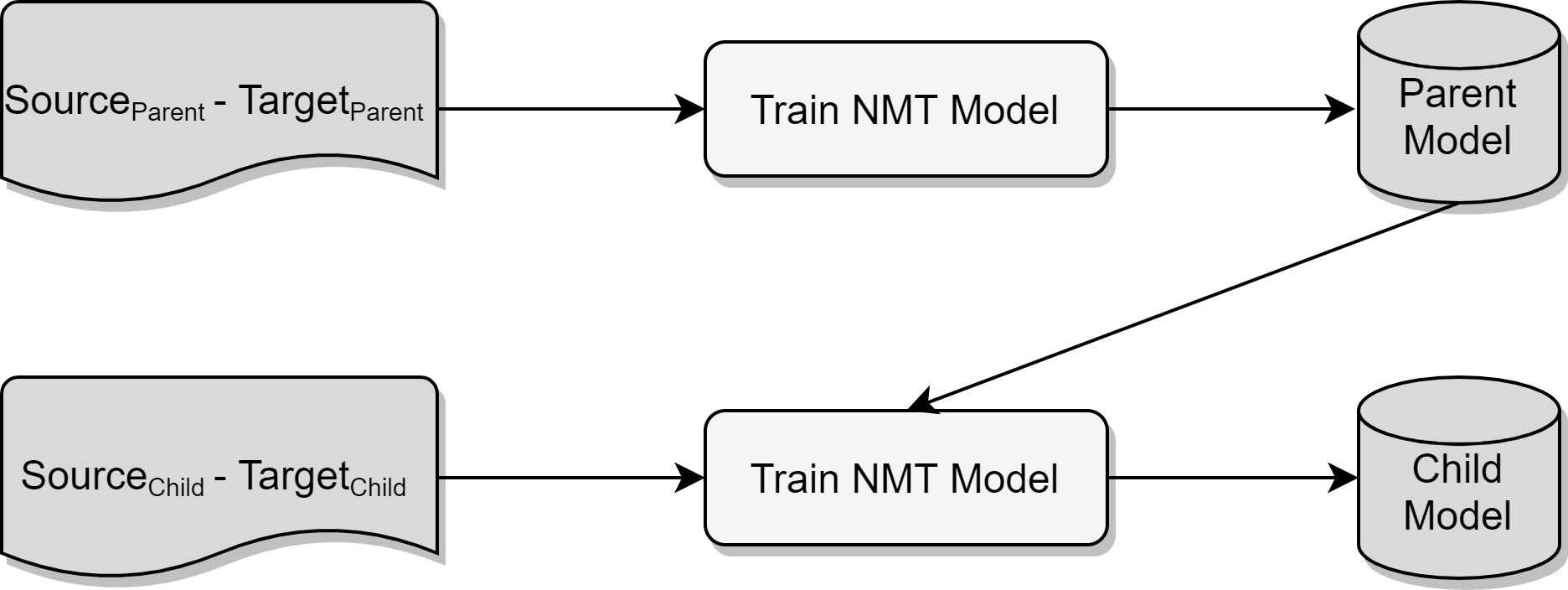}
    \caption{Transfer Learning process.}
    \label{fig:transfer_leraning}
\end{figure}

Transfer learning is a sub-area in Machine Learning that reuses (i.e.~transfers or adapts) knowledge that is gained from solving one particular task, problem, or model (parent) by applying it to a different but related one (child)~\cite{pan2009survey}.~\citet{zoph2016transfer} first introduced the viability of transfer learning for NMT. In NMT, the parent model is first trained on a large corpus of parallel data from a high-resource language pair (or pairs), which is then used to initialize the parameters of a child model that is trained on a relatively smaller parallel corpus of the LRL pair (Figure~\ref{fig:transfer_leraning}).  

The advantages of transferring knowledge from the parent model to the child model include i) reducing the size requirement on child training data, ii) improving the performance of the child task, and iii) faster convergence compared to child models trained from scratch.

The transfer process in NMT models can be broadly categorised as either warm-start and cold-start~\cite{neubig2018rapid}. Due to the availability of child parallel data during parent model training, warm-start systems are more accurate and has been the focus of most of the previous work~\cite{zoph2016transfer, dabre2017empirical, nguyen-2017-transfer, kocmi2018trivial, maimaiti2020enriching}. However, cold-start systems are also of importance due to their resemblance to a real-life scenario where child parallel data is not always available at parent model training time~\cite{lakew2018transfer, kim2019effective, kocmi2020efficiently}. 

As shown in Figure~\ref{fig:transfer_leraning}, the first step in transfer learning is to train a parent model, which could be either bilingual or multilingual (note that the source and target in multi-NMT models can be many-to-one, one-to-many, or many-to-many. A special case of multi-NMT based transfer learning is fine-tuning large-scale multilingual language models such as mBART using small amounts of parallel data~\cite{stickland2020recipes}, as already mentioned in Section~\ref{sec:multiNMT}). However, the bilingual parent model is more common. The majority of the time, the parent and child have the same target language ~\cite{zoph2016transfer, dabre2017empirical, nguyen-2017-transfer, murthy2018addressing, kim2019effective, aji2020neural, maimaiti2020enriching}, while others use the same source language for both the parent and child ~\cite{kocmi2020efficiently}.  However, it is also possible for the parent and child not to have shared languages in common~\cite{kocmi2018trivial, maimaiti2019multi, luo2019hierarchical}. Often,  multi-NMT models used as parents in transfer learning have been trained on the many-to-one setting~\cite{dabre-etal-2019-exploiting, gu2018meta, neubig2018rapid, gu2018universal, lakew2018transfer}. Despite the parent model being trained on either bilingual or multilingual source-target languages, the child has always been bilingual with the exception of~\citet{lakew2018transfer}, which progressively fine-tuned a parent model in order to build a model that adequately performs on multiple language pairs.


Improvements in transfer learning for NMT corresponds to three main aspects: i) minimizing the language space mismatch between languages, ii) fine-tuning technique and iii) the transfer protocol. 

\textbf{Minimizing the language space mismatch}: Transfer learning systems have to address the problem of language space mismatch, since parent and child languages may not have the same feature distribution~\cite{ji2020cross}. When the surface form is used as input, this language mismatch problem becomes a vocabulary mismatch between parent and child models. In the warm-start systems, sub-word segmentation models can be applied to the parent and child training data to build joint vocabularies~\cite{neubig2018rapid, dabre-etal-2019-exploiting}.~\citet{gheini2019universal} took this idea even further and introduced a universal vocabulary for the parent to train on.  
~\citet{lakew2018transfer} showed that a vocabulary based on sub-wording can be employed even in the cold-start scenarios by building a dynamic vocabulary. However, for cold-start scenarios, the better alternative is to pre-train a universal input representation, including child monolingual data, if available~\cite{gu2018universal,kim2019effective, ji2020cross}.       


\textbf{Fine-tuning technique:} Transferring knowledge from the parent model to the child model requires fine-tuning  the parameters trained in the parent model on the child dataset. Conversely, when a particular layer of the parent model is not fine-tuned, this is called freezing that layer of the parent model.
Below we list the research experiments with differing fine-tuning strategies, where the best freezing setup depends on factors such as the neural architecture employed, the translation task, and the dataset size. Thus we do not draw any conclusions on the best fine-tuning strategy.

\textit{No fine-tuning: }The whole parent model is frozen (in other words, copied) to the child~\cite{aji2020neural, ji2020cross}. 

\textit{Fine-tune the embedding layer:} Similarity between parent and child language pairs (e.g. whether parent and child have the same target) determines which embedding has to be fine-tuned. For example, if the parent and child translate to the same target language, parent decoder embeddings can be transferred to the child~\cite{zoph2016transfer}. When the surface form input is used, the most naive way of transferring the embedding layer is to randomly initialize the parent embedding layer before training the child model~\cite{zoph2016transfer}.  A better alternative is to take the parent and child vocabulary overlap  while replacing the rest of the parent embeddings with child embeddings~\cite{nguyen-2017-transfer, lakew2018transfer}.
    
\textit{Fine-tune all of the parent model:} No layer of the parent model is freezed~\cite{lakew2018transfer, maimaiti2019multi, kocmi2020efficiently, ji2020cross}. 

\textit{Fine-tune a custom set of layers:} This includes fine-tuning a selected combination on input, and inner layers of the encoder and decoder ~\cite{zoph2016transfer, kim2019effective, kocmi2020efficiently, aji2020neural}. 


\textbf{Transfer Protocol: }Varying the transfer protocol is also a promising way to improve NMT transfer learning. This can be done in different forms: 
\begin{itemize}

    \item Train a chain of consecutive NMT models by transferring the parameters of a parent model to new LRL pairs~\cite{lakew2018transfer}. 
    \item Train the initial NMT model on a parallel corpus for a resource-rich language pair, fine-tune it with the combined corpus of parent and child (can be more than one child), and finally, fine-tune further with the selected child data only~\cite{dabre-etal-2019-exploiting, maimaiti2020enriching}.
    \item First train the unrelated high-resource language pair, then fine-tune it on a similar intermediate language pair and finally fine-tune on the LRL pair~\cite{imankulova-etal-2019-exploiting, luo2019hierarchical}.
\end{itemize}

Multiple factors determine the success of transfer learning. The relationship between the languages used in parent and child models has been identified as the most crucial~\cite{zoph2016transfer, dabre2017empirical, nguyen-2017-transfer}. 
High relatedness between languages guarantees high vocabulary overlap when the surface form is used as input and would result in more meaningful cross-lingual embeddings as well. Much research has exploited vocabulary overlap of related languages using sub-word segmentation to achieve good results with transfer learning~\cite{nguyen-2017-transfer}, even when parent and child have no language in common~\cite{kocmi2018trivial}. An important consideration is the size of the sub-word vocabulary. It should be selected in such a way that the child is not overwhelmed by the parent~\cite{neubig2018rapid, kocmi2018trivial}. For related languages, transliteration has shown to reduce lexical divergence~\cite{nguyen-2017-transfer, maimaiti2019multi, goyal-etal-2020-efficient}. The syntactic divergence between parent and child can be reduced by re-ordering the parent data~\cite{murthy2018addressing, kim2019effective}. Other factors that influence transfer learning include the size of the parent and child corpora, domains of the parent and child data, the number of shared words (vocabulary overlap), and the language script~\cite{kocmi2018trivial, lin2019choosing, aji2020neural,dabre-etal-2019-exploiting, kocmi2020efficiently}.


\subsubsection{Transfer Learning for Low-Resource Languages}
Transfer learning was originally introduced as a solution to low-resource (both domain and language) NMT. With respect to translation of LRL pairs, transfer learning using a high-resource pair always yielded better results than training the child model from scratch. This holds even for extremely LR children as well~\cite{lakew2018transfer}. Interestingly, some research has shown that transfer learning is better than training a child pair (or a set of pairs ) with one or more parent pairs in a multi-NMT manner~\cite{lakew2018transfer, kim2019effective, maimaiti2019multi, maimaiti2020enriching}. However, that research has been conducted against an early multi-NMT model~\cite{johnson2017google}, considering very few languages. Whether the same observation would hold if a more novel multi-NMT model (discussed in Section~\ref{sec:multiNMT}) is used along with a large number of language pairs should be subject to more research. On the other hand, transfer learning using pre-trained multi-NMT parent models has received only limited attention~\cite{neubig2018rapid, gu2018universal, goyal-etal-2020-efficient}. As mentioned above, multiple factors affect the success of transfer learning. Thus the impact of these factors should be evaluated extensively to determine their exact impact on LRL-NMT. Zero-shot translation adds an extra condition to the cold-start scenario, meaning that child parallel data unavailable, as discussed in Section~\ref{sec:zero_shot}. 

\subsection{Zero-shot NMT}
\label{sec:zero_shot}
In the zero-shot scenario, no parallel corpus is available for the considered source(X)-target(Z) language pair. We have identified pivoting, transfer learning, multi-NMT and unsupervised NMT as existing solutions in the literature for zero-shot NMT. 

\textbf{Pivot-based solutions:}

An initial solution for zero-shot translation was the pivot-based translation, also known as pivoting. Pivoting relies on the availability of an intermediate high-resource language (Y), called the `pivot language'. In pivoting, the translation of X-Z is decomposed into the problem of training the two high-resource independent models: source-pivot (X-Y) and pivot-target (Y-Z). A source sentence is first translated using the X-Y model, the output of which is again translated using the Y-Z model to obtain the target sentence.\\
This basic form of pivoting has two main limitations. First, it suffers from the error propagation problem. Since the source-pivot and pivot-target models are independently trained, errors made in the first phase are propagated into the second phase. This is particularly the case when the source-pivot and pivot-target languages are distantly related. Second,  as models have to be independently trained, the total time complexity is increased. 

To reduce the problem of error propagation, source-pivot and pivot-target models can be allowed to interact with each other during training by sharing the word embedding of the pivot language~\cite{cheng2017joint}. Another solution is to combine pivoting with transfer learning~\cite{kim2019pivot}. Here, the high-resource source-pivot and pivot-target models are first independently trained as in the basic pivoting technique, acting as the parent models. Then the source-target model (child model) is initialized with the source encoder from the pre-trained source-pivot model, and the target decoder from the pivot-target model. In addition to reducing the error propagation, this method reduces time complexity, since only one trained model is used for translation. Another way to reduce error propagation is to use a source-pivot parallel corpus to guide the learning process of a pivot-target model~\cite{chen2017teacher}.~\citet{zheng2017maximum} proposed a similar approach by training the source-target model via Maximum Likelihood Estimation , where the training objective is to maximize the expectation concerning a pivot-source model for the intended source-to-target model on a pivot-target parallel corpus.

It has been shown that adding even small amounts of true parallel source-target sentences (thus the extremely low-resource scenario) does increase the translation accuracy in pivoting~\cite{cheng2017joint,kim2019pivot,chen2017teacher, ren2018triangular} . 
Another possibility is to make use of monolingual data to generate synthetic parallel data. Pivot monolingual data is preferred because compared to source or target, pivot language would have much more monolingual data~\cite{currey2019zero}.

A common observation of the above discussed pivoting research (with the exception of~\cite{ren2018triangular, currey2019zero}) is that, although the focus is on zero-shot translation between a source and target, large parallel corpora have been employed for source-pivot and pivot-target pairs. However, some LRLs may not have large parallel datasets even with a high-resource language such as English. Moreover, as empirically shown by~\citet{liu2018pivot}, the performance of pivoting depends on the relatedness of the selected languages. Thus, we believe that more research is needed to determine the impact of pivoting for zero-shot NMT in the context of LRL pairs.




\textbf{Transfer Learning-based Solutions}:
As mentioned in Section~\ref{sec:transfer_learning}, transfer learning can be considered a form of zero-shot translation, when no parallel data is available for the child model.~\citet{ji2020cross} explored transfer learning for zero-shot translation by mimicking the pivoting technique, assuming a high-resource pivot language. They use source-pivot data to build a universal encoder, which is then used to initialize a pivot-target model. This model is used to directly translate source sentences into target sentences.


\textbf{Multi-NMT-based Solutions:}
Although pivot-based models have been the solution for zero-shot NMT for a long time, recent research showed that multi-NMT models can outperform the pivot-based zero-shot translation~\cite{arivazhagan2019massively}. Many-to-many models have recently shown to beat the English-centric multi-NMT models in zero-shot settings~\cite{fan2020englishcentric}.  

A multi-NMT model can provide a reasonable translation between a source-target pair when the two languages are included in the model in the form of parallel data with any other language because the multi-NMT model is capable of learning an `interlingua' (see Section~\ref{sec:multiNMT}). If the multi-NMT model is truly capable of learning a language-independent interlingua, there should be less correlation between the source and target languages. However, when the model is trained with a large number of languages, the modelling capacity (loosely measured in terms of the number of free parameters for neural networks~\cite{aharoni-etal-2019-massively}) has to be distributed across all the languages, suggesting that the overall model capacity is faced with a bottleneck~\cite{arivazhagan2019massively, zhang2020improving}. Thus, the learned interlingua is not fully language independent. Two solutions have been presented to solve this problem: to explicitly make the source and target languages independent~\cite{gu-etal-2019-improved, sestorain2018zero, arivazhagan2019massively, pham2019improving}, and to improve model capacity~\cite{zhang2020improving, fan2020englishcentric}. 

As another solution, synthetic data can be generated between zero-shot language pairs using techniques such as pivoting \cite{firat-etal-2016-zero} and back-translation \cite{gu-etal-2019-improved, lakew2019multilingual, sen2019multilingual, zhang2020improving}. This synthetic parallel data is included in the multilingual corpus used to train the multi-NMT model. 


\textbf{Unsupervised NMT-based solutions:}
Unsupervised NMT techniques discussed in Section~\ref{unsupervised_nmt} rely only on monolingual data. Thus, this translation task can be considered as a zero-shot translation task.

\subsection{Analysis on the Popularity of LRL-NMT Techniques } 
\label{techniques_trend}
So far we have discussed seven different techniques that can be used for the translation of LRL pairs, as well as zero-shot translation . This section provides a quantitative view of the use of these techniques in the related literature.      

\begin{figure}
    \centering
    \includegraphics[width=\linewidth]{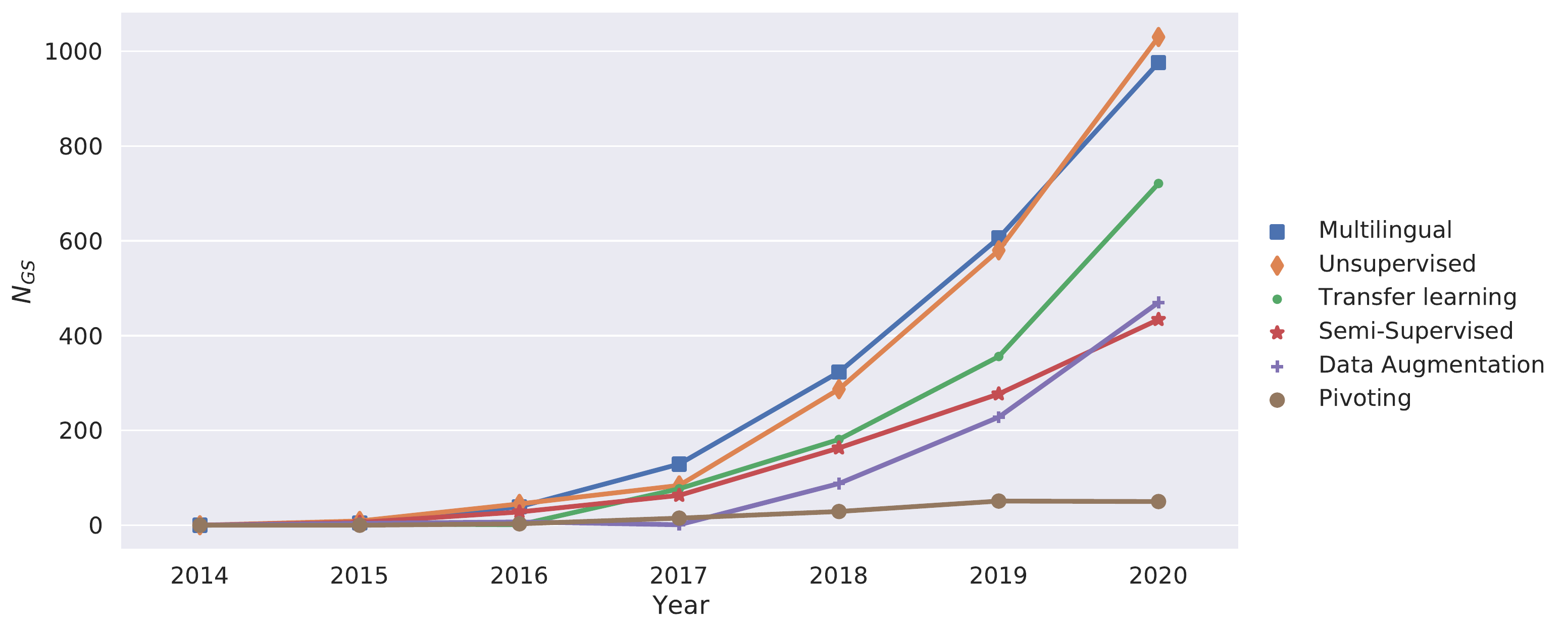}
    \caption{Number of Google Scholar search results($N_{GS}$) for different techniques from 2014-2020}
    \label{fig:tech_time_GS}
\end{figure}

 Figure~\ref{fig:tech_time_GS} shows how the use of different techniques varied from 2014 onwards based on the research papers indexed in Google Scholar. For each of the technique, Google Scholar was searched with the following query: ``\textit{<technique\_name>}'' +  ``\textit{low-resource}''  + ``\textit{neural machine translation}'' for the year range 2014-2020. However, we  acknowledge that the search results contain noise. For example, in certain cases, unsupervised NMT research was referred in unsupervised text generation papers. However, here we are only interested in a comparative view, thus we assume the noise is equally distributed across the search results for all the techniques. 

Figure~\ref{fig:tech_time_GS} shows that multi-NMT had the highest number of papers till 2019, however, unsupervised techniques have surpassed it marginally after that. 
In particular, the use of multi-NMT for LRL pairs started growing with the promising results shown by~\citet{firat-etal-2016-multi} and~\citet{johnson2017google} around 2016 and 2017. Transfer learning, and semi-supervised had similar growth till 2018, whereas from 2019 onwards transfer learning has seen a steep increase in popularity. Data augmentation techniques have gained popularity in 2020 as well. However, pivoting seems to have lost its traction, which may be due to the recent advancements in multi-NMT that outperformed pivoting for zero-shot translation~\cite{arivazhagan2019massively}. Overall, it can be seen that the interest of the NMT research community towards LRLs is steadily increasing irrespective of the type of technique.

\section{Guidelines to select a technique for a given data specification}
\label{technique_selection}

The effectiveness and viability of the techniques presented in Section~\ref{section:label-NMT-Techniques-for-Low-Resource-Languages} depend on the size and nature of the available parallel and monolingual data and the computational resources at hand. This section gives a set of guidelines to advise practitioners of LRL-NMT on the suitable technique for a particular data setup. These guidelines should not be construed as rigid criteria but only as an advisory for the practitioners. 

\begin{figure}[h]
    \centering
    \includegraphics[scale = 0.5]{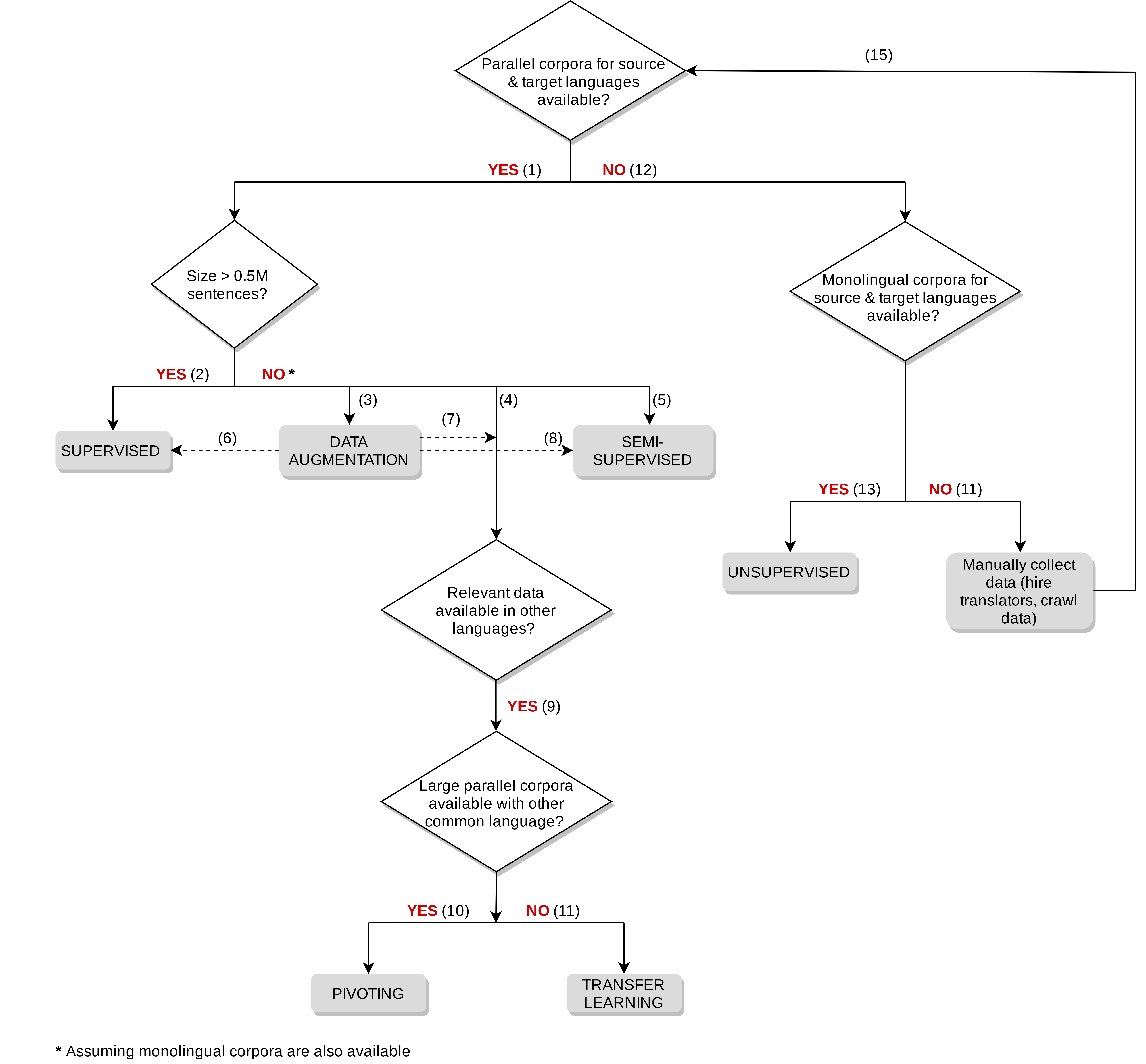}
    \caption{Flowchart describing selection of an appropriate technique for a given data specification}
    \label{fig:flowchart}
\end{figure}

Figure \ref{fig:flowchart} shows a possible process that can be followed in  selecting an NMT technique. This flowchart only provides guidelines for the bilingual scenario where parallel data is available for a pair of languages. However, it should be noted that if sufficient computing resources are available, the multilingual versions of all the techniques can be used as they have shown promising results with respect to LRL pairs. We did not show the multilingual scenario just for the sake of clarity.

We considered the availability and size of the parallel corpus, the availability of monolingual corpora, and language similarity as the major factors to select a technique.
The foremost factor that we have considered is the availability of parallel corpora. If a parallel corpus is available for a language pair, the next step is to check its size (step 1). There is no definite threshold suggested in the literature for the size of the parallel corpus to be considered an LRL scenario in NMT. However, following our discussion in Section~\ref{LR-Definition}, here we considered an LRL scenario where a particular language  pair has less than 0.5M parallel sentences. This is not a hard threshold but a mere suggestion. If a particular language pair has more than 0.5M sentences, we can achieve a reasonable performance through supervised NMT techniques (step 2). 

If the parallel corpus has less than 0.5M sentences, there could be multiple steps taken by a practitioner (as shown by steps 3-5 in Figure \ref{fig:flowchart}). One of the steps could be to increase the size of the dataset by using data augmentation (step 3) which is further followed by a supervised NMT technique (step 6). Data augmentation can be performed by using resources such as bilingual dictionaries and monolingual data. The other option could be to integrate the available monolingual and parallel data to perform semi-supervised NMT (step 5).

If the source and target languages have parallel data available with some other common language such as English, then we can also recommend attempting pivoting (step 10). However, if such parallel datasets are not available, a practitioner can attempt transfer learning (step 11). For this scenario, a parallel corpus between two high-resource languages can be used to build the parent model, which can further be fine-tuned to the LRL child. Transfer learning can be performed on multi-NMT models as well even when high-end GPU machines are not available. As discussed in Section ~\ref{sec:transfer_learning}, the effectiveness of transfer learning depends on the language relatedness, therefore the selection of parent model has to be done carefully. It should be noted that it is always possible to increase the original dataset size by applying data augmentation techniques, before applying pivoting, transfer learning, or semi-supervised solutions.

If the considered LRLs do not have parallel data but have a reasonable amount of monolingual corpora (which is a reasonable assumption for most of the LRLs), unsupervised NMT can be applied (step 13). If a considered language pair neither has any parallel data, nor the monolingual corpora\footnote{We refer to electronic resources, which could be the case with endangered languages that do not have any web presence.}, the only option is to manually create parallel and/or monolingual corpora (step 14).

We would like to conclude with the following two remarks. First, for each of the discussed LRL-NMT techniques, a large body of past related research is available; therefore, practitioners have to carefully select the most appropriate technique to be used as the baseline for their considered languages. This decision not only depends on the exact size of the available datasets but also the language characteristics and any other associated language/linguistic resources such as POS taggers and bilingual dictionaries. Second, LRL-NMT should be considered as an iterative process. For example, as shown by step (15), once a parallel corpus is manually created, the considered language pair now has a parallel  dataset less than 0.5M. With that, either transfer learning, or semi-supervised NMT can be tried out. With this trained model, more parallel data can be generated. Although the generated parallel data might not be 100\% accurate, the noise can be removed via post-processing (manual/automatic/hybrid) to obtain cleaner data. It is possible to train a new model with this newly cleaned data by integrating it with the original corpus.

\section{Landscape of Low-Resource Languages and NMT Research}
\label{section:trend_analysis}

There are over 7000 languages being spoken around the world. A look into the related research reported in Section~\ref{section:label-NMT-Techniques-for-Low-Resource-Languages} reveals the NMT techniques have mostly been tested on the same set of languages. Identifying reasons for this imbalance in language selection would lead to efforts for more language diversity and inclusion in NMT research. We built upon the work by  \citet{joshi2020state} in which $2485$ languages have been divided into 6 classes (see Table ~\ref{tab:language_categories}) based on the amount of publicly available un-annotated and annotated corpora. Although \citet{joshi2020state} did not specifically refer to parallel data available for NMT, we hypothesize that there exists a strong correlation between the language class (and consequently the amount of publicly available data for that language) and the amount of NMT research available for this language.

\subsection{Methodology}
 We queried Google Scholar with the query ``\textit{neural machine translation}'' + ``\textit{language}'' (e.g. ``\textit{neural machine translation}'' + ``\textit{Hindi}''). We excluded results before 2014 along with patents and citations. It should be noted that the search results were noisy; the most common among them being the ambiguity in the language name, where the language name is the same as the other entities such as location or author name. For example, the language `Swati', a Bantoid language spoken in Africa is also a common Indian name. Therefore, we manually checked and removed 240 such languages from our analysis.  

In order to find the LRLs that have been frequently used by the NMT community, we studied the outlier languages in language Classes 0-2 in \citet{joshi2020state}, using the obtained Google Scholar search results ($N_{GS}$). For each class $c$, the outlier languages were identified using the following equation:
\begin{equation}
 N_{GS}^l > Q_3^c + 1.5IQR^c
\label{eq:outliers}
\end{equation} 
where $N_{GS}^l$ represents the number of Google Scholar results obtained for a language $l$, $Q_3^c$ is the third quartile and $IQR^c$ is the interquartile range of Google Scholar results for a language class $c$. In order to ascertain the factors responsible for the interest of the researchers towards these languages, we manually selected some of these outliers with geographical variations and plotted the number of search results with respect to the year. 
These languages are selected such that it has a diverse mix of language class\footnote{Hausa and Swahili (from African macroarea) were not among the outlier languages, yet they were included in the plot to show geographical diversity.}.



\subsection{Results and Discussions}

\begin{figure}
    \begin{subfigure}
        \centering
        \includegraphics[width=0.45\textwidth, height=4.5cm]{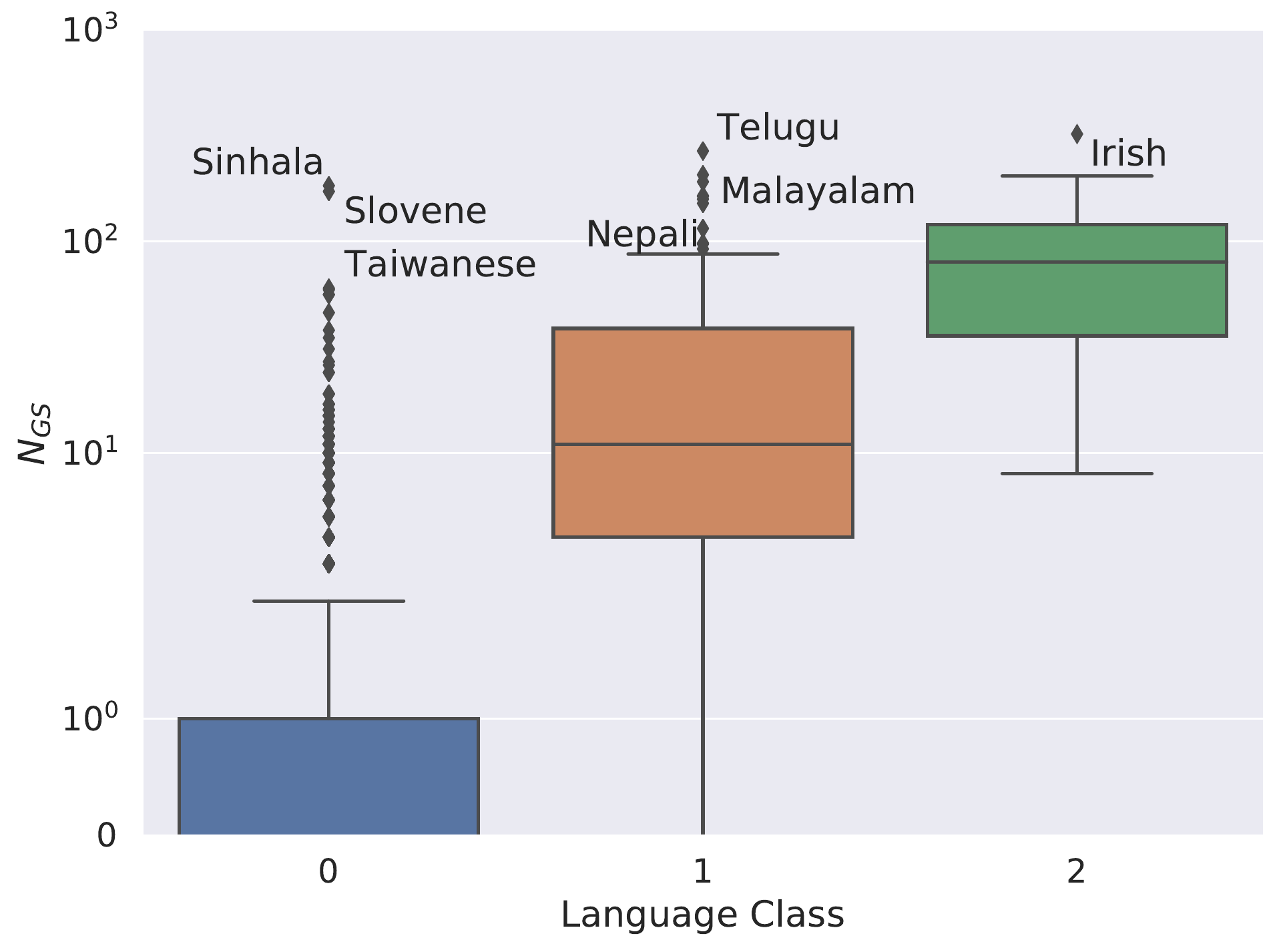}
    \end{subfigure}
    \begin{subfigure}
         \centering
        \includegraphics[width=0.45\textwidth, height=4.5cm]{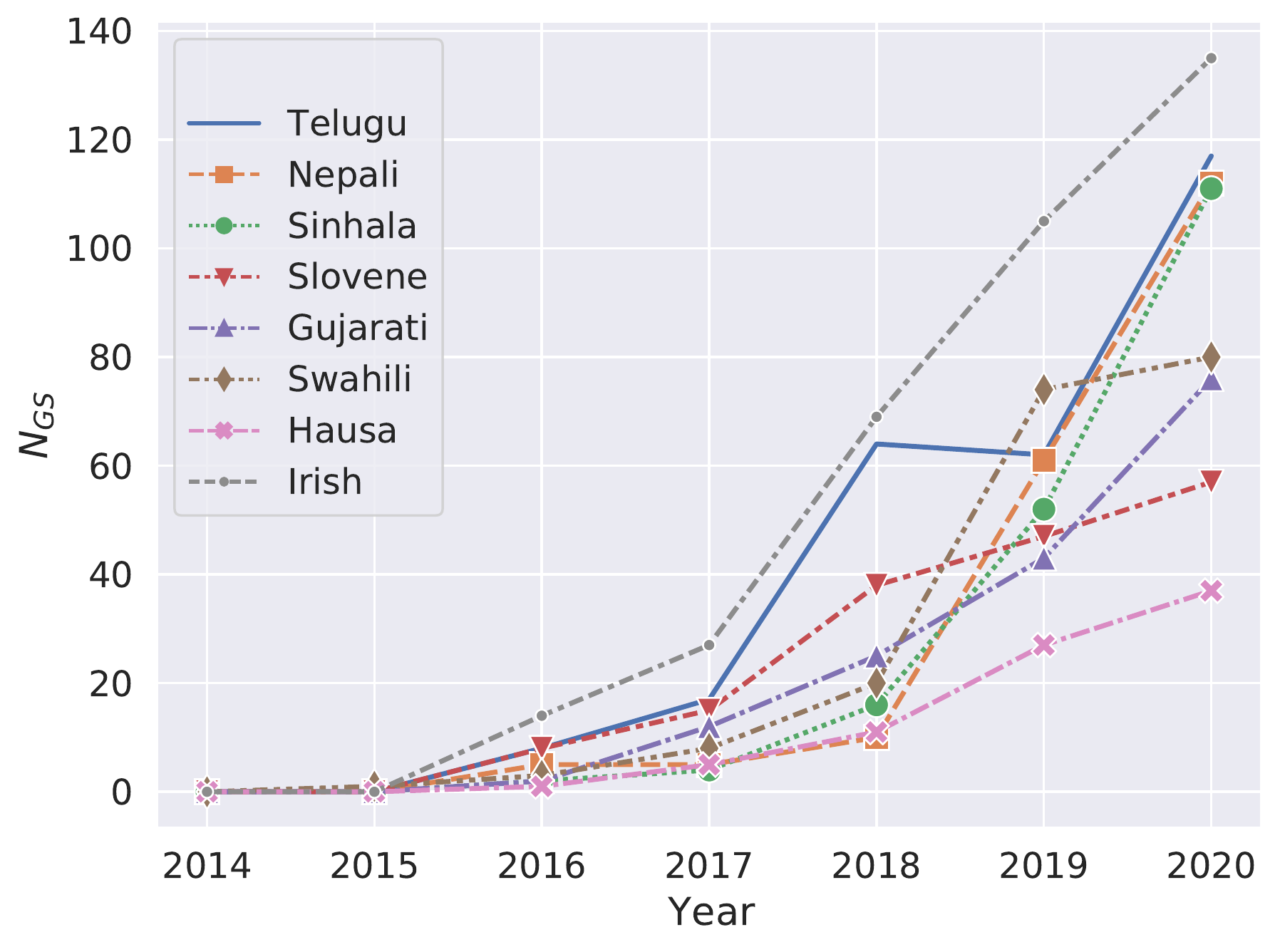}
    \end{subfigure}
    
    \caption{ a) Boxplot showing the distribution of Google Scholar search results ($N_{GS}$) for different language classes. Outliers are marked by black markers and have been calculated independently for each class b)Time-varying trends of a few selected outlier languages from Classes 0-2. Y-axis represents the year and the x-axis represents the number of Google Scholar search results($N_{GS}$) }
    \label{fig:boxplot_time}
\end{figure}

We found 12.6\%, 11.2\% and 7.1\% of languages as outliers for Class 0-2 respectively. A few random outlier examples are Sinhala \& Slovene (Class 0),  Nepali \&  Telugu (Class 1), and Irish (Class 2), as shown in Figure~\ref{fig:boxplot_time}(a). We identified the possible factors responsible for the growth of research for some languages and put them into four categories: geographic considerations, dataset availability, open source frameworks and models, and community involvement.

\begin{figure}
    \centering
        \begin{subfigure}
         \centering
           \includegraphics[width=0.45\textwidth, height=4.5cm]{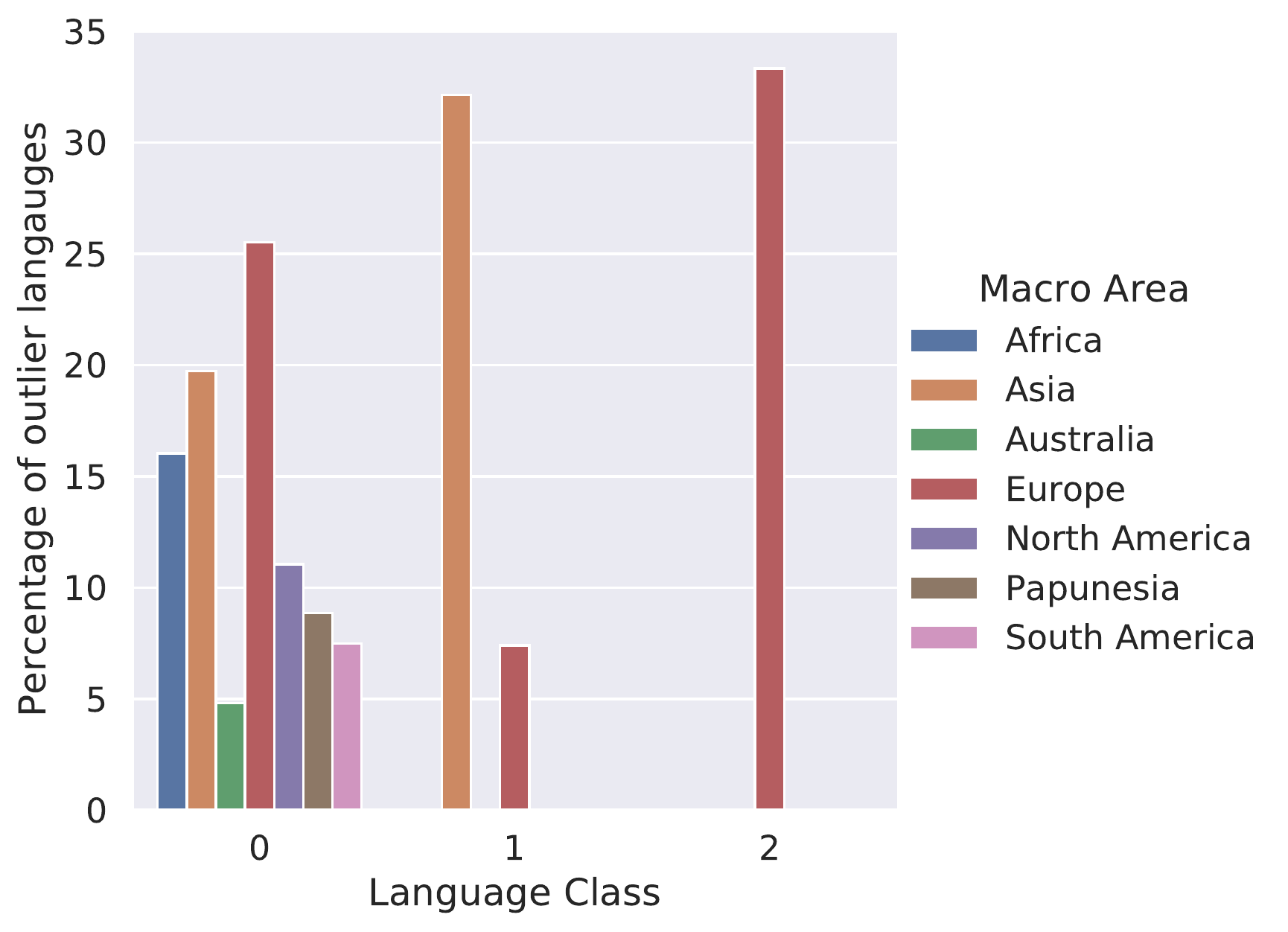}
        \end{subfigure}
    \begin{subfigure}{}
         \includegraphics[width=0.45\textwidth, height= 4.5cm ]{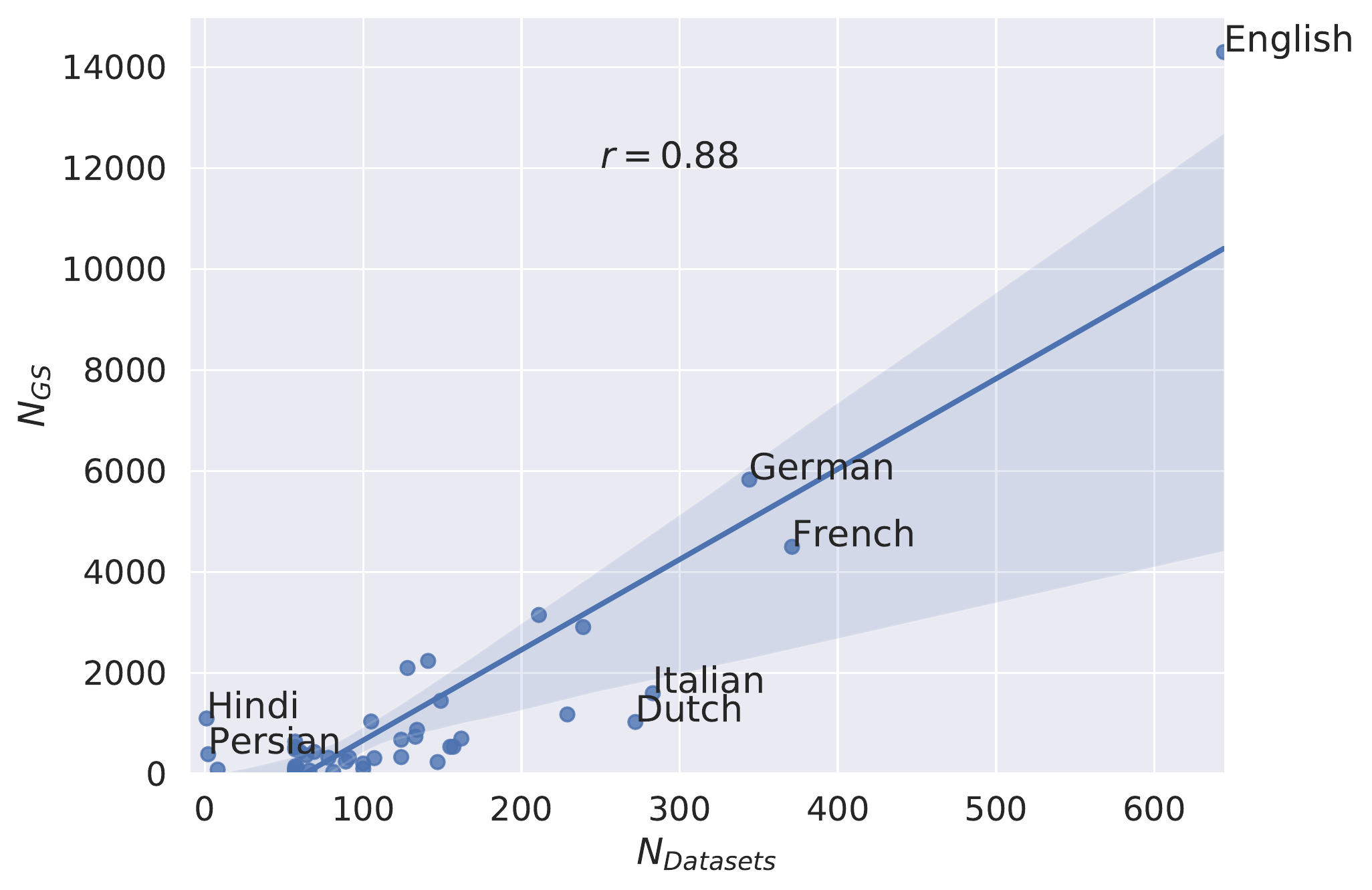}
        \end{subfigure}
    \caption{ a) Percentage of outliers from 7 different geographical regions for Classes 0-2 b) Relationship between the number of datasets available and number of Google Scholar search results ($N_{GS}$)}
        
\label{fig:geo_corr}
\end{figure}

\textbf{Geographic considerations }: We hypothesize that the geographical location where a language is spoken might play an important role in the growth of that language.
To validate the importance of geography, we looked at the outlier languages from Class 0-2 with respect to their geographical location\footnote{The geographical area of a language is determined by using WALS data~\cite{ dryer2011martin}}. In Figure~\ref{fig:geo_corr}(a), we plot the percentage of outlier vs its geographical region. We found that in Class 0, approximately 25\% of the outliers are languages from the European region, whereas in Class 1, approx 7\% of the total languages from Europe are outliers.

Thus it is safe to assume that the early growth for NMT was mostly driven by the geographical location of the language. This could be due to the availability of funds, resources, and regional level joint projects. One of the prominent examples is the growth of European languages. Some of the recent trends are also supporting this. For example, the steady increase in the research activity for Irish and Slovene, outliers from Class 2 and Class 0 respectively (see Figure ~\ref{fig:boxplot_time}) might be due to their presence in the European region.  



\textbf{Dataset availability}: The next source of growth is driven by the availability of the datasets. For example, Sinhala and Nepali, outlier languages from Class 0 and Class 1, respectively have seen a steep rise from 2018-19 onwards (see Fig~\ref{fig:boxplot_time}(b)). One reason could be due to the release of the FLoRes Evaluation Datasets ~\cite{guzman-etal-2019-flores}, which includes both Sinhala-English and Nepali-English.  Our analysis revealed that the inclusion of a language in the standard yearly challenge such as WMT has a considerable impact on its growth in terms of NMT. For example, WMT-2019~\cite{koehn-etal-2019-findings} shared task had Nepali-Hindi parallel corpus. Similarly, Nepali and Sinhala were also part of Google's 102 languages corpus~\cite{aharoni-etal-2019-massively}.
Similarly, the increase in the number of publications for the Gujarati language from 2019 onwards could be attributed to the fact that the Gujarati-English language pair was included in the Shared Machine Translation task in WMT 2019~\cite{barrault-etal-2019-findings}.

To quantify the relationship between the availability of datasets and research activity around that language, we used the resource matrix\footnote{More information on this source: http://matrix.statmt.org/resources/matrix?set=all}. 
This contains the details of the number of monolingual corpora and parallel corpora for 64 languages. Even though the list is not exhaustive, it is helpful for the growth analysis as it contains the languages from all the classes.  In Figure~\ref{fig:geo_corr}(b), we plot the total number of datasets available v.s. the research activity (number of Google Scholar results for NMT) for a particular language. The number of datasets (X-axis) has been calculated by summing the number of monolingual datasets available for a source language and parallel corpora available between a source language and the other target languages.  It can be observed that the availability of datasets is directly correlated with the research activity ($r=0.88$), which further strengthens our claim that the NMT growth for a particular language is directly proportional to the data availability.

\textbf{Open-source frameworks and models accessibility}
The availability of open-source frameworks and models is a major contributing factor towards the growth of research in the area of NMT. Frameworks such as OpenNMT~\citep{klein-etal-2017-opennmt}, and fairseq \cite{ott-etal-2019-fairseq}, as well as pre-trained models such as mBART \cite{liu2020multilingual} provide an easy and scalable implementation that helps in building a baseline and improving it for existing and new languages. These open-source projects are periodically maintained, flexible, and provide most of the latest NMT-related techniques. Since these projects provide standardized codes, it becomes easy to adapt for the LRLs even by novice researchers. It eliminates the need to develop the codes from scratch and helps in accelerating the research process.

\textbf{Community involvement}:  A recent development is a group of like-minded researchers coming together to increase the visibility of MT systems in the context of languages used in a particular region. It consists of both dataset building and the development of the standardized code and also focuses on training a new generation of enthusiasts to carry forward the work. One of the prominent examples is the Masakhane project~\citep{nekoto2020participatory}, which aims to put the Africa AI, specifically African language MT, into the world map. Within about two years, the Masakhane community has covered more than 38 African languages and resulted in multiple publications~\cite{nekoto2020participatory}. 
As we could see from Figure~\ref{fig:boxplot_time}(b), two of the representative languages, Swahili and Hausa, have a steep growth after 2018, which coincides with the inception of the Masakhane project.

Our results and analysis highlight i) the importance of community building and region-level projects, ii) the inclusion of LRL datasets into yearly challenges and large multilingual datasets, and iii) the availability of open source models and frameworks to increase the focus on LRLs in the NMT landscape. This analysis could provide a cue to the researchers and funding agencies worldwide for the development of LRL resources.

\section{Discussion}
This section discusses the open questions in LRL-NMT research and provides the answers to our initial research questions.

\subsection{Open Questions for LRL-NMT and Future Directions}
While notable advances have been made in LRL-NMT in recent years, there remain interesting directions for exploration in model improvements as well as equitable and inclusive access.

\subsubsection{Model Improvements}
Based on the various LRL-NMT techniques discussed, there are multiple improvements that can be applied to the model:  allowing the multilingual models to include more LRL pairs, making the models more robust to limitations in the input dataset, expanding the interpretability and explainability of the model in understanding their behaviour on  LRL pairs, and mitigating biases in the model.

\paragraph{Model Capacity to Include LRLs} 
Massive multi-NMT models is a promising technique, especially for systems produced by multinational tech companies ~\cite{arivazhagan2019massively, aharoni-etal-2019-massively, fan2020englishcentric}. In particular, multi-NMT for zero-shot translation is an important line of research, as it eliminates the need for parallel datasets between every possible language. The work of~\citet{fan2020englishcentric} is of particular interest, as it deals with a non-English-centric multilingual dataset, yet managed to outperform English-centric models in zero-shot translation.
However, despite multi-NMT being able to cover about 100 languages ~\cite{arivazhagan2019massively, aharoni-etal-2019-massively}, only a small fraction of LRLs are included from the more than 7000 possible languages. Therefore, more efforts need to be invested in scaling multi-NMT models that are capable of handling a larger number of languages, which would inherently cover LRLs and extremely LRLs. It is important to investigate how these LRLs are better represented in these massive models without compromising the performance of high-resource languages.~\citet{fan2020englishcentric} recently reported promising results on this line, which should be further explored.

\paragraph{Model Robustness to Limiting Input Factors:} 
Techniques discussed in Section~\ref{section:label-NMT-Techniques-for-Low-Resource-Languages} are limited by various input requirements such as the size of the datasets, domain of the datasets, and language relatedness. Although some ablation studies~\cite{arivazhagan2019massively, aji2020neural, edunov-etal-2018-understanding} have been done to understand the effect of input requirements, they are not sufficient or not exhaustive in considering model robustness. 

For the example of language similarity, LRLs that have syntactic differences between spoken and written forms (e.g. Sinhala) is a challenge. Another challenge is translating code-mixed data.
In terms of domain relatedness, most publicly available datasets are from sources that do not resemble real-life translations. Thus, more focus should be given to multilingual domain adaption research, where pre-trained multi-MNT models can be fine-tuned to the task at hand (e.g.~translating legal or medical records). 
Thus, empirical experimentation with the goal to extend the field to new approaches and architectures to address these limitations must be conducted.

\paragraph{Model Interpretability and Explainability:}
Explaining and interpreting neural models is a challenge for researchers in Deep Learning.  
For multi-NMT, there have been experiments \cite{johnson2017google,kudugunta-etal-2019-investigating} that delve into models to examine how the learning takes place, and how different languages behave. There have also been attempts to provide theoretical interpretations on aspects of NMT models~\cite{zhao2020learning}. However, we believe more research in NMT model interpretability (such as interlingua learning and parent-child transfer) would help developing solutions specific to LRLs.

\paragraph{Mitigating Model Bias:}It has already been shown that gender bias exists in NMT models when trained on high-resource languages~\cite{stanovsky2019evaluating, DBLP:journals/corr/abs-2010-02428}. Recently, the existence of biases in the context of LRL MT is a problem that has not come into light as far as we know. In particular, when translating between languages belonging to different regions and cultures in multilingual settings, there can be undesirable influences from high-resource languages. As discussed later in this section, parallel data extracted from the web tends to contain bias. While there should be parallel efforts to free datasets of bias, it is important to develop models that are robust to dataset bias in order to prevent the ramification of propagating stereotypes from the social context of high-resource languages in to LRLs.

\subsubsection{Equitable and Inclusive Access}
Findings in our trend analysis (Section~\ref{techniques_trend}) suggest that more resources should be made available to underrepresented geographic regions, especially for communities that are traditionally excluded in technological development and those who face social-economic inequities.  Therefore the inclusion of these communities can be prioritized when creating datasets and when providing  accessibility to substantial computing resources required for building time-consuming and expensive neural models. These communities will also benefit from open-source tools and frameworks, as well as the availability of trained models.


\paragraph{Creation  of Datasets:} 
Most of the datasets used in NMT have originated in a small number of regions in the world focusing on English-centric translations, however about 40\% the content on the Internet is now non-English\footnote{\url{https://www.visualcapitalist.com/the-most-used-languages-on-the-internet/}}.
Although LRL-NMT was initially applied to large corpora by sub-sampling HRLs such as English, French, and German, \cite{artetxe-etal-2017-learning, artetxe-etal-2019-effective}
it was later adapted for LRL pairs such as English-Esperanto, \cite{conneau2017word},
English-Urdu, English-Romanian, and English-Russian \cite{lample-etal-2018-phrase, lample2019cross}\footnote{\cite{lample2018unsupervised} excluded because low-resource was suggested but not experimented}.
This focus shifted to European LRLs, and more recently to non-European languages such as Indian, African, East Asian, and Middle Eastern languages ~\cite{barrault-etal-2019-findings, nekoto2020participatory, guzman-etal-2019-flores, koehn-etal-2019-findings}.
However, the amount of such datasets is less than high resource counterparts. 

Therefore, on par with current trends in some Machine Learning communities, more focus can be given to those that present new LRL datasets rather than the novelty of the employed technique, when accepting papers to conferences and evaluating value of this type of research. Contribution of conferences such as LREC, and journals such as the LREC journal is commendable, in this regard.  Projects such as ParaCrawl~\cite{banon2020paracrawl} have automatically mined a large amount of parallel data from the web for multiple language pairs, including LRL such as Irish and Nepalese.  
In addition, regional communities can take the lead in dataset creation due to their expertise in their own cultural context thus could provide better judgement on the bias (discussed in the next point). 

More recently, the problem of dataset bias has received significant attention from the community \cite{blodgett2020language}.  For example, the public crawl data available demonstrate a narrow sub-segment of population and may encode values and perspectives that exist within Western society.  More specifically, it has been shown that scrapped text contains geographical bias~\cite{graham2014uneven, mcguffie2020radicalization}, as well as an age and gender bias~\cite{cohen2008encyclopedia}. 
Furthermore, these web crawl datasets contain the potential to harm due to abusive language, hate speech, microaggressions, dehumanization, and social-political biases~\cite{auxier2021social}. Thus, data pre-processing mechanisms can be employed with input from experts such as social scientists, regional communities, and linguists familiar with different languages. 


\paragraph{Open-source Tools and Frameworks}
Same as creating new LRL datasets, accessibility to tools and frameworks for LRLs are critical in advancing the field. Therefore, creating and making them open-source for free access is of tremendous benefit to LRL community. We like to note the positive impact created by open source initiatives such as HuggingFace\footnote{\url{https://huggingface.co/}}.  

\paragraph{Availability of Trained models:}
Although the community has a recent focus on developing computationally efficient NMT models~\cite{birch-etal-2018-findings}, the public release of large-scale multi-NMT models has been limited, with the exception of the work by~\cite{liu2020multilingual}.  By giving public access to these models, these time-consuming and expensive models can be used as the parent models to be transferred to the LRL child models.  
Therefore publicly releasing NMT models including massive multi-NMT models would be tremendously beneficial to those working in LRL-NMT as well as advancing the field in other areas.



\paragraph{Availability of Computational Resources:} 
The community has a recent focus on developing computationally efficient NMT models and providing computational resources for researchers \cite{birch-etal-2018-findings}\footnote{Some tech giants also provide research grants to use their cloud GPUs platforms.}. However, more efforts need to be put forth by the research community, industry organizations, and governments in distributing resources and attention. Furthermore, computational resources can also be made available as part of conferences and challenges to further encourage LRLs researchers to participate.

\subsection{Answering the Research Questions}
Key findings of this survey can be summarized as follows:  
\begin{enumerate}
    \item \textbf{Techniques and Trends:}  Our survey found that there is a substantial amount of LRL-NMT research, and the trend continues. All the LRL-NMT techniques (data augmentation, unsupervised NMT, semi-supervised NMT, multi-NMT, and transfer learning for NMT) except pivoting show an upward trend with respect to the research publications, suggesting that these techniques have established themselves as the de facto solutions for LRL-NMT.
    \item \textbf{Technique Selection:} The decision chart we produced on technique selection can be taken as a guide in selecting the most appropriate NMT technique for a given data specification, also considering the availability of computational resources. However, we note that this selection also depends on other factors such as language relatedness, and the domain of data. These factors were discussed with respect to individual LRL-NMT techniques.   
    \item \textbf{Future Directions:} We identified multiple research areas for the research community to collectively increase efforts on LRL-NMT. These initiatives were broadly categorised as model improvements and equitable and inclusive access.
\end{enumerate}
\section{Conclusions} \label{label-Conclusion}
Due to the recent advancements in the field, NMT is no longer an unattainable goal for LRLs. However, the sheer volume as well as the acceleration of research taking place makes it difficult to select the state-of-the-art LRL-NMT techniques for a given data specification, as no guidelines are available for the selection of the most appropriate NMT technique for a given data setup. The contribution of this survey paper is to give a comprehensive picture of the LRL-NMT landscape, highlighting the recent trends in technological advancements and provide a guideline to select the most appropriate LRL-NMT technique for a given data specification. Based on our findings through research publications and our quantitative analysis, we provided a set of recommendations to advance the LRL-NMT solutions. We believe that these recommendations would be positively received by the NMT research community.
\section{Acknowledgement}
The first author (SR) would like to thank the postgraduate students of the National Language Processing Center, University of Moratuwa, and the undergraduates of the Department of Computer  Science and Engineering, University of Moratuwa, for their support.

	\bibliographystyle{ACM-Reference-Format}
	\bibliography{lowMT.bib}


\begin{thebibliography}{196}


\ifx \showCODEN    \undefined \def \showCODEN     #1{\unskip}     \fi
\ifx \showDOI      \undefined \def \showDOI       #1{#1}\fi
\ifx \showISBNx    \undefined \def \showISBNx     #1{\unskip}     \fi
\ifx \showISBNxiii \undefined \def \showISBNxiii  #1{\unskip}     \fi
\ifx \showISSN     \undefined \def \showISSN      #1{\unskip}     \fi
\ifx \showLCCN     \undefined \def \showLCCN      #1{\unskip}     \fi
\ifx \shownote     \undefined \def \shownote      #1{#1}          \fi
\ifx \showarticletitle \undefined \def \showarticletitle #1{#1}   \fi
\ifx \showURL      \undefined \def \showURL       {\relax}        \fi
\providecommand\bibfield[2]{#2}
\providecommand\bibinfo[2]{#2}
\providecommand\natexlab[1]{#1}
\providecommand\showeprint[2][]{arXiv:#2}

\bibitem[\protect\citeauthoryear{Abdou, Glon{\v{c}}{\'a}k, and Bojar}{Abdou
  et~al\mbox{.}}{2017}]%
        {abdou2017variable}
\bibfield{author}{\bibinfo{person}{Mostafa Abdou}, \bibinfo{person}{Vladan
  Glon{\v{c}}{\'a}k}, {and} \bibinfo{person}{Ond{\v{r}}ej Bojar}.}
  \bibinfo{year}{2017}\natexlab{}.
\newblock \showarticletitle{Variable mini-batch sizing and pre-trained
  embeddings}. In \bibinfo{booktitle}{\emph{Proceedings of the Second
  Conference on Machine Translation}}. \bibinfo{pages}{680--686}.
\newblock


\bibitem[\protect\citeauthoryear{A{\c{c}}ar{\c{c}}i{\c{c}}ek,
  {\c{C}}olako{\u{g}}lu, Aktan~Hatipo{\u{g}}lu, Huang, and
  Peng}{A{\c{c}}ar{\c{c}}i{\c{c}}ek et~al\mbox{.}}{2020}]%
        {acarcicek2020filtering}
\bibfield{author}{\bibinfo{person}{Haluk A{\c{c}}ar{\c{c}}i{\c{c}}ek},
  \bibinfo{person}{Talha {\c{C}}olako{\u{g}}lu}, \bibinfo{person}{P{\i}nar~Ece
  Aktan~Hatipo{\u{g}}lu}, \bibinfo{person}{Chong~Hsuan Huang}, {and}
  \bibinfo{person}{Wei Peng}.} \bibinfo{year}{2020}\natexlab{}.
\newblock \showarticletitle{Filtering Noisy Parallel Corpus using Transformers
  with Proxy Task Learning}. In \bibinfo{booktitle}{\emph{Proceedings of the
  Fifth Conference on Machine Translation}}. \bibinfo{pages}{940--946}.
\newblock


\bibitem[\protect\citeauthoryear{Aharoni, Johnson, and Firat}{Aharoni
  et~al\mbox{.}}{2019}]%
        {aharoni-etal-2019-massively}
\bibfield{author}{\bibinfo{person}{Roee Aharoni}, \bibinfo{person}{Melvin
  Johnson}, {and} \bibinfo{person}{Orhan Firat}.}
  \bibinfo{year}{2019}\natexlab{}.
\newblock \showarticletitle{Massively Multilingual Neural Machine Translation}.
  In \bibinfo{booktitle}{\emph{Proceedings of the 2019 Conference of the North
  {A}merican Chapter of the Association for Computational Linguistics: Human
  Language Technologies, Volume 1 (Long and Short Papers)}}.
  \bibinfo{pages}{3874--3884}.
\newblock


\bibitem[\protect\citeauthoryear{Aji, Bogoychev, Heafield, and Sennrich}{Aji
  et~al\mbox{.}}{2020}]%
        {aji2020neural}
\bibfield{author}{\bibinfo{person}{Alham~Fikri Aji}, \bibinfo{person}{Nikolay
  Bogoychev}, \bibinfo{person}{Kenneth Heafield}, {and} \bibinfo{person}{Rico
  Sennrich}.} \bibinfo{year}{2020}\natexlab{}.
\newblock \showarticletitle{In Neural Machine Translation, What Does Transfer
  Learning Transfer?}. In \bibinfo{booktitle}{\emph{Proceedings of the 58th
  Annual Meeting of the ACL}}. \bibinfo{pages}{7701--7710}.
\newblock


\bibitem[\protect\citeauthoryear{Alsohybe, Dahan, and Ba-Alwi}{Alsohybe
  et~al\mbox{.}}{2017}]%
        {alsohybe2017machine}
\bibfield{author}{\bibinfo{person}{Nabeel~T Alsohybe},
  \bibinfo{person}{Neama~Abdulaziz Dahan}, {and} \bibinfo{person}{Fadl~Mutaher
  Ba-Alwi}.} \bibinfo{year}{2017}\natexlab{}.
\newblock \showarticletitle{Machine-translation history and evolution: survey
  for Arabic-English translations}.
\newblock \bibinfo{journal}{\emph{arXiv preprint arXiv:1709.04685}}
  (\bibinfo{year}{2017}).
\newblock


\bibitem[\protect\citeauthoryear{Arivazhagan, Bapna, Firat, Lepikhin, Johnson,
  Krikun, Chen, Cao, Foster, Cherry, et~al\mbox{.}}{Arivazhagan
  et~al\mbox{.}}{2019}]%
        {arivazhagan2019massively}
\bibfield{author}{\bibinfo{person}{Naveen Arivazhagan}, \bibinfo{person}{Ankur
  Bapna}, \bibinfo{person}{Orhan Firat}, \bibinfo{person}{Dmitry Lepikhin},
  \bibinfo{person}{Melvin Johnson}, \bibinfo{person}{Maxim Krikun},
  \bibinfo{person}{Mia~Xu Chen}, \bibinfo{person}{Yuan Cao},
  \bibinfo{person}{George Foster}, \bibinfo{person}{Colin Cherry},
  {et~al\mbox{.}}} \bibinfo{year}{2019}\natexlab{}.
\newblock \showarticletitle{Massively multilingual neural machine translation
  in the wild: Findings and challenges}.
\newblock \bibinfo{journal}{\emph{arXiv preprint arXiv:1907.05019}}
  (\bibinfo{year}{2019}).
\newblock


\bibitem[\protect\citeauthoryear{Artetxe, Labaka, and Agirre}{Artetxe
  et~al\mbox{.}}{2017}]%
        {artetxe-etal-2017-learning}
\bibfield{author}{\bibinfo{person}{Mikel Artetxe}, \bibinfo{person}{Gorka
  Labaka}, {and} \bibinfo{person}{Eneko Agirre}.}
  \bibinfo{year}{2017}\natexlab{}.
\newblock \showarticletitle{Learning bilingual word embeddings with (almost) no
  bilingual data}. In \bibinfo{booktitle}{\emph{Proceedings of the 55th Annual
  Meeting of the Association for Computational Linguistics}}.
  \bibinfo{pages}{451--462}.
\newblock


\bibitem[\protect\citeauthoryear{Artetxe, Labaka, and Agirre}{Artetxe
  et~al\mbox{.}}{2019}]%
        {artetxe-etal-2019-effective}
\bibfield{author}{\bibinfo{person}{Mikel Artetxe}, \bibinfo{person}{Gorka
  Labaka}, {and} \bibinfo{person}{Eneko Agirre}.}
  \bibinfo{year}{2019}\natexlab{}.
\newblock \showarticletitle{An Effective Approach to Unsupervised Machine
  Translation}. In \bibinfo{booktitle}{\emph{Proceedings of the 57th Annual
  Meeting of the Association for Computational Linguistics}}.
  \bibinfo{pages}{194--203}.
\newblock


\bibitem[\protect\citeauthoryear{Artetxe, Labaka, Agirre, and Cho}{Artetxe
  et~al\mbox{.}}{2018}]%
        {artetxe2018unsupervised}
\bibfield{author}{\bibinfo{person}{Mikel Artetxe}, \bibinfo{person}{Gorka
  Labaka}, \bibinfo{person}{Eneko Agirre}, {and} \bibinfo{person}{Kyunghyun
  Cho}.} \bibinfo{year}{2018}\natexlab{}.
\newblock \showarticletitle{Unsupervised Neural Machine Translation}. In
  \bibinfo{booktitle}{\emph{6th International Conference on Learning
  Representations, {ICLR} 2018, Conference Track Proceedings}}.
\newblock


\bibitem[\protect\citeauthoryear{Artetxe, Labaka, Casas, and Agirre}{Artetxe
  et~al\mbox{.}}{2020}]%
        {artetxe2020all}
\bibfield{author}{\bibinfo{person}{Mikel Artetxe}, \bibinfo{person}{Gorka
  Labaka}, \bibinfo{person}{Noe Casas}, {and} \bibinfo{person}{Eneko Agirre}.}
  \bibinfo{year}{2020}\natexlab{}.
\newblock \showarticletitle{Do all roads lead to Rome? Understanding the role
  of initialization in iterative back-translation}.
\newblock \bibinfo{journal}{\emph{Knowledge-Based Systems}}
  \bibinfo{volume}{206} (\bibinfo{year}{2020}), \bibinfo{pages}{106401}.
\newblock


\bibitem[\protect\citeauthoryear{Artetxe and Schwenk}{Artetxe and
  Schwenk}{2019}]%
        {artetxe-schwenk-2019-margin}
\bibfield{author}{\bibinfo{person}{Mikel Artetxe} {and} \bibinfo{person}{Holger
  Schwenk}.} \bibinfo{year}{2019}\natexlab{}.
\newblock \showarticletitle{Margin-based Parallel Corpus Mining with
  Multilingual Sentence Embeddings}. In \bibinfo{booktitle}{\emph{Proceedings
  of the 57th Annual Meeting of the Association for Computational
  Linguistics}}. \bibinfo{pages}{3197--3203}.
\newblock


\bibitem[\protect\citeauthoryear{Auxier and Anderson}{Auxier and
  Anderson}{2021}]%
        {auxier2021social}
\bibfield{author}{\bibinfo{person}{Brooke Auxier} {and} \bibinfo{person}{Monica
  Anderson}.} \bibinfo{year}{2021}\natexlab{}.
\newblock \showarticletitle{Social media use in 2021}.
\newblock \bibinfo{journal}{\emph{Pew Research Center}} (\bibinfo{year}{2021}).
\newblock


\bibitem[\protect\citeauthoryear{Bahdanau, Cho, and Bengio}{Bahdanau
  et~al\mbox{.}}{2015}]%
        {38ed090f8de94fb3b0b46b86f9133623}
\bibfield{author}{\bibinfo{person}{Dzmitry Bahdanau},
  \bibinfo{person}{Kyunghyun Cho}, {and} \bibinfo{person}{Yoshua Bengio}.}
  \bibinfo{year}{2015}\natexlab{}.
\newblock \showarticletitle{Neural Machine Translation by Jointly Learning to
  Align and Translate}. In \bibinfo{booktitle}{\emph{3rd International
  Conference on Learning Representations, {ICLR} 2015}}.
\newblock


\bibitem[\protect\citeauthoryear{Ba{\~n}{\'o}n, Chen, Haddow, Heafield, Hoang,
  Espl{\`a}-Gomis, Forcada, Kamran, Kirefu, Koehn, Ortiz~Rojas, Pla~Sempere,
  Ram{\'\i}rez-S{\'a}nchez, Sarr{\'\i}as, Strelec, Thompson, Waites, Wiggins,
  and Zaragoza}{Ba{\~n}{\'o}n et~al\mbox{.}}{2020}]%
        {banon2020paracrawl}
\bibfield{author}{\bibinfo{person}{Marta Ba{\~n}{\'o}n},
  \bibinfo{person}{Pinzhen Chen}, \bibinfo{person}{Barry Haddow},
  \bibinfo{person}{Kenneth Heafield}, \bibinfo{person}{Hieu Hoang},
  \bibinfo{person}{Miquel Espl{\`a}-Gomis}, \bibinfo{person}{Mikel~L. Forcada},
  \bibinfo{person}{Amir Kamran}, \bibinfo{person}{Faheem Kirefu},
  \bibinfo{person}{Philipp Koehn}, \bibinfo{person}{Sergio Ortiz~Rojas},
  \bibinfo{person}{Leopoldo Pla~Sempere}, \bibinfo{person}{Gema
  Ram{\'\i}rez-S{\'a}nchez}, \bibinfo{person}{Elsa Sarr{\'\i}as},
  \bibinfo{person}{Marek Strelec}, \bibinfo{person}{Brian Thompson},
  \bibinfo{person}{William Waites}, \bibinfo{person}{Dion Wiggins}, {and}
  \bibinfo{person}{Jaume Zaragoza}.} \bibinfo{year}{2020}\natexlab{}.
\newblock \showarticletitle{{P}ara{C}rawl: Web-Scale Acquisition of Parallel
  Corpora}. In \bibinfo{booktitle}{\emph{Proceedings of the 58th Annual Meeting
  of the ACL}}. \bibinfo{pages}{4555--4567}.
\newblock


\bibitem[\protect\citeauthoryear{Baziotis, Haddow, and Birch}{Baziotis
  et~al\mbox{.}}{2020}]%
        {baziotis2020language}
\bibfield{author}{\bibinfo{person}{Christos Baziotis}, \bibinfo{person}{Barry
  Haddow}, {and} \bibinfo{person}{Alexandra Birch}.}
  \bibinfo{year}{2020}\natexlab{}.
\newblock \showarticletitle{Language Model Prior for Low-Resource Neural
  Machine Translation}. In \bibinfo{booktitle}{\emph{Proceedings of the 2020
  Conference on Empirical Methods in NLP}}. \bibinfo{pages}{7622--7634}.
\newblock


\bibitem[\protect\citeauthoryear{Besacier, Barnard, Karpov, and
  Schultz}{Besacier et~al\mbox{.}}{2014}]%
        {besacier2014automatic}
\bibfield{author}{\bibinfo{person}{Laurent Besacier}, \bibinfo{person}{Etienne
  Barnard}, \bibinfo{person}{Alexey Karpov}, {and} \bibinfo{person}{Tanja
  Schultz}.} \bibinfo{year}{2014}\natexlab{}.
\newblock \showarticletitle{Automatic speech recognition for under-resourced
  languages: A survey}.
\newblock \bibinfo{journal}{\emph{Speech Communication}}  \bibinfo{volume}{56}
  (\bibinfo{year}{2014}), \bibinfo{pages}{85--100}.
\newblock


\bibitem[\protect\citeauthoryear{Birch, Finch, Luong, Neubig, and Oda}{Birch
  et~al\mbox{.}}{2018}]%
        {birch-etal-2018-findings}
\bibfield{author}{\bibinfo{person}{Alexandra Birch}, \bibinfo{person}{Andrew
  Finch}, \bibinfo{person}{Minh-Thang Luong}, \bibinfo{person}{Graham Neubig},
  {and} \bibinfo{person}{Yusuke Oda}.} \bibinfo{year}{2018}\natexlab{}.
\newblock \showarticletitle{Findings of the Second Workshop on Neural Machine
  Translation and Generation}. In \bibinfo{booktitle}{\emph{Proceedings of the
  2nd Workshop on Neural Machine Translation and Generation}}.
  \bibinfo{pages}{1--10}.
\newblock


\bibitem[\protect\citeauthoryear{Blackwood, Ballesteros, and Ward}{Blackwood
  et~al\mbox{.}}{2018}]%
        {blackwood-etal-2018-multilingual}
\bibfield{author}{\bibinfo{person}{Graeme Blackwood}, \bibinfo{person}{Miguel
  Ballesteros}, {and} \bibinfo{person}{Todd Ward}.}
  \bibinfo{year}{2018}\natexlab{}.
\newblock \showarticletitle{Multilingual Neural Machine Translation with
  Task-Specific Attention}. In \bibinfo{booktitle}{\emph{Proceedings of the
  27th International Conference on Computational Linguistics}}.
  \bibinfo{pages}{3112--3122}.
\newblock


\bibitem[\protect\citeauthoryear{Blodgett, Barocas, Daum{\'e}~III, and
  Wallach}{Blodgett et~al\mbox{.}}{2020}]%
        {blodgett2020language}
\bibfield{author}{\bibinfo{person}{Su~Lin Blodgett}, \bibinfo{person}{Solon
  Barocas}, \bibinfo{person}{Hal Daum{\'e}~III}, {and} \bibinfo{person}{Hanna
  Wallach}.} \bibinfo{year}{2020}\natexlab{}.
\newblock \showarticletitle{Language (Technology) is Power: A Critical Survey
  of {``}Bias{''} in {NLP}}. In \bibinfo{booktitle}{\emph{Proceedings of the
  58th Annual Meeting of the ACL}}. \bibinfo{pages}{5454--5476}.
\newblock


\bibitem[\protect\citeauthoryear{Bojar, Chatterjee, Federmann, Graham, Haddow,
  Huang, Huck, Koehn, Liu, Logacheva, Monz, Negri, Post, Rubino, Specia, and
  Turchi}{Bojar et~al\mbox{.}}{2017}]%
        {barrault-etal-2019-findings}
\bibfield{author}{\bibinfo{person}{Ond{\v{r}}ej Bojar}, \bibinfo{person}{Rajen
  Chatterjee}, \bibinfo{person}{Christian Federmann}, \bibinfo{person}{Yvette
  Graham}, \bibinfo{person}{Barry Haddow}, \bibinfo{person}{Shujian Huang},
  \bibinfo{person}{Matthias Huck}, \bibinfo{person}{Philipp Koehn},
  \bibinfo{person}{Qun Liu}, \bibinfo{person}{Varvara Logacheva},
  \bibinfo{person}{Christof Monz}, \bibinfo{person}{Matteo Negri},
  \bibinfo{person}{Matt Post}, \bibinfo{person}{Raphael Rubino},
  \bibinfo{person}{Lucia Specia}, {and} \bibinfo{person}{Marco Turchi}.}
  \bibinfo{year}{2017}\natexlab{}.
\newblock \showarticletitle{Findings of the 2017 Conference on Machine
  Translation ({WMT}17)}. In \bibinfo{booktitle}{\emph{Proceedings of the
  Second Conference on Machine Translation}}. \bibinfo{pages}{169--214}.
\newblock


\bibitem[\protect\citeauthoryear{Caswell, Chelba, and Grangier}{Caswell
  et~al\mbox{.}}{2019}]%
        {caswell2019tagged}
\bibfield{author}{\bibinfo{person}{Isaac Caswell}, \bibinfo{person}{Ciprian
  Chelba}, {and} \bibinfo{person}{David Grangier}.}
  \bibinfo{year}{2019}\natexlab{}.
\newblock \showarticletitle{Tagged Back-Translation}. In
  \bibinfo{booktitle}{\emph{Proceedings of the Fourth Conference on Machine
  Translation (Volume 1: Research Papers)}}. \bibinfo{pages}{53--63}.
\newblock


\bibitem[\protect\citeauthoryear{Chen, Liu, Cheng, and Li}{Chen
  et~al\mbox{.}}{2017}]%
        {chen2017teacher}
\bibfield{author}{\bibinfo{person}{Yun Chen}, \bibinfo{person}{Yang Liu},
  \bibinfo{person}{Yong Cheng}, {and} \bibinfo{person}{Victor~O.K. Li}.}
  \bibinfo{year}{2017}\natexlab{}.
\newblock \showarticletitle{A Teacher-Student Framework for Zero-Resource
  Neural Machine Translation}. In \bibinfo{booktitle}{\emph{Proceedings of the
  55th Annual Meeting of the ACL}}. \bibinfo{pages}{1925--1935}.
\newblock


\bibitem[\protect\citeauthoryear{Cheng, Xu, He, He, Wu, Sun, and Liu}{Cheng
  et~al\mbox{.}}{2016}]%
        {cheng2016semi}
\bibfield{author}{\bibinfo{person}{Yong Cheng}, \bibinfo{person}{Wei Xu},
  \bibinfo{person}{Zhongjun He}, \bibinfo{person}{Wei He}, \bibinfo{person}{Hua
  Wu}, \bibinfo{person}{Maosong Sun}, {and} \bibinfo{person}{Yang Liu}.}
  \bibinfo{year}{2016}\natexlab{}.
\newblock \showarticletitle{Semi-Supervised Learning for Neural Machine
  Translation}. In \bibinfo{booktitle}{\emph{Proceedings of the 54th Annual
  Meeting of the Association for Computational Linguistics (Volume 1: Long
  Papers)}}. \bibinfo{pages}{1965--1974}.
\newblock


\bibitem[\protect\citeauthoryear{Cheng, Yang, Liu, Sun, and Xu}{Cheng
  et~al\mbox{.}}{2017}]%
        {cheng2017joint}
\bibfield{author}{\bibinfo{person}{Yong Cheng}, \bibinfo{person}{Qian Yang},
  \bibinfo{person}{Yang Liu}, \bibinfo{person}{Maosong Sun}, {and}
  \bibinfo{person}{Wei Xu}.} \bibinfo{year}{2017}\natexlab{}.
\newblock \showarticletitle{Joint Training for Pivot-based Neural Machine
  Translation}. In \bibinfo{booktitle}{\emph{Proceedings of the Twenty-Sixth
  International Joint Conference on AI 2017}}. \bibinfo{pages}{3974--3980}.
\newblock


\bibitem[\protect\citeauthoryear{Cho, van Merri{\"e}nboer, Bahdanau, and
  Bengio}{Cho et~al\mbox{.}}{2014}]%
        {cho-etal-2014-properties}
\bibfield{author}{\bibinfo{person}{Kyunghyun Cho}, \bibinfo{person}{Bart van
  Merri{\"e}nboer}, \bibinfo{person}{Dzmitry Bahdanau}, {and}
  \bibinfo{person}{Yoshua Bengio}.} \bibinfo{year}{2014}\natexlab{}.
\newblock \showarticletitle{On the Properties of Neural Machine Translation:
  Encoder{--}Decoder Approaches}. In \bibinfo{booktitle}{\emph{Proceedings of
  {SSST}-8, Eighth Workshop on Syntax, Semantics and Structure in Statistical
  Translation}}. \bibinfo{pages}{103--111}.
\newblock


\bibitem[\protect\citeauthoryear{Chronopoulou, Stojanovski, and
  Fraser}{Chronopoulou et~al\mbox{.}}{2020}]%
        {chronopoulou-etal-2020-reusing}
\bibfield{author}{\bibinfo{person}{Alexandra Chronopoulou},
  \bibinfo{person}{Dario Stojanovski}, {and} \bibinfo{person}{Alexander
  Fraser}.} \bibinfo{year}{2020}\natexlab{}.
\newblock \showarticletitle{{R}eusing a {P}retrained {L}anguage {M}odel on
  {L}anguages with {L}imited {C}orpora for {U}nsupervised {NMT}}. In
  \bibinfo{booktitle}{\emph{Proceedings of the 2020 Conference on Empirical
  Methods in Natural Language Processing (EMNLP)}}.
  \bibinfo{pages}{2703--2711}.
\newblock


\bibitem[\protect\citeauthoryear{Chu and Wang}{Chu and Wang}{2018}]%
        {chu-wang-2018-survey}
\bibfield{author}{\bibinfo{person}{Chenhui Chu} {and} \bibinfo{person}{Rui
  Wang}.} \bibinfo{year}{2018}\natexlab{}.
\newblock \showarticletitle{A Survey of Domain Adaptation for Neural Machine
  Translation}. In \bibinfo{booktitle}{\emph{Proceedings of the 27th
  International Conference on Computational Linguistics}}.
  \bibinfo{pages}{1304--1319}.
\newblock


\bibitem[\protect\citeauthoryear{Cohen}{Cohen}{2008}]%
        {cohen2008encyclopedia}
\bibfield{author}{\bibinfo{person}{M Cohen}.} \bibinfo{year}{2008}\natexlab{}.
\newblock \showarticletitle{Encyclopedia idiotica}.
\newblock \bibinfo{journal}{\emph{Times Higher Education}}
  \bibinfo{volume}{28} (\bibinfo{year}{2008}).
\newblock


\bibitem[\protect\citeauthoryear{Conneau and Lample}{Conneau and
  Lample}{2019}]%
        {lample2019cross}
\bibfield{author}{\bibinfo{person}{Alexis Conneau} {and}
  \bibinfo{person}{Guillaume Lample}.} \bibinfo{year}{2019}\natexlab{}.
\newblock \showarticletitle{Cross-lingual Language Model Pretraining}. In
  \bibinfo{booktitle}{\emph{Advances in Neural Information Processing Systems
  32: Annual Conference on NeurIPS 2019}}. \bibinfo{pages}{7057--7067}.
\newblock


\bibitem[\protect\citeauthoryear{Cooper~Stickland, Li, and
  Ghazvininejad}{Cooper~Stickland et~al\mbox{.}}{2021}]%
        {stickland2020recipes}
\bibfield{author}{\bibinfo{person}{Asa Cooper~Stickland}, \bibinfo{person}{Xian
  Li}, {and} \bibinfo{person}{Marjan Ghazvininejad}.}
  \bibinfo{year}{2021}\natexlab{}.
\newblock \showarticletitle{Recipes for Adapting Pre-trained Monolingual and
  Multilingual Models to Machine Translation}. In
  \bibinfo{booktitle}{\emph{Proceedings of the 16th Conference of the European
  Chapter of the Association for Computational Linguistics: Main Volume}}.
  \bibinfo{pages}{3440--3453}.
\newblock


\bibitem[\protect\citeauthoryear{Cotterell and Kreutzer}{Cotterell and
  Kreutzer}{2018}]%
        {cotterell2018explaining}
\bibfield{author}{\bibinfo{person}{Ryan Cotterell} {and} \bibinfo{person}{Julia
  Kreutzer}.} \bibinfo{year}{2018}\natexlab{}.
\newblock \showarticletitle{Explaining and generalizing back-translation
  through wake-sleep}.
\newblock \bibinfo{journal}{\emph{arXiv preprint arXiv:1806.04402}}
  (\bibinfo{year}{2018}).
\newblock


\bibitem[\protect\citeauthoryear{Currey and Heafield}{Currey and
  Heafield}{2019}]%
        {currey2019zero}
\bibfield{author}{\bibinfo{person}{Anna Currey} {and} \bibinfo{person}{Kenneth
  Heafield}.} \bibinfo{year}{2019}\natexlab{}.
\newblock \showarticletitle{Zero-Resource Neural Machine Translation with
  Monolingual Pivot Data}. In \bibinfo{booktitle}{\emph{Proceedings of the 3rd
  Workshop on Neural Generation and Translation}}. \bibinfo{pages}{99--107}.
\newblock


\bibitem[\protect\citeauthoryear{Currey, Miceli~Barone, and Heafield}{Currey
  et~al\mbox{.}}{2017}]%
        {currey-etal-2017-copied}
\bibfield{author}{\bibinfo{person}{Anna Currey},
  \bibinfo{person}{Antonio~Valerio Miceli~Barone}, {and}
  \bibinfo{person}{Kenneth Heafield}.} \bibinfo{year}{2017}\natexlab{}.
\newblock \showarticletitle{Copied Monolingual Data Improves Low-Resource
  Neural Machine Translation}. In \bibinfo{booktitle}{\emph{Proceedings of the
  Second Conference on Machine Translation}}. \bibinfo{pages}{148--156}.
\newblock


\bibitem[\protect\citeauthoryear{Dabre, Chu, and Kunchukuttan}{Dabre
  et~al\mbox{.}}{2020}]%
        {dabre2020survey}
\bibfield{author}{\bibinfo{person}{Raj Dabre}, \bibinfo{person}{Chenhui Chu},
  {and} \bibinfo{person}{Anoop Kunchukuttan}.} \bibinfo{year}{2020}\natexlab{}.
\newblock \showarticletitle{A survey of multilingual neural machine
  translation}.
\newblock \bibinfo{journal}{\emph{ACM Computing Surveys (CSUR)}}
  \bibinfo{volume}{53}, \bibinfo{number}{5} (\bibinfo{year}{2020}),
  \bibinfo{pages}{1--38}.
\newblock


\bibitem[\protect\citeauthoryear{Dabre, Fujita, and Chu}{Dabre
  et~al\mbox{.}}{2019}]%
        {dabre-etal-2019-exploiting}
\bibfield{author}{\bibinfo{person}{Raj Dabre}, \bibinfo{person}{Atsushi
  Fujita}, {and} \bibinfo{person}{Chenhui Chu}.}
  \bibinfo{year}{2019}\natexlab{}.
\newblock \showarticletitle{Exploiting Multilingualism through Multistage
  Fine-Tuning for Low-Resource Neural Machine Translation}. In
  \bibinfo{booktitle}{\emph{Proceedings of the 2019 Conference on Empirical
  Methods in Natural Language Processing and the 9th International Joint
  Conference on Natural Language Processing}}. \bibinfo{pages}{1410--1416}.
\newblock


\bibitem[\protect\citeauthoryear{Dabre, Nakagawa, and Kazawa}{Dabre
  et~al\mbox{.}}{2017}]%
        {dabre2017empirical}
\bibfield{author}{\bibinfo{person}{Raj Dabre}, \bibinfo{person}{Tetsuji
  Nakagawa}, {and} \bibinfo{person}{Hideto Kazawa}.}
  \bibinfo{year}{2017}\natexlab{}.
\newblock \showarticletitle{An Empirical Study of Language Relatedness for
  Transfer Learning in Neural Machine Translation}. In
  \bibinfo{booktitle}{\emph{Proceedings of the 31st Pacific Asia Conference on
  Language, Information and Computation}}. \bibinfo{pages}{282--286}.
\newblock


\bibitem[\protect\citeauthoryear{Domhan and Hieber}{Domhan and Hieber}{2017}]%
        {domhan2017using}
\bibfield{author}{\bibinfo{person}{Tobias Domhan} {and} \bibinfo{person}{Felix
  Hieber}.} \bibinfo{year}{2017}\natexlab{}.
\newblock \showarticletitle{Using Target-side Monolingual Data for Neural
  Machine Translation through Multi-task Learning}. In
  \bibinfo{booktitle}{\emph{Proceedings of the 2017 Conference on Empirical
  Methods in NLP}}. \bibinfo{pages}{1500--1505}.
\newblock


\bibitem[\protect\citeauthoryear{Dong, Wu, He, Yu, and Wang}{Dong
  et~al\mbox{.}}{2015}]%
        {dong2015multi}
\bibfield{author}{\bibinfo{person}{Daxiang Dong}, \bibinfo{person}{Hua Wu},
  \bibinfo{person}{Wei He}, \bibinfo{person}{Dianhai Yu}, {and}
  \bibinfo{person}{Haifeng Wang}.} \bibinfo{year}{2015}\natexlab{}.
\newblock \showarticletitle{Multi-Task Learning for Multiple Language
  Translation}. In \bibinfo{booktitle}{\emph{Proceedings of the 53rd Annual
  Meeting of the Association for Computational Linguistics and the 7th
  International Joint Conference on Natural Language Processing (Volume 1: Long
  Papers)}}. \bibinfo{pages}{1723--1732}.
\newblock


\bibitem[\protect\citeauthoryear{Dou, Anastasopoulos, and Neubig}{Dou
  et~al\mbox{.}}{2020}]%
        {dou2020dynamic}
\bibfield{author}{\bibinfo{person}{Zi-Yi Dou}, \bibinfo{person}{Antonios
  Anastasopoulos}, {and} \bibinfo{person}{Graham Neubig}.}
  \bibinfo{year}{2020}\natexlab{}.
\newblock \showarticletitle{Dynamic Data Selection and Weighting for Iterative
  Back-Translation}. In \bibinfo{booktitle}{\emph{Proceedings of the 2020
  Conference on Empirical Methods in Natural Language Processing (EMNLP)}}.
  \bibinfo{pages}{5894--5904}.
\newblock


\bibitem[\protect\citeauthoryear{Dryer}{Dryer}{2011}]%
        {dryer2011martin}
\bibfield{author}{\bibinfo{person}{Matthew~S Dryer}.}
  \bibinfo{year}{2011}\natexlab{}.
\newblock \showarticletitle{Martin Haspelmath, editors. 2013}.
\newblock \bibinfo{journal}{\emph{WALS Online. Max Planck Institute for
  Evolutionary Anthropology, Leipzig}} (\bibinfo{year}{2011}).
\newblock


\bibitem[\protect\citeauthoryear{Duan, Zhao, Zhang, and Wang}{Duan
  et~al\mbox{.}}{2020b}]%
        {duan2020syntax}
\bibfield{author}{\bibinfo{person}{Sufeng Duan}, \bibinfo{person}{Hai Zhao},
  \bibinfo{person}{Dongdong Zhang}, {and} \bibinfo{person}{Rui Wang}.}
  \bibinfo{year}{2020}\natexlab{b}.
\newblock \showarticletitle{Syntax-aware data augmentation for neural machine
  translation}.
\newblock \bibinfo{journal}{\emph{arXiv preprint arXiv:2004.14200}}
  (\bibinfo{year}{2020}).
\newblock


\bibitem[\protect\citeauthoryear{Duan, Ji, Jia, Tan, Zhang, Chen, Luo, and
  Zhang}{Duan et~al\mbox{.}}{2020a}]%
        {duan2020bilingual}
\bibfield{author}{\bibinfo{person}{Xiangyu Duan}, \bibinfo{person}{Baijun Ji},
  \bibinfo{person}{Hao Jia}, \bibinfo{person}{Min Tan}, \bibinfo{person}{Min
  Zhang}, \bibinfo{person}{Boxing Chen}, \bibinfo{person}{Weihua Luo}, {and}
  \bibinfo{person}{Yue Zhang}.} \bibinfo{year}{2020}\natexlab{a}.
\newblock \showarticletitle{Bilingual Dictionary Based Neural Machine
  Translation without Using Parallel Sentences}. In
  \bibinfo{booktitle}{\emph{Proceedings of the 58th Annual Meeting of the
  Association for Computational Linguistics}}. \bibinfo{pages}{1570--1579}.
\newblock


\bibitem[\protect\citeauthoryear{Edman, Toral, and van Noord}{Edman
  et~al\mbox{.}}{2020}]%
        {edman-etal-2020-low}
\bibfield{author}{\bibinfo{person}{Lukas Edman}, \bibinfo{person}{Antonio
  Toral}, {and} \bibinfo{person}{Gertjan van Noord}.}
  \bibinfo{year}{2020}\natexlab{}.
\newblock \showarticletitle{Low-Resource Unsupervised {NMT}: Diagnosing the
  Problem and Providing a Linguistically Motivated Solution}. In
  \bibinfo{booktitle}{\emph{Proceedings of the 22nd Annual Conference of the
  European Association for Machine Translation}}. \bibinfo{pages}{81--90}.
\newblock


\bibitem[\protect\citeauthoryear{Edunov, Ott, Auli, and Grangier}{Edunov
  et~al\mbox{.}}{2018}]%
        {edunov-etal-2018-understanding}
\bibfield{author}{\bibinfo{person}{Sergey Edunov}, \bibinfo{person}{Myle Ott},
  \bibinfo{person}{Michael Auli}, {and} \bibinfo{person}{David Grangier}.}
  \bibinfo{year}{2018}\natexlab{}.
\newblock \showarticletitle{Understanding Back-Translation at Scale}. In
  \bibinfo{booktitle}{\emph{Proceedings of the 2018 Conference on Empirical
  Methods in Natural Language Processing}}. \bibinfo{pages}{489--500}.
\newblock


\bibitem[\protect\citeauthoryear{Edunov, Ott, Ranzato, and Auli}{Edunov
  et~al\mbox{.}}{2020}]%
        {edunov-etal-2020-evaluation}
\bibfield{author}{\bibinfo{person}{Sergey Edunov}, \bibinfo{person}{Myle Ott},
  \bibinfo{person}{Marc{'}Aurelio Ranzato}, {and} \bibinfo{person}{Michael
  Auli}.} \bibinfo{year}{2020}\natexlab{}.
\newblock \showarticletitle{On The Evaluation of Machine Translation Systems
  Trained With Back-Translation}. In \bibinfo{booktitle}{\emph{Proceedings of
  the 58th Annual Meeting of the ACL}}. \bibinfo{pages}{2836--2846}.
\newblock


\bibitem[\protect\citeauthoryear{Espa{\~n}a-Bonet and van
  Genabith}{Espa{\~n}a-Bonet and van Genabith}{2017}]%
        {espana2017going}
\bibfield{author}{\bibinfo{person}{Cristina Espa{\~n}a-Bonet} {and}
  \bibinfo{person}{Josef van Genabith}.} \bibinfo{year}{2017}\natexlab{}.
\newblock \showarticletitle{Going beyond zero-shot MT: combining phonological,
  morphological and semantic factors. The UdS-DFKI System at IWSLT 2017}.
\newblock \bibinfo{journal}{\emph{Proc. of IWSLT}} (\bibinfo{year}{2017}).
\newblock


\bibitem[\protect\citeauthoryear{Fadaee, Bisazza, and Monz}{Fadaee
  et~al\mbox{.}}{2017}]%
        {fadaee2017data}
\bibfield{author}{\bibinfo{person}{Marzieh Fadaee}, \bibinfo{person}{Arianna
  Bisazza}, {and} \bibinfo{person}{Christof Monz}.}
  \bibinfo{year}{2017}\natexlab{}.
\newblock \showarticletitle{Data Augmentation for Low-Resource Neural Machine
  Translation}. In \bibinfo{booktitle}{\emph{Proceedings of the 55th Annual
  Meeting of the Association for Computational Linguistics (Volume 2: Short
  Papers)}}. \bibinfo{pages}{567--573}.
\newblock


\bibitem[\protect\citeauthoryear{Fadaee and Monz}{Fadaee and Monz}{2018}]%
        {fadaee-monz-2018-back}
\bibfield{author}{\bibinfo{person}{Marzieh Fadaee} {and}
  \bibinfo{person}{Christof Monz}.} \bibinfo{year}{2018}\natexlab{}.
\newblock \showarticletitle{Back-Translation Sampling by Targeting Difficult
  Words in Neural Machine Translation}. In
  \bibinfo{booktitle}{\emph{Proceedings of the 2018 Conference on Empirical
  Methods in Natural Language Processing}}. \bibinfo{pages}{436--446}.
\newblock


\bibitem[\protect\citeauthoryear{Fan, Bhosale, Schwenk, Ma, El-Kishky, Goyal,
  Baines, Celebi, Wenzek, Chaudhary, Goyal, Birch, Liptchinsky, Edunov, Grave,
  Auli, and Joulin}{Fan et~al\mbox{.}}{2020}]%
        {fan2020englishcentric}
\bibfield{author}{\bibinfo{person}{Angela Fan}, \bibinfo{person}{Shruti
  Bhosale}, \bibinfo{person}{Holger Schwenk}, \bibinfo{person}{Zhiyi Ma},
  \bibinfo{person}{Ahmed El-Kishky}, \bibinfo{person}{Siddharth Goyal},
  \bibinfo{person}{Mandeep Baines}, \bibinfo{person}{Onur Celebi},
  \bibinfo{person}{Guillaume Wenzek}, \bibinfo{person}{Vishrav Chaudhary},
  \bibinfo{person}{Naman Goyal}, \bibinfo{person}{Tom Birch},
  \bibinfo{person}{Vitaliy Liptchinsky}, \bibinfo{person}{Sergey Edunov},
  \bibinfo{person}{Edouard Grave}, \bibinfo{person}{Michael Auli}, {and}
  \bibinfo{person}{Armand Joulin}.} \bibinfo{year}{2020}\natexlab{}.
\newblock \bibinfo{title}{Beyond English-Centric Multilingual Machine
  Translation}.
\newblock
\newblock


\bibitem[\protect\citeauthoryear{Firat, Cho, and Bengio}{Firat
  et~al\mbox{.}}{2016a}]%
        {firat-etal-2016-multi}
\bibfield{author}{\bibinfo{person}{Orhan Firat}, \bibinfo{person}{Kyunghyun
  Cho}, {and} \bibinfo{person}{Yoshua Bengio}.}
  \bibinfo{year}{2016}\natexlab{a}.
\newblock \showarticletitle{Multi-Way, Multilingual Neural Machine Translation
  with a Shared Attention Mechanism}. In \bibinfo{booktitle}{\emph{Proceedings
  of the 2016 Conference of the North {A}merican Chapter of the Association for
  Computational Linguistics: Human Language Technologies}}.
  \bibinfo{pages}{866--875}.
\newblock


\bibitem[\protect\citeauthoryear{Firat, Sankaran, Al-onaizan, Yarman~Vural, and
  Cho}{Firat et~al\mbox{.}}{2016b}]%
        {firat-etal-2016-zero}
\bibfield{author}{\bibinfo{person}{Orhan Firat}, \bibinfo{person}{Baskaran
  Sankaran}, \bibinfo{person}{Yaser Al-onaizan}, \bibinfo{person}{Fatos~T.
  Yarman~Vural}, {and} \bibinfo{person}{Kyunghyun Cho}.}
  \bibinfo{year}{2016}\natexlab{b}.
\newblock \showarticletitle{Zero-Resource Translation with Multi-Lingual Neural
  Machine Translation}. In \bibinfo{booktitle}{\emph{Proceedings of the 2016
  Conference on Empirical Methods in Natural Language Processing}}.
  \bibinfo{pages}{268--277}.
\newblock


\bibitem[\protect\citeauthoryear{Gao, Zhu, Wu, Xia, Qin, Cheng, Zhou, and
  Liu}{Gao et~al\mbox{.}}{2019}]%
        {gao-etal-2019-soft}
\bibfield{author}{\bibinfo{person}{Fei Gao}, \bibinfo{person}{Jinhua Zhu},
  \bibinfo{person}{Lijun Wu}, \bibinfo{person}{Yingce Xia},
  \bibinfo{person}{Tao Qin}, \bibinfo{person}{Xueqi Cheng},
  \bibinfo{person}{Wengang Zhou}, {and} \bibinfo{person}{Tie-Yan Liu}.}
  \bibinfo{year}{2019}\natexlab{}.
\newblock \showarticletitle{Soft Contextual Data Augmentation for Neural
  Machine Translation}. In \bibinfo{booktitle}{\emph{Proceedings of the 57th
  Annual Meeting of the Association for Computational Linguistics}}.
  \bibinfo{pages}{5539--5544}.
\newblock


\bibitem[\protect\citeauthoryear{Garcia, Foret, Sellam, and Parikh}{Garcia
  et~al\mbox{.}}{2020}]%
        {garcia2020multilingual}
\bibfield{author}{\bibinfo{person}{Xavier Garcia}, \bibinfo{person}{Pierre
  Foret}, \bibinfo{person}{Thibault Sellam}, {and} \bibinfo{person}{Ankur
  Parikh}.} \bibinfo{year}{2020}\natexlab{}.
\newblock \showarticletitle{A Multilingual View of Unsupervised Machine
  Translation}. In \bibinfo{booktitle}{\emph{Findings of the Association for
  Computational Linguistics: EMNLP 2020}}. \bibinfo{pages}{3160--3170}.
\newblock


\bibitem[\protect\citeauthoryear{Garmash and Monz}{Garmash and Monz}{2016}]%
        {garmash2016ensemble}
\bibfield{author}{\bibinfo{person}{Ekaterina Garmash} {and}
  \bibinfo{person}{Christof Monz}.} \bibinfo{year}{2016}\natexlab{}.
\newblock \showarticletitle{Ensemble Learning for Multi-Source Neural Machine
  Translation}. In \bibinfo{booktitle}{\emph{Proceedings of {COLING} 2016, the
  26th International Conference on CL: Technical Papers}}.
  \bibinfo{pages}{1409--1418}.
\newblock


\bibitem[\protect\citeauthoryear{Gheini and May}{Gheini and May}{2019}]%
        {gheini2019universal}
\bibfield{author}{\bibinfo{person}{Mozhdeh Gheini} {and}
  \bibinfo{person}{Jonathan May}.} \bibinfo{year}{2019}\natexlab{}.
\newblock \showarticletitle{A universal parent model for low-resource neural
  machine translation transfer}.
\newblock \bibinfo{journal}{\emph{arXiv preprint arXiv:1909.06516}}
  (\bibinfo{year}{2019}).
\newblock


\bibitem[\protect\citeauthoryear{Gibadullin, Valeev, Khusainova, and
  Khan}{Gibadullin et~al\mbox{.}}{2019}]%
        {gibadullin2019survey}
\bibfield{author}{\bibinfo{person}{Ilshat Gibadullin}, \bibinfo{person}{Aidar
  Valeev}, \bibinfo{person}{Albina Khusainova}, {and} \bibinfo{person}{Adil
  Khan}.} \bibinfo{year}{2019}\natexlab{}.
\newblock \showarticletitle{A Survey of Methods to Leverage Monolingual Data in
  Low-resource Neural Machine Translation}.
\newblock \bibinfo{journal}{\emph{arXiv preprint arXiv:1910.00373}}
  (\bibinfo{year}{2019}).
\newblock


\bibitem[\protect\citeauthoryear{Goyal, Kumar, and Sharma}{Goyal
  et~al\mbox{.}}{2020}]%
        {goyal-etal-2020-efficient}
\bibfield{author}{\bibinfo{person}{Vikrant Goyal}, \bibinfo{person}{Sourav
  Kumar}, {and} \bibinfo{person}{Dipti~Misra Sharma}.}
  \bibinfo{year}{2020}\natexlab{}.
\newblock \showarticletitle{Efficient Neural Machine Translation for
  Low-Resource Languages via Exploiting Related Languages}. In
  \bibinfo{booktitle}{\emph{Proceedings of the 58th Annual Meeting of the
  Association for Computational Linguistics: Student Research Workshop}}.
  \bibinfo{pages}{162--168}.
\newblock


\bibitem[\protect\citeauthoryear{Gra{\c{c}}a, Kim, Schamper, Khadivi, and
  Ney}{Gra{\c{c}}a et~al\mbox{.}}{2019}]%
        {gracca2019generalizing}
\bibfield{author}{\bibinfo{person}{Miguel Gra{\c{c}}a}, \bibinfo{person}{Yunsu
  Kim}, \bibinfo{person}{Julian Schamper}, \bibinfo{person}{Shahram Khadivi},
  {and} \bibinfo{person}{Hermann Ney}.} \bibinfo{year}{2019}\natexlab{}.
\newblock \showarticletitle{Generalizing Back-Translation in Neural Machine
  Translation}. In \bibinfo{booktitle}{\emph{Proceedings of the 4th Conference
  on Machine Translation}}. \bibinfo{pages}{45--52}.
\newblock


\bibitem[\protect\citeauthoryear{Graham, Hogan, Straumann, and Medhat}{Graham
  et~al\mbox{.}}{2014}]%
        {graham2014uneven}
\bibfield{author}{\bibinfo{person}{Mark Graham}, \bibinfo{person}{Bernie
  Hogan}, \bibinfo{person}{Ralph~K Straumann}, {and} \bibinfo{person}{Ahmed
  Medhat}.} \bibinfo{year}{2014}\natexlab{}.
\newblock \showarticletitle{Uneven geographies of user-generated information:
  Patterns of increasing informational poverty}.
\newblock \bibinfo{journal}{\emph{Annals of the Association of American
  Geographers}} \bibinfo{volume}{104}, \bibinfo{number}{4}
  (\bibinfo{year}{2014}), \bibinfo{pages}{746--764}.
\newblock


\bibitem[\protect\citeauthoryear{Gu, Hassan, Devlin, and Li}{Gu
  et~al\mbox{.}}{2018a}]%
        {gu2018universal}
\bibfield{author}{\bibinfo{person}{Jiatao Gu}, \bibinfo{person}{Hany Hassan},
  \bibinfo{person}{Jacob Devlin}, {and} \bibinfo{person}{Victor~O.K. Li}.}
  \bibinfo{year}{2018}\natexlab{a}.
\newblock \showarticletitle{Universal Neural Machine Translation for Extremely
  Low Resource Languages}. In \bibinfo{booktitle}{\emph{Proceedings of the 2018
  Conference of the NAACL: Human Language Technologies, Volume 1 (Long
  Papers)}}. \bibinfo{pages}{344--354}.
\newblock


\bibitem[\protect\citeauthoryear{Gu, Wang, Chen, Li, and Cho}{Gu
  et~al\mbox{.}}{2018b}]%
        {gu2018meta}
\bibfield{author}{\bibinfo{person}{Jiatao Gu}, \bibinfo{person}{Yong Wang},
  \bibinfo{person}{Yun Chen}, \bibinfo{person}{Victor O.~K. Li}, {and}
  \bibinfo{person}{Kyunghyun Cho}.} \bibinfo{year}{2018}\natexlab{b}.
\newblock \showarticletitle{Meta-Learning for Low-Resource Neural Machine
  Translation}. In \bibinfo{booktitle}{\emph{Proceedings of the 2018 Conference
  on Empirical Methods in NLP}}. \bibinfo{pages}{3622--3631}.
\newblock


\bibitem[\protect\citeauthoryear{Gu, Wang, Cho, and Li}{Gu
  et~al\mbox{.}}{2019}]%
        {gu-etal-2019-improved}
\bibfield{author}{\bibinfo{person}{Jiatao Gu}, \bibinfo{person}{Yong Wang},
  \bibinfo{person}{Kyunghyun Cho}, {and} \bibinfo{person}{Victor~O.K. Li}.}
  \bibinfo{year}{2019}\natexlab{}.
\newblock \showarticletitle{Improved Zero-shot Neural Machine Translation via
  Ignoring Spurious Correlations}. In \bibinfo{booktitle}{\emph{Proceedings of
  the 57th Annual Meeting of the ACL}}. \bibinfo{pages}{1258--1268}.
\newblock


\bibitem[\protect\citeauthoryear{Gulcehre, Firat, Xu, Cho, Barrault, Lin,
  Bougares, Schwenk, and Bengio}{Gulcehre et~al\mbox{.}}{2015}]%
        {gulcehre2015using}
\bibfield{author}{\bibinfo{person}{Caglar Gulcehre}, \bibinfo{person}{Orhan
  Firat}, \bibinfo{person}{Kelvin Xu}, \bibinfo{person}{Kyunghyun Cho},
  \bibinfo{person}{Loic Barrault}, \bibinfo{person}{Huei-Chi Lin},
  \bibinfo{person}{Fethi Bougares}, \bibinfo{person}{Holger Schwenk}, {and}
  \bibinfo{person}{Yoshua Bengio}.} \bibinfo{year}{2015}\natexlab{}.
\newblock \showarticletitle{On using monolingual corpora in neural machine
  translation}.
\newblock \bibinfo{journal}{\emph{arXiv preprint arXiv:1503.03535}}
  (\bibinfo{year}{2015}).
\newblock


\bibitem[\protect\citeauthoryear{Guo, Shen, Yang, Ge, Cer, Hernandez~Abrego,
  Stevens, Constant, Sung, Strope, and Kurzweil}{Guo et~al\mbox{.}}{2018}]%
        {guo2018effective}
\bibfield{author}{\bibinfo{person}{Mandy Guo}, \bibinfo{person}{Qinlan Shen},
  \bibinfo{person}{Yinfei Yang}, \bibinfo{person}{Heming Ge},
  \bibinfo{person}{Daniel Cer}, \bibinfo{person}{Gustavo Hernandez~Abrego},
  \bibinfo{person}{Keith Stevens}, \bibinfo{person}{Noah Constant},
  \bibinfo{person}{Yun-Hsuan Sung}, \bibinfo{person}{Brian Strope}, {and}
  \bibinfo{person}{Ray Kurzweil}.} \bibinfo{year}{2018}\natexlab{}.
\newblock \showarticletitle{Effective Parallel Corpus Mining using Bilingual
  Sentence Embeddings}. In \bibinfo{booktitle}{\emph{Proceedings of the Third
  Conference on Machine Translation: Research Papers}}.
  \bibinfo{pages}{165--176}.
\newblock


\bibitem[\protect\citeauthoryear{Guzm{\'a}n, Chen, Ott, Pino, Lample, Koehn,
  Chaudhary, and Ranzato}{Guzm{\'a}n et~al\mbox{.}}{2019}]%
        {guzman-etal-2019-flores}
\bibfield{author}{\bibinfo{person}{Francisco Guzm{\'a}n},
  \bibinfo{person}{Peng-Jen Chen}, \bibinfo{person}{Myle Ott},
  \bibinfo{person}{Juan Pino}, \bibinfo{person}{Guillaume Lample},
  \bibinfo{person}{Philipp Koehn}, \bibinfo{person}{Vishrav Chaudhary}, {and}
  \bibinfo{person}{Marc{'}Aurelio Ranzato}.} \bibinfo{year}{2019}\natexlab{}.
\newblock \showarticletitle{The {FLORES} Evaluation Datasets for Low-Resource
  Machine Translation: {N}epali{--}{E}nglish and {S}inhala{--}{E}nglish}. In
  \bibinfo{booktitle}{\emph{Proceedings of the 2019 Conference on Empirical
  Methods in Natural Language Processing and the 9th International Joint
  Conference on Natural Language Processing (EMNLP-IJCNLP)}}.
  \bibinfo{pages}{6098--6111}.
\newblock


\bibitem[\protect\citeauthoryear{Ha, Niehues, and Waibel}{Ha
  et~al\mbox{.}}{2016}]%
        {ha2016toward}
\bibfield{author}{\bibinfo{person}{Thanh-Le Ha}, \bibinfo{person}{Jan Niehues},
  {and} \bibinfo{person}{Alexander Waibel}.} \bibinfo{year}{2016}\natexlab{}.
\newblock \showarticletitle{Toward multilingual neural machine translation with
  universal encoder and decoder}.
\newblock \bibinfo{journal}{\emph{arXiv preprint arXiv:1611.04798}}
  (\bibinfo{year}{2016}).
\newblock


\bibitem[\protect\citeauthoryear{Ha, Niehues, and Waibel}{Ha
  et~al\mbox{.}}{2017}]%
        {ha2017effective}
\bibfield{author}{\bibinfo{person}{Thanh-Le Ha}, \bibinfo{person}{Jan Niehues},
  {and} \bibinfo{person}{Alexander Waibel}.} \bibinfo{year}{2017}\natexlab{}.
\newblock \showarticletitle{Effective strategies in zero-shot neural machine
  translation}.
\newblock \bibinfo{journal}{\emph{arXiv preprint arXiv:1711.07893}}
  (\bibinfo{year}{2017}).
\newblock


\bibitem[\protect\citeauthoryear{Hangya and Fraser}{Hangya and Fraser}{2019}]%
        {hangya2019unsupervised}
\bibfield{author}{\bibinfo{person}{Viktor Hangya} {and}
  \bibinfo{person}{Alexander Fraser}.} \bibinfo{year}{2019}\natexlab{}.
\newblock \showarticletitle{Unsupervised Parallel Sentence Extraction with
  Parallel Segment Detection Helps Machine Translation}. In
  \bibinfo{booktitle}{\emph{Proceedings of the 57th Annual Meeting of the
  ACL}}. \bibinfo{pages}{1224--1234}.
\newblock


\bibitem[\protect\citeauthoryear{Harmon}{Harmon}{1995}]%
        {harmon1995status}
\bibfield{author}{\bibinfo{person}{David Harmon}.}
  \bibinfo{year}{1995}\natexlab{}.
\newblock \showarticletitle{The Status of the World's Languages as Reported in"
  Ethnologue".}
\newblock \bibinfo{journal}{\emph{Southwest Journal of Linguistics}}
  \bibinfo{volume}{14} (\bibinfo{year}{1995}), \bibinfo{pages}{1--28}.
\newblock


\bibitem[\protect\citeauthoryear{He, Xia, Qin, Wang, Yu, Liu, and Ma}{He
  et~al\mbox{.}}{2016}]%
        {he2016dual}
\bibfield{author}{\bibinfo{person}{Di He}, \bibinfo{person}{Yingce Xia},
  \bibinfo{person}{Tao Qin}, \bibinfo{person}{Liwei Wang},
  \bibinfo{person}{Nenghai Yu}, \bibinfo{person}{Tie{-}Yan Liu}, {and}
  \bibinfo{person}{Wei{-}Ying Ma}.} \bibinfo{year}{2016}\natexlab{}.
\newblock \showarticletitle{Dual Learning for Machine Translation}. In
  \bibinfo{booktitle}{\emph{Advances in Neural Information Processing Systems
  29: Annual Conference on Neural Information Processing Systems 2016}}.
  \bibinfo{pages}{820--828}.
\newblock


\bibitem[\protect\citeauthoryear{Hedderich, Lange, Adel, Str{\"o}tgen, and
  Klakow}{Hedderich et~al\mbox{.}}{2021}]%
        {hedderich2020survey}
\bibfield{author}{\bibinfo{person}{Michael~A. Hedderich},
  \bibinfo{person}{Lukas Lange}, \bibinfo{person}{Heike Adel},
  \bibinfo{person}{Jannik Str{\"o}tgen}, {and} \bibinfo{person}{Dietrich
  Klakow}.} \bibinfo{year}{2021}\natexlab{}.
\newblock \showarticletitle{A Survey on Recent Approaches for Natural Language
  Processing in Low-Resource Scenarios}. In
  \bibinfo{booktitle}{\emph{Proceedings of the 2021 Conference of the North
  American Chapter of the Association for Computational Linguistics: Human
  Language Technologies}}. \bibinfo{pages}{2545--2568}.
\newblock


\bibitem[\protect\citeauthoryear{Hoang, Koehn, Haffari, and Cohn}{Hoang
  et~al\mbox{.}}{2018}]%
        {hoang2018iterative}
\bibfield{author}{\bibinfo{person}{Vu~Cong~Duy Hoang}, \bibinfo{person}{Philipp
  Koehn}, \bibinfo{person}{Gholamreza Haffari}, {and} \bibinfo{person}{Trevor
  Cohn}.} \bibinfo{year}{2018}\natexlab{}.
\newblock \showarticletitle{Iterative Back-Translation for Neural Machine
  Translation}. In \bibinfo{booktitle}{\emph{Proceedings of the 2nd Workshop on
  NMT and Generation}}. \bibinfo{pages}{18--24}.
\newblock


\bibitem[\protect\citeauthoryear{Hokamp, Glover, and
  Gholipour~Ghalandari}{Hokamp et~al\mbox{.}}{2019}]%
        {hokamp2019evaluating}
\bibfield{author}{\bibinfo{person}{Chris Hokamp}, \bibinfo{person}{John
  Glover}, {and} \bibinfo{person}{Demian Gholipour~Ghalandari}.}
  \bibinfo{year}{2019}\natexlab{}.
\newblock \showarticletitle{Evaluating the Supervised and Zero-shot Performance
  of Multi-lingual Translation Models}. In
  \bibinfo{booktitle}{\emph{Proceedings of the Fourth Conference on Machine
  Translation (Volume 2: Shared Task Papers, Day 1)}}.
  \bibinfo{pages}{209--217}.
\newblock


\bibitem[\protect\citeauthoryear{Hutchins}{Hutchins}{1997}]%
        {hutchins1997first}
\bibfield{author}{\bibinfo{person}{John Hutchins}.}
  \bibinfo{year}{1997}\natexlab{}.
\newblock \showarticletitle{From first conception to first demonstration: the
  nascent years of machine translation, 1947--1954. a chronology}.
\newblock \bibinfo{journal}{\emph{Machine Translation}} \bibinfo{volume}{12},
  \bibinfo{number}{3} (\bibinfo{year}{1997}), \bibinfo{pages}{195--252}.
\newblock


\bibitem[\protect\citeauthoryear{Imamura, Fujita, and Sumita}{Imamura
  et~al\mbox{.}}{2018}]%
        {imamura2018enhancement}
\bibfield{author}{\bibinfo{person}{Kenji Imamura}, \bibinfo{person}{Atsushi
  Fujita}, {and} \bibinfo{person}{Eiichiro Sumita}.}
  \bibinfo{year}{2018}\natexlab{}.
\newblock \showarticletitle{Enhancement of Encoder and Attention Using Target
  Monolingual Corpora in Neural Machine Translation}. In
  \bibinfo{booktitle}{\emph{Proceedings of the 2nd Workshop on Neural Machine
  Translation and Generation}}. \bibinfo{pages}{55--63}.
\newblock


\bibitem[\protect\citeauthoryear{Imankulova, Dabre, Fujita, and
  Imamura}{Imankulova et~al\mbox{.}}{2019a}]%
        {imankulova-etal-2019-exploiting}
\bibfield{author}{\bibinfo{person}{Aizhan Imankulova}, \bibinfo{person}{Raj
  Dabre}, \bibinfo{person}{Atsushi Fujita}, {and} \bibinfo{person}{Kenji
  Imamura}.} \bibinfo{year}{2019}\natexlab{a}.
\newblock \showarticletitle{Exploiting Out-of-Domain Parallel Data through
  Multilingual Transfer Learning for Low-Resource Neural Machine Translation}.
  In \bibinfo{booktitle}{\emph{Proceedings of Machine Translation Summit XVII
  Volume 1: Research Track}}. \bibinfo{pages}{128--139}.
\newblock


\bibitem[\protect\citeauthoryear{Imankulova, Sato, and Komachi}{Imankulova
  et~al\mbox{.}}{2017}]%
        {imankulova2017improving}
\bibfield{author}{\bibinfo{person}{Aizhan Imankulova},
  \bibinfo{person}{Takayuki Sato}, {and} \bibinfo{person}{Mamoru Komachi}.}
  \bibinfo{year}{2017}\natexlab{}.
\newblock \showarticletitle{Improving Low-Resource Neural Machine Translation
  with Filtered Pseudo-Parallel Corpus}. In
  \bibinfo{booktitle}{\emph{Proceedings of the 4th Workshop on {A}sian
  Translation ({WAT}2017)}}. \bibinfo{pages}{70--78}.
\newblock


\bibitem[\protect\citeauthoryear{Imankulova, Sato, and Komachi}{Imankulova
  et~al\mbox{.}}{2019b}]%
        {imankulova2019filtered}
\bibfield{author}{\bibinfo{person}{Aizhan Imankulova},
  \bibinfo{person}{Takayuki Sato}, {and} \bibinfo{person}{Mamoru Komachi}.}
  \bibinfo{year}{2019}\natexlab{b}.
\newblock \showarticletitle{Filtered pseudo-parallel corpus improves
  low-resource neural machine translation}.
\newblock \bibinfo{journal}{\emph{ACM Transactions on Asian and Low-Resource
  Language Information Processing (TALLIP)}} \bibinfo{volume}{19},
  \bibinfo{number}{2} (\bibinfo{year}{2019}), \bibinfo{pages}{1--16}.
\newblock


\bibitem[\protect\citeauthoryear{Ji, Zhang, Duan, Zhang, Chen, and Luo}{Ji
  et~al\mbox{.}}{2020}]%
        {ji2020cross}
\bibfield{author}{\bibinfo{person}{Baijun Ji}, \bibinfo{person}{Zhirui Zhang},
  \bibinfo{person}{Xiangyu Duan}, \bibinfo{person}{Min Zhang},
  \bibinfo{person}{Boxing Chen}, {and} \bibinfo{person}{Weihua Luo}.}
  \bibinfo{year}{2020}\natexlab{}.
\newblock \showarticletitle{Cross-Lingual Pre-Training Based Transfer for
  Zero-Shot Neural Machine Translation}. In \bibinfo{booktitle}{\emph{The
  Thirty-Fourth {AAAI} Conference on Artificial Intelligence, {AAAI} 2020, The
  Thirty-Second Innovative Applications of Artificial Intelligence Conference,
  {IAAI} 2020, The Tenth {AAAI} Symposium on Educational Advances in Artificial
  Intelligence, {EAAI} 2020}}. \bibinfo{pages}{115--122}.
\newblock


\bibitem[\protect\citeauthoryear{Johnson, Schuster, Le, Krikun, Wu, Chen,
  Thorat, Vi{\'e}gas, Wattenberg, Corrado, Hughes, and Dean}{Johnson
  et~al\mbox{.}}{2017a}]%
        {johnson2017google}
\bibfield{author}{\bibinfo{person}{Melvin Johnson}, \bibinfo{person}{Mike
  Schuster}, \bibinfo{person}{Quoc~V. Le}, \bibinfo{person}{Maxim Krikun},
  \bibinfo{person}{Yonghui Wu}, \bibinfo{person}{Zhifeng Chen},
  \bibinfo{person}{Nikhil Thorat}, \bibinfo{person}{Fernanda Vi{\'e}gas},
  \bibinfo{person}{Martin Wattenberg}, \bibinfo{person}{Greg Corrado},
  \bibinfo{person}{Macduff Hughes}, {and} \bibinfo{person}{Jeffrey Dean}.}
  \bibinfo{year}{2017}\natexlab{a}.
\newblock \showarticletitle{{G}oogle{'}s Multilingual Neural Machine
  Translation System: Enabling Zero-Shot Translation}.
\newblock \bibinfo{journal}{\emph{Transactions of the Association for
  Computational Linguistics}}  \bibinfo{volume}{5} (\bibinfo{year}{2017}),
  \bibinfo{pages}{339--351}.
\newblock


\bibitem[\protect\citeauthoryear{Johnson, Schuster, Le, Krikun, Wu, Chen,
  Thorat, Vi{\'e}gas, Wattenberg, Corrado, Hughes, and Dean}{Johnson
  et~al\mbox{.}}{2017b}]%
        {DBLP:journals/corr/JohnsonSLKWCTVW16}
\bibfield{author}{\bibinfo{person}{Melvin Johnson}, \bibinfo{person}{Mike
  Schuster}, \bibinfo{person}{Quoc~V. Le}, \bibinfo{person}{Maxim Krikun},
  \bibinfo{person}{Yonghui Wu}, \bibinfo{person}{Zhifeng Chen},
  \bibinfo{person}{Nikhil Thorat}, \bibinfo{person}{Fernanda Vi{\'e}gas},
  \bibinfo{person}{Martin Wattenberg}, \bibinfo{person}{Greg Corrado},
  \bibinfo{person}{Macduff Hughes}, {and} \bibinfo{person}{Jeffrey Dean}.}
  \bibinfo{year}{2017}\natexlab{b}.
\newblock \showarticletitle{{G}oogle{'}s Multilingual Neural Machine
  Translation System: Enabling Zero-Shot Translation}.
\newblock \bibinfo{journal}{\emph{Transactions of the ACL}}
  \bibinfo{volume}{5} (\bibinfo{year}{2017}), \bibinfo{pages}{339--351}.
\newblock


\bibitem[\protect\citeauthoryear{Joshi, Santy, Budhiraja, Bali, and
  Choudhury}{Joshi et~al\mbox{.}}{2020}]%
        {joshi2020state}
\bibfield{author}{\bibinfo{person}{Pratik Joshi}, \bibinfo{person}{Sebastin
  Santy}, \bibinfo{person}{Amar Budhiraja}, \bibinfo{person}{Kalika Bali},
  {and} \bibinfo{person}{Monojit Choudhury}.} \bibinfo{year}{2020}\natexlab{}.
\newblock \showarticletitle{The State and Fate of Linguistic Diversity and
  Inclusion in the {NLP} World}. In \bibinfo{booktitle}{\emph{Proceedings of
  the 58th Annual Meeting of the Association for Computational Linguistics}}.
  \bibinfo{pages}{6282--6293}.
\newblock


\bibitem[\protect\citeauthoryear{Karakanta, Dehdari, and van
  Genabith}{Karakanta et~al\mbox{.}}{2018}]%
        {karakanta2018neural}
\bibfield{author}{\bibinfo{person}{Alina Karakanta}, \bibinfo{person}{Jon
  Dehdari}, {and} \bibinfo{person}{Josef van Genabith}.}
  \bibinfo{year}{2018}\natexlab{}.
\newblock \showarticletitle{Neural machine translation for low-resource
  languages without parallel corpora}.
\newblock \bibinfo{journal}{\emph{Machine Translation}} \bibinfo{volume}{32},
  \bibinfo{number}{1-2} (\bibinfo{year}{2018}), \bibinfo{pages}{167--189}.
\newblock


\bibitem[\protect\citeauthoryear{Keung, Salazar, Lu, and Smith}{Keung
  et~al\mbox{.}}{2020}]%
        {keung2020unsupervised}
\bibfield{author}{\bibinfo{person}{Phillip Keung}, \bibinfo{person}{Julian
  Salazar}, \bibinfo{person}{Yichao Lu}, {and} \bibinfo{person}{Noah~A.
  Smith}.} \bibinfo{year}{2020}\natexlab{}.
\newblock \showarticletitle{Unsupervised Bitext Mining and Translation via
  Self-Trained Contextual Embeddings}.
\newblock \bibinfo{journal}{\emph{Transactions of the Association for
  Computational Linguistics}}  \bibinfo{volume}{8} (\bibinfo{year}{2020}),
  \bibinfo{pages}{828--841}.
\newblock


\bibitem[\protect\citeauthoryear{Khatri and Bhattacharyya}{Khatri and
  Bhattacharyya}{2020}]%
        {khatri2020filtering}
\bibfield{author}{\bibinfo{person}{Jyotsana Khatri} {and}
  \bibinfo{person}{Pushpak Bhattacharyya}.} \bibinfo{year}{2020}\natexlab{}.
\newblock \showarticletitle{Filtering Back-Translated Data in Unsupervised
  Neural Machine Translation}. In \bibinfo{booktitle}{\emph{Proceedings of the
  28th International Conference on Computational Linguistics}}.
  \bibinfo{pages}{4334--4339}.
\newblock


\bibitem[\protect\citeauthoryear{Kim, Gao, and Ney}{Kim et~al\mbox{.}}{2019a}]%
        {kim2019effective}
\bibfield{author}{\bibinfo{person}{Yunsu Kim}, \bibinfo{person}{Yingbo Gao},
  {and} \bibinfo{person}{Hermann Ney}.} \bibinfo{year}{2019}\natexlab{a}.
\newblock \showarticletitle{Effective Cross-lingual Transfer of Neural Machine
  Translation Models without Shared Vocabularies}. In
  \bibinfo{booktitle}{\emph{Proceedings of the 57th Annual Meeting of the
  ACL}}. \bibinfo{pages}{1246--1257}.
\newblock


\bibitem[\protect\citeauthoryear{Kim, Gra{\c{c}}a, and Ney}{Kim
  et~al\mbox{.}}{2020}]%
        {kim-etal-2020-unsupervised}
\bibfield{author}{\bibinfo{person}{Yunsu Kim}, \bibinfo{person}{Miguel
  Gra{\c{c}}a}, {and} \bibinfo{person}{Hermann Ney}.}
  \bibinfo{year}{2020}\natexlab{}.
\newblock \showarticletitle{When and Why is Unsupervised Neural Machine
  Translation Useless?}. In \bibinfo{booktitle}{\emph{Proceedings of the 22nd
  Annual Conference of the European Association for Machine Translation}}.
  \bibinfo{pages}{35--44}.
\newblock


\bibitem[\protect\citeauthoryear{Kim, Petrov, Petrushkov, Khadivi, and Ney}{Kim
  et~al\mbox{.}}{2019b}]%
        {kim2019pivot}
\bibfield{author}{\bibinfo{person}{Yunsu Kim}, \bibinfo{person}{Petre Petrov},
  \bibinfo{person}{Pavel Petrushkov}, \bibinfo{person}{Shahram Khadivi}, {and}
  \bibinfo{person}{Hermann Ney}.} \bibinfo{year}{2019}\natexlab{b}.
\newblock \showarticletitle{Pivot-based Transfer Learning for Neural Machine
  Translation between Non-{E}nglish Languages}. In
  \bibinfo{booktitle}{\emph{Proceedings of the 2019 Conference on Empirical
  Methods in NLP and the 9th International Joint Conference on Natural Language
  Processing}}. \bibinfo{pages}{866--876}.
\newblock


\bibitem[\protect\citeauthoryear{Klein, Kim, Deng, Senellart, and Rush}{Klein
  et~al\mbox{.}}{2017}]%
        {klein-etal-2017-opennmt}
\bibfield{author}{\bibinfo{person}{Guillaume Klein}, \bibinfo{person}{Yoon
  Kim}, \bibinfo{person}{Yuntian Deng}, \bibinfo{person}{Jean Senellart}, {and}
  \bibinfo{person}{Alexander Rush}.} \bibinfo{year}{2017}\natexlab{}.
\newblock \showarticletitle{{O}pen{NMT}: Open-Source Toolkit for Neural Machine
  Translation}. In \bibinfo{booktitle}{\emph{Proceedings of {ACL} 2017, System
  Demonstrations}}. \bibinfo{pages}{67--72}.
\newblock


\bibitem[\protect\citeauthoryear{Kocmi and Bojar}{Kocmi and Bojar}{2018}]%
        {kocmi2018trivial}
\bibfield{author}{\bibinfo{person}{Tom Kocmi} {and}
  \bibinfo{person}{Ond{\v{r}}ej Bojar}.} \bibinfo{year}{2018}\natexlab{}.
\newblock \showarticletitle{Trivial Transfer Learning for Low-Resource Neural
  Machine Translation}. In \bibinfo{booktitle}{\emph{Proceedings of the Third
  Conference on Machine Translation: Research Papers}}.
  \bibinfo{pages}{244--252}.
\newblock


\bibitem[\protect\citeauthoryear{Kocmi and Bojar}{Kocmi and Bojar}{2020}]%
        {kocmi2020efficiently}
\bibfield{author}{\bibinfo{person}{Tom Kocmi} {and}
  \bibinfo{person}{Ond{\v{r}}ej Bojar}.} \bibinfo{year}{2020}\natexlab{}.
\newblock \showarticletitle{Efficiently Reusing Old Models Across Languages via
  Transfer Learning}. In \bibinfo{booktitle}{\emph{Proceedings of the 22nd
  Annual Conference of the European Association for Machine Translation}}.
  \bibinfo{pages}{19--28}.
\newblock


\bibitem[\protect\citeauthoryear{Koehn, Guzm{\'a}n, Chaudhary, and Pino}{Koehn
  et~al\mbox{.}}{2019}]%
        {koehn-etal-2019-findings}
\bibfield{author}{\bibinfo{person}{Philipp Koehn}, \bibinfo{person}{Francisco
  Guzm{\'a}n}, \bibinfo{person}{Vishrav Chaudhary}, {and} \bibinfo{person}{Juan
  Pino}.} \bibinfo{year}{2019}\natexlab{}.
\newblock \showarticletitle{Findings of the {WMT} 2019 Shared Task on Parallel
  Corpus Filtering for Low-Resource Conditions}. In
  \bibinfo{booktitle}{\emph{Proceedings of the Fourth Conference on Machine
  Translation (Volume 3: Shared Task Papers, Day 2)}}. \bibinfo{pages}{54--72}.
\newblock


\bibitem[\protect\citeauthoryear{Koehn and Knowles}{Koehn and Knowles}{2017}]%
        {koehn2017six}
\bibfield{author}{\bibinfo{person}{Philipp Koehn} {and}
  \bibinfo{person}{Rebecca Knowles}.} \bibinfo{year}{2017}\natexlab{}.
\newblock \showarticletitle{Six Challenges for Neural Machine Translation}. In
  \bibinfo{booktitle}{\emph{Proceedings of the First Workshop on Neural Machine
  Translation}}. \bibinfo{pages}{28--39}.
\newblock


\bibitem[\protect\citeauthoryear{Kudugunta, Bapna, Caswell, and
  Firat}{Kudugunta et~al\mbox{.}}{2019}]%
        {kudugunta-etal-2019-investigating}
\bibfield{author}{\bibinfo{person}{Sneha Kudugunta}, \bibinfo{person}{Ankur
  Bapna}, \bibinfo{person}{Isaac Caswell}, {and} \bibinfo{person}{Orhan
  Firat}.} \bibinfo{year}{2019}\natexlab{}.
\newblock \showarticletitle{Investigating Multilingual {NMT} Representations at
  Scale}. In \bibinfo{booktitle}{\emph{Proceedings of the 2019 Conference on
  Empirical Methods in Natural Language Processing and the 9th International
  Joint Conference on Natural Language Processing (EMNLP-IJCNLP)}}.
  \bibinfo{pages}{1565--1575}.
\newblock


\bibitem[\protect\citeauthoryear{Lai, Dai, and Yang}{Lai et~al\mbox{.}}{2020}]%
        {lai2020unsupervised}
\bibfield{author}{\bibinfo{person}{Guokun Lai}, \bibinfo{person}{Zihang Dai},
  {and} \bibinfo{person}{Yiming Yang}.} \bibinfo{year}{2020}\natexlab{}.
\newblock \showarticletitle{Unsupervised Parallel Corpus Mining on Web Data}.
\newblock \bibinfo{journal}{\emph{arXiv preprint arXiv:2009.08595}}
  (\bibinfo{year}{2020}).
\newblock


\bibitem[\protect\citeauthoryear{Lakew, Cettolo, and Federico}{Lakew
  et~al\mbox{.}}{2018a}]%
        {lakew-etal-2018-comparison}
\bibfield{author}{\bibinfo{person}{Surafel~Melaku Lakew},
  \bibinfo{person}{Mauro Cettolo}, {and} \bibinfo{person}{Marcello Federico}.}
  \bibinfo{year}{2018}\natexlab{a}.
\newblock \showarticletitle{A Comparison of Transformer and Recurrent Neural
  Networks on Multilingual Neural Machine Translation}. In
  \bibinfo{booktitle}{\emph{Proceedings of the 27th International Conference on
  Computational Linguistics}}. \bibinfo{pages}{641--652}.
\newblock


\bibitem[\protect\citeauthoryear{Lakew, Erofeeva, Negri, Federico, and
  Turchi}{Lakew et~al\mbox{.}}{2018b}]%
        {lakew2018transfer}
\bibfield{author}{\bibinfo{person}{Surafel~M Lakew}, \bibinfo{person}{Aliia
  Erofeeva}, \bibinfo{person}{Matteo Negri}, \bibinfo{person}{Marcello
  Federico}, {and} \bibinfo{person}{Marco Turchi}.}
  \bibinfo{year}{2018}\natexlab{b}.
\newblock \showarticletitle{Transfer learning in multilingual neural machine
  translation with dynamic vocabulary}.
\newblock \bibinfo{journal}{\emph{arXiv preprint arXiv:1811.01137}}
  (\bibinfo{year}{2018}).
\newblock


\bibitem[\protect\citeauthoryear{Lakew, Federico, Negri, and Turchi}{Lakew
  et~al\mbox{.}}{2019a}]%
        {lakew2019multilingual}
\bibfield{author}{\bibinfo{person}{Surafel~M Lakew}, \bibinfo{person}{Marcello
  Federico}, \bibinfo{person}{Matteo Negri}, {and} \bibinfo{person}{Marco
  Turchi}.} \bibinfo{year}{2019}\natexlab{a}.
\newblock \showarticletitle{Multilingual Neural Machine Translation for
  Zero-Resource Languages}.
\newblock \bibinfo{journal}{\emph{arXiv preprint arXiv:1909.07342}}
  (\bibinfo{year}{2019}).
\newblock


\bibitem[\protect\citeauthoryear{Lakew, Karakanta, Federico, Negri, and
  Turchi}{Lakew et~al\mbox{.}}{2019b}]%
        {lakew2019adapting}
\bibfield{author}{\bibinfo{person}{Surafel~M Lakew}, \bibinfo{person}{Alina
  Karakanta}, \bibinfo{person}{Marcello Federico}, \bibinfo{person}{Matteo
  Negri}, {and} \bibinfo{person}{Marco Turchi}.}
  \bibinfo{year}{2019}\natexlab{b}.
\newblock \showarticletitle{Adapting Multilingual Neural Machine Translation to
  Unseen Languages}.
\newblock \bibinfo{journal}{\emph{arXiv preprint arXiv:1910.13998}}
  (\bibinfo{year}{2019}).
\newblock


\bibitem[\protect\citeauthoryear{Lample, Conneau, Denoyer, and Ranzato}{Lample
  et~al\mbox{.}}{2018a}]%
        {lample2018unsupervised}
\bibfield{author}{\bibinfo{person}{Guillaume Lample}, \bibinfo{person}{Alexis
  Conneau}, \bibinfo{person}{Ludovic Denoyer}, {and}
  \bibinfo{person}{Marc'Aurelio Ranzato}.} \bibinfo{year}{2018}\natexlab{a}.
\newblock \showarticletitle{Unsupervised Machine Translation Using Monolingual
  Corpora Only}. In \bibinfo{booktitle}{\emph{6th International Conference on
  Learning Representations, {ICLR} 2018, Conference Track Proceedings}}.
\newblock


\bibitem[\protect\citeauthoryear{Lample, Conneau, Ranzato, Denoyer, and
  J{\'{e}}gou}{Lample et~al\mbox{.}}{2018b}]%
        {conneau2017word}
\bibfield{author}{\bibinfo{person}{Guillaume Lample}, \bibinfo{person}{Alexis
  Conneau}, \bibinfo{person}{Marc'Aurelio Ranzato}, \bibinfo{person}{Ludovic
  Denoyer}, {and} \bibinfo{person}{Herv{\'{e}} J{\'{e}}gou}.}
  \bibinfo{year}{2018}\natexlab{b}.
\newblock \showarticletitle{Word translation without parallel data}. In
  \bibinfo{booktitle}{\emph{6th International Conference on Learning
  Representations, Conference Track Proceedings}}.
\newblock


\bibitem[\protect\citeauthoryear{Lample, Ott, Conneau, Denoyer, and
  Ranzato}{Lample et~al\mbox{.}}{2018c}]%
        {lample-etal-2018-phrase}
\bibfield{author}{\bibinfo{person}{Guillaume Lample}, \bibinfo{person}{Myle
  Ott}, \bibinfo{person}{Alexis Conneau}, \bibinfo{person}{Ludovic Denoyer},
  {and} \bibinfo{person}{Marc{'}Aurelio Ranzato}.}
  \bibinfo{year}{2018}\natexlab{c}.
\newblock \showarticletitle{Phrase-Based {\&} Neural Unsupervised Machine
  Translation}. In \bibinfo{booktitle}{\emph{Proceedings of the 2018 Conference
  on Empirical Methods in Natural Language Processing}}.
  \bibinfo{pages}{5039--5049}.
\newblock


\bibitem[\protect\citeauthoryear{Li, Wang, and Yu}{Li et~al\mbox{.}}{2020b}]%
        {li2019metamta}
\bibfield{author}{\bibinfo{person}{Rumeng Li}, \bibinfo{person}{Xun Wang},
  {and} \bibinfo{person}{Hong Yu}.} \bibinfo{year}{2020}\natexlab{b}.
\newblock \showarticletitle{MetaMT, a Meta Learning Method Leveraging Multiple
  Domain Data for Low Resource Machine Translation}. In
  \bibinfo{booktitle}{\emph{The Thirty-Fourth {AAAI} Conference on Artificial
  Intelligence, {AAAI} 2020, The Thirty-Second Innovative Applications of
  Artificial Intelligence Conference, {IAAI} 2020, The Tenth {AAAI} Symposium
  on Educational Advances in Artificial Intelligence, {EAAI} 2020}}.
  \bibinfo{pages}{8245--8252}.
\newblock


\bibitem[\protect\citeauthoryear{Li, Khot, Khashabi, Sabharwal, and
  Srikumar}{Li et~al\mbox{.}}{2020a}]%
        {DBLP:journals/corr/abs-2010-02428}
\bibfield{author}{\bibinfo{person}{Tao Li}, \bibinfo{person}{Tushar Khot},
  \bibinfo{person}{Daniel Khashabi}, \bibinfo{person}{Ashish Sabharwal}, {and}
  \bibinfo{person}{Vivek Srikumar}.} \bibinfo{year}{2020}\natexlab{a}.
\newblock \showarticletitle{UnQovering Stereotyping Biases via Underspecified
  Questions}.
\newblock \bibinfo{journal}{\emph{CoRR}}  \bibinfo{volume}{abs/2010.02428}
  (\bibinfo{year}{2020}).
\newblock


\bibitem[\protect\citeauthoryear{Li, Zhao, Wang, Utiyama, and Sumita}{Li
  et~al\mbox{.}}{2020c}]%
        {li2020reference}
\bibfield{author}{\bibinfo{person}{Zuchao Li}, \bibinfo{person}{Hai Zhao},
  \bibinfo{person}{Rui Wang}, \bibinfo{person}{Masao Utiyama}, {and}
  \bibinfo{person}{Eiichiro Sumita}.} \bibinfo{year}{2020}\natexlab{c}.
\newblock \showarticletitle{Reference Language based Unsupervised Neural
  Machine Translation}. In \bibinfo{booktitle}{\emph{Findings of the
  Association for Computational Linguistics: EMNLP 2020}}.
  \bibinfo{pages}{4151--4162}.
\newblock


\bibitem[\protect\citeauthoryear{Lin, Chen, Lee, Li, Zhang, Xia, Rijhwani, He,
  Zhang, Ma, Anastasopoulos, Littell, and Neubig}{Lin et~al\mbox{.}}{2019}]%
        {lin2019choosing}
\bibfield{author}{\bibinfo{person}{Yu-Hsiang Lin}, \bibinfo{person}{Chian-Yu
  Chen}, \bibinfo{person}{Jean Lee}, \bibinfo{person}{Zirui Li},
  \bibinfo{person}{Yuyan Zhang}, \bibinfo{person}{Mengzhou Xia},
  \bibinfo{person}{Shruti Rijhwani}, \bibinfo{person}{Junxian He},
  \bibinfo{person}{Zhisong Zhang}, \bibinfo{person}{Xuezhe Ma},
  \bibinfo{person}{Antonios Anastasopoulos}, \bibinfo{person}{Patrick Littell},
  {and} \bibinfo{person}{Graham Neubig}.} \bibinfo{year}{2019}\natexlab{}.
\newblock \showarticletitle{Choosing Transfer Languages for Cross-Lingual
  Learning}. In \bibinfo{booktitle}{\emph{Proceedings of the 57th Annual
  Meeting of the ACL}}. \bibinfo{pages}{3125--3135}.
\newblock


\bibitem[\protect\citeauthoryear{Liu, Silva, Wang, and Way}{Liu
  et~al\mbox{.}}{2018}]%
        {liu2018pivot}
\bibfield{author}{\bibinfo{person}{Chao-Hong Liu},
  \bibinfo{person}{Catarina~Cruz Silva}, \bibinfo{person}{Longyue Wang}, {and}
  \bibinfo{person}{Andy Way}.} \bibinfo{year}{2018}\natexlab{}.
\newblock \showarticletitle{Pivot machine translation using Chinese as pivot
  language}. In \bibinfo{booktitle}{\emph{China Workshop on Machine
  Translation}}. Springer, \bibinfo{pages}{74--85}.
\newblock


\bibitem[\protect\citeauthoryear{Liu, Kusner, and Blunsom}{Liu
  et~al\mbox{.}}{2021}]%
        {liu2021counterfactual}
\bibfield{author}{\bibinfo{person}{Qi Liu}, \bibinfo{person}{Matt Kusner},
  {and} \bibinfo{person}{Phil Blunsom}.} \bibinfo{year}{2021}\natexlab{}.
\newblock \showarticletitle{Counterfactual Data Augmentation for Neural Machine
  Translation}. In \bibinfo{booktitle}{\emph{Proceedings of the 2021 Conference
  of the North American Chapter of the Association for Computational
  Linguistics: Human Language Technologies}}. \bibinfo{pages}{187--197}.
\newblock


\bibitem[\protect\citeauthoryear{Liu, Gu, Goyal, Li, Edunov, Ghazvininejad,
  Lewis, and Zettlemoyer}{Liu et~al\mbox{.}}{2020}]%
        {liu2020multilingual}
\bibfield{author}{\bibinfo{person}{Yinhan Liu}, \bibinfo{person}{Jiatao Gu},
  \bibinfo{person}{Naman Goyal}, \bibinfo{person}{Xian Li},
  \bibinfo{person}{Sergey Edunov}, \bibinfo{person}{Marjan Ghazvininejad},
  \bibinfo{person}{Mike Lewis}, {and} \bibinfo{person}{Luke Zettlemoyer}.}
  \bibinfo{year}{2020}\natexlab{}.
\newblock \showarticletitle{Multilingual Denoising Pre-training for Neural
  Machine Translation}.
\newblock \bibinfo{journal}{\emph{Transactions of the Association for
  Computational Linguistics}}  \bibinfo{volume}{8} (\bibinfo{year}{2020}),
  \bibinfo{pages}{726--742}.
\newblock


\bibitem[\protect\citeauthoryear{Lu, Keung, Ladhak, Bhardwaj, Zhang, and
  Sun}{Lu et~al\mbox{.}}{2018}]%
        {lu2018neural}
\bibfield{author}{\bibinfo{person}{Yichao Lu}, \bibinfo{person}{Phillip Keung},
  \bibinfo{person}{Faisal Ladhak}, \bibinfo{person}{Vikas Bhardwaj},
  \bibinfo{person}{Shaonan Zhang}, {and} \bibinfo{person}{Jason Sun}.}
  \bibinfo{year}{2018}\natexlab{}.
\newblock \showarticletitle{A neural interlingua for multilingual machine
  translation}. In \bibinfo{booktitle}{\emph{Proceedings of the 3rd Conference
  on MT: Research Papers}}. \bibinfo{pages}{84--92}.
\newblock


\bibitem[\protect\citeauthoryear{Luo, Yang, Yuan, Chen, and Ainiwaer}{Luo
  et~al\mbox{.}}{2019}]%
        {luo2019hierarchical}
\bibfield{author}{\bibinfo{person}{Gongxu Luo}, \bibinfo{person}{Yating Yang},
  \bibinfo{person}{Yang Yuan}, \bibinfo{person}{Zhanheng Chen}, {and}
  \bibinfo{person}{Aizimaiti Ainiwaer}.} \bibinfo{year}{2019}\natexlab{}.
\newblock \showarticletitle{Hierarchical transfer learning architecture for
  low-resource neural machine translation}.
\newblock \bibinfo{journal}{\emph{IEEE Access}}  \bibinfo{volume}{7}
  (\bibinfo{year}{2019}), \bibinfo{pages}{154157--154166}.
\newblock


\bibitem[\protect\citeauthoryear{Maimaiti, Liu, Luan, and Sun}{Maimaiti
  et~al\mbox{.}}{2019}]%
        {maimaiti2019multi}
\bibfield{author}{\bibinfo{person}{Mieradilijiang Maimaiti},
  \bibinfo{person}{Yang Liu}, \bibinfo{person}{Huanbo Luan}, {and}
  \bibinfo{person}{Maosong Sun}.} \bibinfo{year}{2019}\natexlab{}.
\newblock \showarticletitle{Multi-Round Transfer Learning for Low-Resource NMT
  Using Multiple High-Resource Languages}.
\newblock \bibinfo{journal}{\emph{ACM Transactions on Asian and Low-Resource
  Language Information Processing (TALLIP)}} \bibinfo{volume}{18},
  \bibinfo{number}{4} (\bibinfo{year}{2019}), \bibinfo{pages}{1--26}.
\newblock


\bibitem[\protect\citeauthoryear{Maimaiti, Liu, Luan, and Sun}{Maimaiti
  et~al\mbox{.}}{2020}]%
        {maimaiti2020enriching}
\bibfield{author}{\bibinfo{person}{Mieradilijiang Maimaiti},
  \bibinfo{person}{Yang Liu}, \bibinfo{person}{Huanbo Luan}, {and}
  \bibinfo{person}{Maosong Sun}.} \bibinfo{year}{2020}\natexlab{}.
\newblock \showarticletitle{Enriching the Transfer Learning with Pre-Trained
  Lexicon Embedding for Low-Resource Neural Machine Translation}.
\newblock \bibinfo{journal}{\emph{Tsinghua Science and Technology}}
  (\bibinfo{year}{2020}), \bibinfo{pages}{1}.
\newblock


\bibitem[\protect\citeauthoryear{Marchisio, Duh, and Koehn}{Marchisio
  et~al\mbox{.}}{2020}]%
        {DBLP:conf/wmt/MarchisioDK20}
\bibfield{author}{\bibinfo{person}{Kelly Marchisio}, \bibinfo{person}{Kevin
  Duh}, {and} \bibinfo{person}{Philipp Koehn}.}
  \bibinfo{year}{2020}\natexlab{}.
\newblock \showarticletitle{When Does Unsupervised Machine Translation Work?}.
  In \bibinfo{booktitle}{\emph{Proceedings of the 5th Conference on Machine
  Translation 2020, Online, November 19-20, 2020}}. \bibinfo{pages}{571--583}.
\newblock


\bibitem[\protect\citeauthoryear{Marie, Rubino, and Fujita}{Marie
  et~al\mbox{.}}{2020}]%
        {marie2020tagged}
\bibfield{author}{\bibinfo{person}{Benjamin Marie}, \bibinfo{person}{Raphael
  Rubino}, {and} \bibinfo{person}{Atsushi Fujita}.}
  \bibinfo{year}{2020}\natexlab{}.
\newblock \showarticletitle{Tagged Back-translation Revisited: Why Does It
  Really Work?}. In \bibinfo{booktitle}{\emph{Proceedings of the 58th Annual
  Meeting of the Association for Computational Linguistics}}.
  \bibinfo{pages}{5990--5997}.
\newblock


\bibitem[\protect\citeauthoryear{Maruf, Saleh, and Haffari}{Maruf
  et~al\mbox{.}}{2019}]%
        {maruf2019survey}
\bibfield{author}{\bibinfo{person}{Sameen Maruf}, \bibinfo{person}{Fahimeh
  Saleh}, {and} \bibinfo{person}{Gholamreza Haffari}.}
  \bibinfo{year}{2019}\natexlab{}.
\newblock \showarticletitle{A survey on document-level machine translation:
  Methods and evaluation}.
\newblock \bibinfo{journal}{\emph{arXiv preprint arXiv:1912.08494}}
  (\bibinfo{year}{2019}).
\newblock


\bibitem[\protect\citeauthoryear{McGuffie and Newhouse}{McGuffie and
  Newhouse}{2020}]%
        {mcguffie2020radicalization}
\bibfield{author}{\bibinfo{person}{Kris McGuffie} {and} \bibinfo{person}{Alex
  Newhouse}.} \bibinfo{year}{2020}\natexlab{}.
\newblock \showarticletitle{The radicalization risks of GPT-3 and advanced
  neural language models}.
\newblock \bibinfo{journal}{\emph{arXiv preprint arXiv:2009.06807}}
  (\bibinfo{year}{2020}).
\newblock


\bibitem[\protect\citeauthoryear{Murthy, Kunchukuttan, and
  Bhattacharyya}{Murthy et~al\mbox{.}}{2019}]%
        {murthy2018addressing}
\bibfield{author}{\bibinfo{person}{Rudra Murthy}, \bibinfo{person}{Anoop
  Kunchukuttan}, {and} \bibinfo{person}{Pushpak Bhattacharyya}.}
  \bibinfo{year}{2019}\natexlab{}.
\newblock \showarticletitle{Addressing word-order Divergence in Multilingual
  Neural Machine Translation for extremely Low Resource Languages}. In
  \bibinfo{booktitle}{\emph{Proceedings of the 2019 Conference of the NAACL:
  Human Language Technologies, Volume 1 (Long and Short Papers)}}.
  \bibinfo{pages}{3868--3873}.
\newblock


\bibitem[\protect\citeauthoryear{Nag, Kale, Lakshminarasimhan, and
  Singhavi}{Nag et~al\mbox{.}}{2020}]%
        {nag2020incorporating}
\bibfield{author}{\bibinfo{person}{Sreyashi Nag}, \bibinfo{person}{Mihir Kale},
  \bibinfo{person}{Varun Lakshminarasimhan}, {and} \bibinfo{person}{Swapnil
  Singhavi}.} \bibinfo{year}{2020}\natexlab{}.
\newblock \bibinfo{title}{Incorporating Bilingual Dictionaries for Low Resource
  Semi-Supervised Neural Machine Translation}.
\newblock
\newblock


\bibitem[\protect\citeauthoryear{Nekoto, Marivate, Matsila, Fasubaa,
  Fagbohungbe, Akinola, Muhammad, Kabongo~Kabenamualu, Osei, Sackey, Niyongabo,
  Macharm, Ogayo, Ahia, Berhe, Adeyemi, Mokgesi-Selinga, Okegbemi, Martinus,
  Tajudeen, Degila, Ogueji, Siminyu, Kreutzer, Webster, Ali, Abbott, Orife,
  Ezeani, Dangana, Kamper, Elsahar, Duru, Kioko, Espoir, van Biljon, Whitenack,
  Onyefuluchi, Emezue, Dossou, Sibanda, Bassey, Olabiyi, Ramkilowan, {\"O}ktem,
  Akinfaderin, and Bashir}{Nekoto et~al\mbox{.}}{2020}]%
        {nekoto2020participatory}
\bibfield{author}{\bibinfo{person}{Wilhelmina Nekoto}, \bibinfo{person}{Vukosi
  Marivate}, \bibinfo{person}{Tshinondiwa Matsila}, \bibinfo{person}{Timi
  Fasubaa}, \bibinfo{person}{Taiwo Fagbohungbe},
  \bibinfo{person}{Solomon~Oluwole Akinola}, \bibinfo{person}{Shamsuddeen
  Muhammad}, \bibinfo{person}{Salomon Kabongo~Kabenamualu},
  \bibinfo{person}{Salomey Osei}, \bibinfo{person}{Freshia Sackey},
  \bibinfo{person}{Rubungo~Andre Niyongabo}, \bibinfo{person}{Ricky Macharm},
  \bibinfo{person}{Perez Ogayo}, \bibinfo{person}{Orevaoghene Ahia},
  \bibinfo{person}{Musie~Meressa Berhe}, \bibinfo{person}{Mofetoluwa Adeyemi},
  \bibinfo{person}{Masabata Mokgesi-Selinga}, \bibinfo{person}{Lawrence
  Okegbemi}, \bibinfo{person}{Laura Martinus}, \bibinfo{person}{Kolawole
  Tajudeen}, \bibinfo{person}{Kevin Degila}, \bibinfo{person}{Kelechi Ogueji},
  \bibinfo{person}{Kathleen Siminyu}, \bibinfo{person}{Julia Kreutzer},
  \bibinfo{person}{Jason Webster}, \bibinfo{person}{Jamiil~Toure Ali},
  \bibinfo{person}{Jade Abbott}, \bibinfo{person}{Iroro Orife},
  \bibinfo{person}{Ignatius Ezeani}, \bibinfo{person}{Idris~Abdulkadir
  Dangana}, \bibinfo{person}{Herman Kamper}, \bibinfo{person}{Hady Elsahar},
  \bibinfo{person}{Goodness Duru}, \bibinfo{person}{Ghollah Kioko},
  \bibinfo{person}{Murhabazi Espoir}, \bibinfo{person}{Elan van Biljon},
  \bibinfo{person}{Daniel Whitenack}, \bibinfo{person}{Christopher
  Onyefuluchi}, \bibinfo{person}{Chris~Chinenye Emezue},
  \bibinfo{person}{Bonaventure F.~P. Dossou}, \bibinfo{person}{Blessing
  Sibanda}, \bibinfo{person}{Blessing Bassey}, \bibinfo{person}{Ayodele
  Olabiyi}, \bibinfo{person}{Arshath Ramkilowan}, \bibinfo{person}{Alp
  {\"O}ktem}, \bibinfo{person}{Adewale Akinfaderin}, {and}
  \bibinfo{person}{Abdallah Bashir}.} \bibinfo{year}{2020}\natexlab{}.
\newblock \showarticletitle{Participatory Research for Low-resourced Machine
  Translation: A Case Study in {A}frican Languages}. In
  \bibinfo{booktitle}{\emph{Findings of the Association for Computational
  Linguistics: EMNLP 2020}}. \bibinfo{pages}{2144--2160}.
\newblock


\bibitem[\protect\citeauthoryear{Neubig and Hu}{Neubig and Hu}{2018}]%
        {neubig2018rapid}
\bibfield{author}{\bibinfo{person}{Graham Neubig} {and} \bibinfo{person}{Junjie
  Hu}.} \bibinfo{year}{2018}\natexlab{}.
\newblock \showarticletitle{Rapid Adaptation of Neural Machine Translation to
  New Languages}. In \bibinfo{booktitle}{\emph{Proceedings of the 2018
  Conference on Empirical Methods in Natural Language Processing}}.
  \bibinfo{pages}{875--880}.
\newblock


\bibitem[\protect\citeauthoryear{Nguyen and Chiang}{Nguyen and Chiang}{2017}]%
        {nguyen-2017-transfer}
\bibfield{author}{\bibinfo{person}{Toan~Q. Nguyen} {and} \bibinfo{person}{David
  Chiang}.} \bibinfo{year}{2017}\natexlab{}.
\newblock \showarticletitle{Transfer Learning across Low-Resource, Related
  Languages for Neural Machine Translation}. In
  \bibinfo{booktitle}{\emph{Proceedings of the Eighth International Joint
  Conference on NLP}}. \bibinfo{pages}{296--301}.
\newblock


\bibitem[\protect\citeauthoryear{Nishimura, Sudoh, Neubig, and
  Nakamura}{Nishimura et~al\mbox{.}}{2018a}]%
        {nishimura2018multia}
\bibfield{author}{\bibinfo{person}{Yuta Nishimura}, \bibinfo{person}{Katsuhito
  Sudoh}, \bibinfo{person}{Graham Neubig}, {and} \bibinfo{person}{Satoshi
  Nakamura}.} \bibinfo{year}{2018}\natexlab{a}.
\newblock \showarticletitle{Multi-source neural machine translation with data
  augmentation}.
\newblock \bibinfo{journal}{\emph{arXiv preprint arXiv:1810.06826}}
  (\bibinfo{year}{2018}).
\newblock


\bibitem[\protect\citeauthoryear{Nishimura, Sudoh, Neubig, and
  Nakamura}{Nishimura et~al\mbox{.}}{2018b}]%
        {nishimura2018multi}
\bibfield{author}{\bibinfo{person}{Yuta Nishimura}, \bibinfo{person}{Katsuhito
  Sudoh}, \bibinfo{person}{Graham Neubig}, {and} \bibinfo{person}{Satoshi
  Nakamura}.} \bibinfo{year}{2018}\natexlab{b}.
\newblock \showarticletitle{Multi-Source Neural Machine Translation with
  Missing Data}. In \bibinfo{booktitle}{\emph{Proceedings of the 2nd Workshop
  on Neural Machine Translation and Generation}}. \bibinfo{pages}{92--99}.
\newblock


\bibitem[\protect\citeauthoryear{Ott, Edunov, Baevski, Fan, Gross, Ng,
  Grangier, and Auli}{Ott et~al\mbox{.}}{2019}]%
        {ott-etal-2019-fairseq}
\bibfield{author}{\bibinfo{person}{Myle Ott}, \bibinfo{person}{Sergey Edunov},
  \bibinfo{person}{Alexei Baevski}, \bibinfo{person}{Angela Fan},
  \bibinfo{person}{Sam Gross}, \bibinfo{person}{Nathan Ng},
  \bibinfo{person}{David Grangier}, {and} \bibinfo{person}{Michael Auli}.}
  \bibinfo{year}{2019}\natexlab{}.
\newblock \showarticletitle{fairseq: A Fast, Extensible Toolkit for Sequence
  Modeling}. In \bibinfo{booktitle}{\emph{Proceedings of the 2019 Conference of
  the North {A}merican Chapter of the Association for Computational Linguistics
  (Demonstrations)}}. \bibinfo{pages}{48--53}.
\newblock


\bibitem[\protect\citeauthoryear{Pan and Yang}{Pan and Yang}{2009}]%
        {pan2009survey}
\bibfield{author}{\bibinfo{person}{Sinno~Jialin Pan} {and}
  \bibinfo{person}{Qiang Yang}.} \bibinfo{year}{2009}\natexlab{}.
\newblock \showarticletitle{A survey on transfer learning}.
\newblock \bibinfo{journal}{\emph{IEEE Transactions on knowledge and data
  engineering}} \bibinfo{volume}{22}, \bibinfo{number}{10}
  (\bibinfo{year}{2009}), \bibinfo{pages}{1345--1359}.
\newblock


\bibitem[\protect\citeauthoryear{Peng, Huang, Li, Chen, and Liu}{Peng
  et~al\mbox{.}}{2020}]%
        {peng2020dictionary}
\bibfield{author}{\bibinfo{person}{Wei Peng}, \bibinfo{person}{Chongxuan
  Huang}, \bibinfo{person}{Tianhao Li}, \bibinfo{person}{Yun Chen}, {and}
  \bibinfo{person}{Qun Liu}.} \bibinfo{year}{2020}\natexlab{}.
\newblock \showarticletitle{Dictionary-based data augmentation for cross-domain
  neural machine translation}.
\newblock \bibinfo{journal}{\emph{arXiv preprint arXiv:2004.02577}}
  (\bibinfo{year}{2020}).
\newblock


\bibitem[\protect\citeauthoryear{Pham, Niehues, Ha, and Waibel}{Pham
  et~al\mbox{.}}{2019}]%
        {pham2019improving}
\bibfield{author}{\bibinfo{person}{Ngoc-Quan Pham}, \bibinfo{person}{Jan
  Niehues}, \bibinfo{person}{Thanh-Le Ha}, {and} \bibinfo{person}{Alexander
  Waibel}.} \bibinfo{year}{2019}\natexlab{}.
\newblock \showarticletitle{Improving Zero-shot Translation with
  Language-Independent Constraints}. In \bibinfo{booktitle}{\emph{Proceedings
  of the Fourth Conference on Machine Translation}}. \bibinfo{pages}{13--23}.
\newblock


\bibitem[\protect\citeauthoryear{Platanios, Sachan, Neubig, and
  Mitchell}{Platanios et~al\mbox{.}}{2018}]%
        {Platanios2018ContextualPG}
\bibfield{author}{\bibinfo{person}{Emmanouil~Antonios Platanios},
  \bibinfo{person}{Mrinmaya Sachan}, \bibinfo{person}{Graham Neubig}, {and}
  \bibinfo{person}{Tom Mitchell}.} \bibinfo{year}{2018}\natexlab{}.
\newblock \showarticletitle{Contextual Parameter Generation for Universal
  Neural Machine Translation}. In \bibinfo{booktitle}{\emph{Proceedings of the
  2018 Conference on Empirical Methods in Natural Language Processing}}.
  \bibinfo{pages}{425--435}.
\newblock


\bibitem[\protect\citeauthoryear{Popescu-Belis}{Popescu-Belis}{2019}]%
        {popescu2019context}
\bibfield{author}{\bibinfo{person}{Andrei Popescu-Belis}.}
  \bibinfo{year}{2019}\natexlab{}.
\newblock \showarticletitle{Context in neural machine translation: A review of
  models and evaluations}.
\newblock \bibinfo{journal}{\emph{arXiv preprint arXiv:1901.09115}}
  (\bibinfo{year}{2019}).
\newblock


\bibitem[\protect\citeauthoryear{Pourdamghani, Aldarrab, Ghazvininejad, Knight,
  and May}{Pourdamghani et~al\mbox{.}}{2019}]%
        {pourdamghani-etal-2019-translating}
\bibfield{author}{\bibinfo{person}{Nima Pourdamghani}, \bibinfo{person}{Nada
  Aldarrab}, \bibinfo{person}{Marjan Ghazvininejad}, \bibinfo{person}{Kevin
  Knight}, {and} \bibinfo{person}{Jonathan May}.}
  \bibinfo{year}{2019}\natexlab{}.
\newblock \showarticletitle{Translating Translationese: A Two-Step Approach to
  Unsupervised Machine Translation}. In \bibinfo{booktitle}{\emph{Proceedings
  of the 57th Annual Meeting of the Association for Computational
  Linguistics}}. \bibinfo{pages}{3057--3062}.
\newblock


\bibitem[\protect\citeauthoryear{Qi, Sachan, Felix, Padmanabhan, and Neubig}{Qi
  et~al\mbox{.}}{2018}]%
        {qi-etal-2018-pre}
\bibfield{author}{\bibinfo{person}{Ye Qi}, \bibinfo{person}{Devendra Sachan},
  \bibinfo{person}{Matthieu Felix}, \bibinfo{person}{Sarguna Padmanabhan},
  {and} \bibinfo{person}{Graham Neubig}.} \bibinfo{year}{2018}\natexlab{}.
\newblock \showarticletitle{When and Why Are Pre-Trained Word Embeddings Useful
  for Neural Machine Translation?}. In \bibinfo{booktitle}{\emph{Proceedings of
  the 2018 Conference of the North {A}merican Chapter of the Association for
  Computational Linguistics: Human Language Technologies}}.
  \bibinfo{pages}{529--535}.
\newblock


\bibitem[\protect\citeauthoryear{Raganato, V{\'a}zquez, Creutz, and
  Tiedemann}{Raganato et~al\mbox{.}}{2019}]%
        {raganato2019evaluation}
\bibfield{author}{\bibinfo{person}{Alessandro Raganato},
  \bibinfo{person}{Ra{\'u}l V{\'a}zquez}, \bibinfo{person}{Mathias Creutz},
  {and} \bibinfo{person}{J{\"o}rg Tiedemann}.} \bibinfo{year}{2019}\natexlab{}.
\newblock \showarticletitle{An Evaluation of Language-Agnostic
  Inner-Attention-Based Representations in Machine Translation}. In
  \bibinfo{booktitle}{\emph{Proceedings of the 4th Workshop on Representation
  Learning for NLP}}. \bibinfo{pages}{27--32}.
\newblock


\bibitem[\protect\citeauthoryear{Ramachandran, Liu, and Le}{Ramachandran
  et~al\mbox{.}}{2017}]%
        {ramachandran2016unsupervised}
\bibfield{author}{\bibinfo{person}{Prajit Ramachandran}, \bibinfo{person}{Peter
  Liu}, {and} \bibinfo{person}{Quoc Le}.} \bibinfo{year}{2017}\natexlab{}.
\newblock \showarticletitle{Unsupervised Pretraining for Sequence to Sequence
  Learning}. In \bibinfo{booktitle}{\emph{Proceedings of the 2017 Conference on
  Empirical Methods in Natural Language Processing}}.
  \bibinfo{pages}{383--391}.
\newblock


\bibitem[\protect\citeauthoryear{Ren, Chen, Liu, Li, Zhou, and Ma}{Ren
  et~al\mbox{.}}{2018}]%
        {ren2018triangular}
\bibfield{author}{\bibinfo{person}{Shuo Ren}, \bibinfo{person}{Wenhu Chen},
  \bibinfo{person}{Shujie Liu}, \bibinfo{person}{Mu Li}, \bibinfo{person}{Ming
  Zhou}, {and} \bibinfo{person}{Shuai Ma}.} \bibinfo{year}{2018}\natexlab{}.
\newblock \showarticletitle{Triangular Architecture for Rare Language
  Translation}. In \bibinfo{booktitle}{\emph{Proceedings of the 56th Annual
  Meeting of the Association for Computational Linguistics (Volume 1: Long
  Papers)}}. \bibinfo{pages}{56--65}.
\newblock


\bibitem[\protect\citeauthoryear{Ren, Wu, Liu, Zhou, and Ma}{Ren
  et~al\mbox{.}}{2019}]%
        {ren-etal-2019-explicit}
\bibfield{author}{\bibinfo{person}{Shuo Ren}, \bibinfo{person}{Yu Wu},
  \bibinfo{person}{Shujie Liu}, \bibinfo{person}{Ming Zhou}, {and}
  \bibinfo{person}{Shuai Ma}.} \bibinfo{year}{2019}\natexlab{}.
\newblock \showarticletitle{Explicit Cross-lingual Pre-training for
  Unsupervised Machine Translation}. In \bibinfo{booktitle}{\emph{Proceedings
  of the 2019 Conference on Empirical Methods in Natural Language Processing
  and the 9th International Joint Conference on Natural Language Processing
  (EMNLP-IJCNLP)}}. \bibinfo{pages}{770--779}.
\newblock


\bibitem[\protect\citeauthoryear{Ren, Wu, Liu, Zhou, and Ma}{Ren
  et~al\mbox{.}}{2020}]%
        {ren2020retrieve}
\bibfield{author}{\bibinfo{person}{Shuo Ren}, \bibinfo{person}{Yu Wu},
  \bibinfo{person}{Shujie Liu}, \bibinfo{person}{Ming Zhou}, {and}
  \bibinfo{person}{Shuai Ma}.} \bibinfo{year}{2020}\natexlab{}.
\newblock \showarticletitle{A Retrieve-and-Rewrite Initialization Method for
  Unsupervised Machine Translation}. In \bibinfo{booktitle}{\emph{Proceedings
  of the 58th Annual Meeting of the ACL}}. \bibinfo{pages}{3498--3504}.
\newblock


\bibitem[\protect\citeauthoryear{Ruiter, van Genabith, and
  Espa{\~n}a-Bonet}{Ruiter et~al\mbox{.}}{2020}]%
        {ruiter2020self}
\bibfield{author}{\bibinfo{person}{Dana Ruiter}, \bibinfo{person}{Josef van
  Genabith}, {and} \bibinfo{person}{Cristina Espa{\~n}a-Bonet}.}
  \bibinfo{year}{2020}\natexlab{}.
\newblock \showarticletitle{Self-Induced Curriculum Learning in Self-Supervised
  Neural Machine Translation}. In \bibinfo{booktitle}{\emph{Proceedings of the
  2020 Conference on Empirical Methods in Natural Language Processing
  (EMNLP)}}. \bibinfo{pages}{2560--2571}.
\newblock


\bibitem[\protect\citeauthoryear{Sachan and Neubig}{Sachan and Neubig}{2018}]%
        {sachan-neubig-2018-parameter}
\bibfield{author}{\bibinfo{person}{Devendra Sachan} {and}
  \bibinfo{person}{Graham Neubig}.} \bibinfo{year}{2018}\natexlab{}.
\newblock \showarticletitle{Parameter Sharing Methods for Multilingual
  Self-Attentional Translation Models}. In
  \bibinfo{booktitle}{\emph{Proceedings of the Third Conference on Machine
  Translation: Research Papers}}. \bibinfo{pages}{261--271}.
\newblock


\bibitem[\protect\citeauthoryear{Schwenk}{Schwenk}{2018}]%
        {schwenk2018filtering}
\bibfield{author}{\bibinfo{person}{Holger Schwenk}.}
  \bibinfo{year}{2018}\natexlab{}.
\newblock \showarticletitle{Filtering and Mining Parallel Data in a Joint
  Multilingual Space}. In \bibinfo{booktitle}{\emph{Proceedings of the 56th
  Annual Meeting of the Association for Computational Linguistics (Volume 2:
  Short Papers)}}. \bibinfo{pages}{228--234}.
\newblock


\bibitem[\protect\citeauthoryear{Schwenk, Chaudhary, Sun, Gong, and
  Guzm{\'a}n}{Schwenk et~al\mbox{.}}{2021}]%
        {schwenk2019wikimatrix}
\bibfield{author}{\bibinfo{person}{Holger Schwenk}, \bibinfo{person}{Vishrav
  Chaudhary}, \bibinfo{person}{Shuo Sun}, \bibinfo{person}{Hongyu Gong}, {and}
  \bibinfo{person}{Francisco Guzm{\'a}n}.} \bibinfo{year}{2021}\natexlab{}.
\newblock \showarticletitle{{W}iki{M}atrix: Mining 135{M} Parallel Sentences in
  1620 Language Pairs from {W}ikipedia}. In
  \bibinfo{booktitle}{\emph{Proceedings of the 16th Conference of the European
  Chapter of the Association for Computational Linguistics: Main Volume}}.
  \bibinfo{pages}{1351--1361}.
\newblock


\bibitem[\protect\citeauthoryear{Schwenk, Wenzek, Edunov, Grave, and
  Joulin}{Schwenk et~al\mbox{.}}{2019}]%
        {schwenk2019ccmatrix}
\bibfield{author}{\bibinfo{person}{Holger Schwenk}, \bibinfo{person}{Guillaume
  Wenzek}, \bibinfo{person}{Sergey Edunov}, \bibinfo{person}{Edouard Grave},
  {and} \bibinfo{person}{Armand Joulin}.} \bibinfo{year}{2019}\natexlab{}.
\newblock \showarticletitle{Ccmatrix: Mining billions of high-quality parallel
  sentences on the web}.
\newblock \bibinfo{journal}{\emph{arXiv preprint arXiv:1911.04944}}
  (\bibinfo{year}{2019}).
\newblock


\bibitem[\protect\citeauthoryear{Sen, Gupta, Ekbal, and Bhattacharyya}{Sen
  et~al\mbox{.}}{2019}]%
        {sen2019multilingual}
\bibfield{author}{\bibinfo{person}{Sukanta Sen}, \bibinfo{person}{Kamal~Kumar
  Gupta}, \bibinfo{person}{Asif Ekbal}, {and} \bibinfo{person}{Pushpak
  Bhattacharyya}.} \bibinfo{year}{2019}\natexlab{}.
\newblock \showarticletitle{Multilingual Unsupervised {NMT} using Shared
  Encoder and Language-Specific Decoders}. In
  \bibinfo{booktitle}{\emph{Proceedings of the 57th Annual Meeting of the
  Association for Computational Linguistics}}. \bibinfo{pages}{3083--3089}.
\newblock


\bibitem[\protect\citeauthoryear{Sennrich, Haddow, and Birch}{Sennrich
  et~al\mbox{.}}{2016}]%
        {sennrich-etal-2016-improving}
\bibfield{author}{\bibinfo{person}{Rico Sennrich}, \bibinfo{person}{Barry
  Haddow}, {and} \bibinfo{person}{Alexandra Birch}.}
  \bibinfo{year}{2016}\natexlab{}.
\newblock \showarticletitle{Improving Neural Machine Translation Models with
  Monolingual Data}. In \bibinfo{booktitle}{\emph{Proceedings of the 54th
  Annual Meeting of the Association for Computational Linguistics}}.
  \bibinfo{pages}{86--96}.
\newblock


\bibitem[\protect\citeauthoryear{Sestorain, Ciaramita, Buck, and
  Hofmann}{Sestorain et~al\mbox{.}}{2018}]%
        {sestorain2018zero}
\bibfield{author}{\bibinfo{person}{Lierni Sestorain},
  \bibinfo{person}{Massimiliano Ciaramita}, \bibinfo{person}{Christian Buck},
  {and} \bibinfo{person}{Thomas Hofmann}.} \bibinfo{year}{2018}\natexlab{}.
\newblock \showarticletitle{Zero-shot dual machine translation}.
\newblock \bibinfo{journal}{\emph{arXiv preprint arXiv:1805.10338}}
  (\bibinfo{year}{2018}).
\newblock


\bibitem[\protect\citeauthoryear{Siddhant, Bapna, Cao, Firat, Chen, Kudugunta,
  Arivazhagan, and Wu}{Siddhant et~al\mbox{.}}{2020}]%
        {siddhant-etal-2020-leveraging}
\bibfield{author}{\bibinfo{person}{Aditya Siddhant}, \bibinfo{person}{Ankur
  Bapna}, \bibinfo{person}{Yuan Cao}, \bibinfo{person}{Orhan Firat},
  \bibinfo{person}{Mia Chen}, \bibinfo{person}{Sneha Kudugunta},
  \bibinfo{person}{Naveen Arivazhagan}, {and} \bibinfo{person}{Yonghui Wu}.}
  \bibinfo{year}{2020}\natexlab{}.
\newblock \showarticletitle{Leveraging Monolingual Data with Self-Supervision
  for Multilingual Neural Machine Translation}. In
  \bibinfo{booktitle}{\emph{Proceedings of the 58th Annual Meeting of the
  Association for Computational Linguistics}}. \bibinfo{pages}{2827--2835}.
\newblock


\bibitem[\protect\citeauthoryear{Skorokhodov, Rykachevskiy, Emelyanenko,
  Slotin, and Ponkratov}{Skorokhodov et~al\mbox{.}}{2018}]%
        {skorokhodov2018semi}
\bibfield{author}{\bibinfo{person}{Ivan Skorokhodov}, \bibinfo{person}{Anton
  Rykachevskiy}, \bibinfo{person}{Dmitry Emelyanenko}, \bibinfo{person}{Sergey
  Slotin}, {and} \bibinfo{person}{Anton Ponkratov}.}
  \bibinfo{year}{2018}\natexlab{}.
\newblock \showarticletitle{Semi-Supervised Neural Machine Translation with
  Language Models}. In \bibinfo{booktitle}{\emph{Proceedings of the {AMTA} 2018
  Workshop on Technologies for {MT} of Low Resource Languages ({L}o{R}es{MT}
  2018)}}. \bibinfo{pages}{37--44}.
\newblock


\bibitem[\protect\citeauthoryear{S{\o}gaard, Ruder, and Vuli{\'c}}{S{\o}gaard
  et~al\mbox{.}}{2018}]%
        {sogaard-etal-2018-limitations}
\bibfield{author}{\bibinfo{person}{Anders S{\o}gaard},
  \bibinfo{person}{Sebastian Ruder}, {and} \bibinfo{person}{Ivan Vuli{\'c}}.}
  \bibinfo{year}{2018}\natexlab{}.
\newblock \showarticletitle{On the Limitations of Unsupervised Bilingual
  Dictionary Induction}. In \bibinfo{booktitle}{\emph{Proceedings of the 56th
  Annual Meeting of the Association for Computational Linguistics}}.
  \bibinfo{pages}{778--788}.
\newblock


\bibitem[\protect\citeauthoryear{Song, Tan, Qin, Lu, and Liu}{Song
  et~al\mbox{.}}{2019}]%
        {song2019mass}
\bibfield{author}{\bibinfo{person}{Kaitao Song}, \bibinfo{person}{Xu Tan},
  \bibinfo{person}{Tao Qin}, \bibinfo{person}{Jianfeng Lu}, {and}
  \bibinfo{person}{Tie{-}Yan Liu}.} \bibinfo{year}{2019}\natexlab{}.
\newblock \showarticletitle{{MASS:} Masked Sequence to Sequence Pre-training
  for Language Generation}. In \bibinfo{booktitle}{\emph{Proceedings of the
  36th International Conference on Machine Learning, {ICML} 2019, 9-15 June
  2019}} \emph{(\bibinfo{series}{Proceedings of Machine Learning Research})},
  Vol.~\bibinfo{volume}{97}. \bibinfo{pages}{5926--5936}.
\newblock


\bibitem[\protect\citeauthoryear{Stahlberg}{Stahlberg}{2020}]%
        {stahlberg2020neural}
\bibfield{author}{\bibinfo{person}{Felix Stahlberg}.}
  \bibinfo{year}{2020}\natexlab{}.
\newblock \showarticletitle{Neural machine translation: A review}.
\newblock \bibinfo{journal}{\emph{Journal of AI Research}}
  \bibinfo{volume}{69} (\bibinfo{year}{2020}), \bibinfo{pages}{343--418}.
\newblock


\bibitem[\protect\citeauthoryear{Stahlberg, Cross, and Stoyanov}{Stahlberg
  et~al\mbox{.}}{2018}]%
        {stahlberg2018simple}
\bibfield{author}{\bibinfo{person}{Felix Stahlberg}, \bibinfo{person}{James
  Cross}, {and} \bibinfo{person}{Veselin Stoyanov}.}
  \bibinfo{year}{2018}\natexlab{}.
\newblock \showarticletitle{Simple Fusion: Return of the Language Model}. In
  \bibinfo{booktitle}{\emph{Proceedings of the Third Conference on Machine
  Translation: Research Papers}}. \bibinfo{pages}{204--211}.
\newblock


\bibitem[\protect\citeauthoryear{Stanovsky, Smith, and Zettlemoyer}{Stanovsky
  et~al\mbox{.}}{2019}]%
        {stanovsky2019evaluating}
\bibfield{author}{\bibinfo{person}{Gabriel Stanovsky}, \bibinfo{person}{Noah~A.
  Smith}, {and} \bibinfo{person}{Luke Zettlemoyer}.}
  \bibinfo{year}{2019}\natexlab{}.
\newblock \showarticletitle{Evaluating Gender Bias in Machine Translation}. In
  \bibinfo{booktitle}{\emph{Proceedings of the 57th Annual Meeting of the
  Association for Computational Linguistics}}. \bibinfo{pages}{1679--1684}.
\newblock


\bibitem[\protect\citeauthoryear{Sun, Wang, Chen, Utiyama, Sumita, and
  Zhao}{Sun et~al\mbox{.}}{2019}]%
        {sun-etal-2019-unsupervised}
\bibfield{author}{\bibinfo{person}{Haipeng Sun}, \bibinfo{person}{Rui Wang},
  \bibinfo{person}{Kehai Chen}, \bibinfo{person}{Masao Utiyama},
  \bibinfo{person}{Eiichiro Sumita}, {and} \bibinfo{person}{Tiejun Zhao}.}
  \bibinfo{year}{2019}\natexlab{}.
\newblock \showarticletitle{Unsupervised Bilingual Word Embedding Agreement for
  Unsupervised Neural Machine Translation}. In
  \bibinfo{booktitle}{\emph{Proceedings of the 57th Annual Meeting of the
  Association for Computational Linguistics}}. \bibinfo{pages}{1235--1245}.
\newblock


\bibitem[\protect\citeauthoryear{Sun, Wang, Chen, Utiyama, Sumita, and
  Zhao}{Sun et~al\mbox{.}}{2020}]%
        {sun-etal-2020-knowledge}
\bibfield{author}{\bibinfo{person}{Haipeng Sun}, \bibinfo{person}{Rui Wang},
  \bibinfo{person}{Kehai Chen}, \bibinfo{person}{Masao Utiyama},
  \bibinfo{person}{Eiichiro Sumita}, {and} \bibinfo{person}{Tiejun Zhao}.}
  \bibinfo{year}{2020}\natexlab{}.
\newblock \showarticletitle{Knowledge Distillation for Multilingual
  Unsupervised Neural Machine Translation}. In
  \bibinfo{booktitle}{\emph{Proceedings of the 58th Annual Meeting of the
  Association for Computational Linguistics}}. \bibinfo{pages}{3525--3535}.
\newblock


\bibitem[\protect\citeauthoryear{Sun, Zhu, Yifan, and Mi}{Sun
  et~al\mbox{.}}{2021}]%
        {sun2021parallel}
\bibfield{author}{\bibinfo{person}{Yu Sun}, \bibinfo{person}{Shaolin Zhu},
  \bibinfo{person}{Feng Yifan}, {and} \bibinfo{person}{Chenggang Mi}.}
  \bibinfo{year}{2021}\natexlab{}.
\newblock \showarticletitle{Parallel sentences mining with transfer learning in
  an unsupervised setting}. In \bibinfo{booktitle}{\emph{Proceedings of the
  2021 Conference of the North American Chapter of the Association for
  Computational Linguistics: Student Research Workshop}}.
  \bibinfo{pages}{136--142}.
\newblock


\bibitem[\protect\citeauthoryear{Tan, Ren, He, Qin, Zhao, and Liu}{Tan
  et~al\mbox{.}}{2019}]%
        {tan2018multilingual}
\bibfield{author}{\bibinfo{person}{Xu Tan}, \bibinfo{person}{Yi Ren},
  \bibinfo{person}{Di He}, \bibinfo{person}{Tao Qin}, \bibinfo{person}{Zhou
  Zhao}, {and} \bibinfo{person}{Tie{-}Yan Liu}.}
  \bibinfo{year}{2019}\natexlab{}.
\newblock \showarticletitle{Multilingual Neural Machine Translation with
  Knowledge Distillation}. In \bibinfo{booktitle}{\emph{7th International
  Conference on Learning Representations, {ICLR} 2019}}.
\newblock


\bibitem[\protect\citeauthoryear{Tennage, Sandaruwan, Thilakarathne, Herath,
  Ranathunga, Jayasena, and Dias}{Tennage et~al\mbox{.}}{2017}]%
        {tennage2017neural}
\bibfield{author}{\bibinfo{person}{Pasindu Tennage}, \bibinfo{person}{Prabath
  Sandaruwan}, \bibinfo{person}{Malith Thilakarathne}, \bibinfo{person}{Achini
  Herath}, \bibinfo{person}{Surangika Ranathunga}, \bibinfo{person}{Sanath
  Jayasena}, {and} \bibinfo{person}{Gihan Dias}.}
  \bibinfo{year}{2017}\natexlab{}.
\newblock \showarticletitle{Neural machine translation for sinhala and tamil
  languages}. In \bibinfo{booktitle}{\emph{2017 International Conference on
  Asian Language Processing (IALP)}}. IEEE, \bibinfo{pages}{189--192}.
\newblock


\bibitem[\protect\citeauthoryear{Vaswani, Bengio, Brevdo, Chollet, Gomez,
  Gouws, Jones, Kaiser, Kalchbrenner, Parmar, Sepassi, Shazeer, and
  Uszkoreit}{Vaswani et~al\mbox{.}}{2018}]%
        {vaswani2018tensor2tensor}
\bibfield{author}{\bibinfo{person}{Ashish Vaswani}, \bibinfo{person}{Samy
  Bengio}, \bibinfo{person}{Eugene Brevdo}, \bibinfo{person}{Francois Chollet},
  \bibinfo{person}{Aidan Gomez}, \bibinfo{person}{Stephan Gouws},
  \bibinfo{person}{Llion Jones}, \bibinfo{person}{{\L}ukasz Kaiser},
  \bibinfo{person}{Nal Kalchbrenner}, \bibinfo{person}{Niki Parmar},
  \bibinfo{person}{Ryan Sepassi}, \bibinfo{person}{Noam Shazeer}, {and}
  \bibinfo{person}{Jakob Uszkoreit}.} \bibinfo{year}{2018}\natexlab{}.
\newblock \showarticletitle{{T}ensor2{T}ensor for Neural Machine Translation}.
  In \bibinfo{booktitle}{\emph{Proceedings of the 13th Conference of the
  Association for Machine Translation in the {A}mericas (Volume 1: Research
  Track)}}. \bibinfo{pages}{193--199}.
\newblock


\bibitem[\protect\citeauthoryear{Vaswani, Shazeer, Parmar, Uszkoreit, Jones,
  Gomez, Kaiser, and Polosukhin}{Vaswani et~al\mbox{.}}{2017}]%
        {vaswani2017attention}
\bibfield{author}{\bibinfo{person}{Ashish Vaswani}, \bibinfo{person}{Noam
  Shazeer}, \bibinfo{person}{Niki Parmar}, \bibinfo{person}{Jakob Uszkoreit},
  \bibinfo{person}{Llion Jones}, \bibinfo{person}{Aidan~N. Gomez},
  \bibinfo{person}{Lukasz Kaiser}, {and} \bibinfo{person}{Illia Polosukhin}.}
  \bibinfo{year}{2017}\natexlab{}.
\newblock \showarticletitle{Attention is All you Need}. In
  \bibinfo{booktitle}{\emph{Advances in Neural Information Processing Systems
  30: Annual Conference on Neural Information Processing Systems 2017, December
  4-9, 2017}}. \bibinfo{pages}{5998--6008}.
\newblock


\bibitem[\protect\citeauthoryear{V{\'a}zquez, Raganato, Creutz, and
  Tiedemann}{V{\'a}zquez et~al\mbox{.}}{2020}]%
        {vazquez2020systematic}
\bibfield{author}{\bibinfo{person}{Ra{\'u}l V{\'a}zquez},
  \bibinfo{person}{Alessandro Raganato}, \bibinfo{person}{Mathias Creutz},
  {and} \bibinfo{person}{J{\"o}rg Tiedemann}.} \bibinfo{year}{2020}\natexlab{}.
\newblock \showarticletitle{A Systematic Study of Inner-Attention-Based
  Sentence Representations in Multilingual Neural Machine Translation}.
\newblock \bibinfo{journal}{\emph{Computational Linguistics}}
  \bibinfo{volume}{46}, \bibinfo{number}{2} (\bibinfo{year}{2020}),
  \bibinfo{pages}{387--424}.
\newblock


\bibitem[\protect\citeauthoryear{V{\'a}zquez, Raganato, Tiedemann, and
  Creutz}{V{\'a}zquez et~al\mbox{.}}{2019}]%
        {vazquez2018multilingual}
\bibfield{author}{\bibinfo{person}{Ra{\'u}l V{\'a}zquez},
  \bibinfo{person}{Alessandro Raganato}, \bibinfo{person}{J{\"o}rg Tiedemann},
  {and} \bibinfo{person}{Mathias Creutz}.} \bibinfo{year}{2019}\natexlab{}.
\newblock \showarticletitle{Multilingual {NMT} with a Language-Independent
  Attention Bridge}. In \bibinfo{booktitle}{\emph{Proceedings of the 4th
  Workshop on Representation Learning for NLP}}. \bibinfo{pages}{33--39}.
\newblock


\bibitem[\protect\citeauthoryear{Wang, Bai, Li, and Zhao}{Wang
  et~al\mbox{.}}{2021}]%
        {wang2020crosslingual}
\bibfield{author}{\bibinfo{person}{Mingxuan Wang}, \bibinfo{person}{Hongxiao
  Bai}, \bibinfo{person}{Lei Li}, {and} \bibinfo{person}{Hai Zhao}.}
  \bibinfo{year}{2021}\natexlab{}.
\newblock \showarticletitle{Cross-lingual Supervision Improves Unsupervised
  Neural Machine Translation}. In \bibinfo{booktitle}{\emph{Proceedings of the
  2021 Conference of the North American Chapter of the Association for
  Computational Linguistics: Human Language Technologies: Industry Papers}}.
  \bibinfo{pages}{89--96}.
\newblock


\bibitem[\protect\citeauthoryear{Wang, Liu, Wang, Luan, and Sun}{Wang
  et~al\mbox{.}}{2019a}]%
        {wang2019improving}
\bibfield{author}{\bibinfo{person}{Shuo Wang}, \bibinfo{person}{Yang Liu},
  \bibinfo{person}{Chao Wang}, \bibinfo{person}{Huanbo Luan}, {and}
  \bibinfo{person}{Maosong Sun}.} \bibinfo{year}{2019}\natexlab{a}.
\newblock \showarticletitle{Improving Back-Translation with Uncertainty-based
  Confidence Estimation}. In \bibinfo{booktitle}{\emph{Proceedings of the 2019
  Conference on Empirical Methods in Natural Language Processing and the 9th
  International Joint Conference on Natural Language Processing}}.
  \bibinfo{pages}{791--802}.
\newblock


\bibitem[\protect\citeauthoryear{Wang and Neubig}{Wang and Neubig}{2019}]%
        {wang-neubig-2019-target}
\bibfield{author}{\bibinfo{person}{Xinyi Wang} {and} \bibinfo{person}{Graham
  Neubig}.} \bibinfo{year}{2019}\natexlab{}.
\newblock \showarticletitle{Target Conditioned Sampling: Optimizing Data
  Selection for Multilingual Neural Machine Translation}. In
  \bibinfo{booktitle}{\emph{Proceedings of the 57th Annual Meeting of the
  ACL}}. \bibinfo{pages}{5823--5828}.
\newblock


\bibitem[\protect\citeauthoryear{Wang, Pham, Arthur, and Neubig}{Wang
  et~al\mbox{.}}{2019b}]%
        {wang2018multilingual}
\bibfield{author}{\bibinfo{person}{Xinyi Wang}, \bibinfo{person}{Hieu Pham},
  \bibinfo{person}{Philip Arthur}, {and} \bibinfo{person}{Graham Neubig}.}
  \bibinfo{year}{2019}\natexlab{b}.
\newblock \showarticletitle{Multilingual Neural Machine Translation With Soft
  Decoupled Encoding}. In \bibinfo{booktitle}{\emph{7th International
  Conference on Learning Representations, {ICLR} 2019,}}.
\newblock


\bibitem[\protect\citeauthoryear{Wang, Pham, Dai, and Neubig}{Wang
  et~al\mbox{.}}{2018a}]%
        {wang-etal-2018-switchout}
\bibfield{author}{\bibinfo{person}{Xinyi Wang}, \bibinfo{person}{Hieu Pham},
  \bibinfo{person}{Zihang Dai}, {and} \bibinfo{person}{Graham Neubig}.}
  \bibinfo{year}{2018}\natexlab{a}.
\newblock \showarticletitle{{S}witch{O}ut: an Efficient Data Augmentation
  Algorithm for Neural Machine Translation}. In
  \bibinfo{booktitle}{\emph{Proceedings of the 2018 Conference on Empirical
  Methods in NLP}}. \bibinfo{pages}{856--861}.
\newblock


\bibitem[\protect\citeauthoryear{Wang, Tsvetkov, and Neubig}{Wang
  et~al\mbox{.}}{2020a}]%
        {wang2020balancing}
\bibfield{author}{\bibinfo{person}{Xinyi Wang}, \bibinfo{person}{Yulia
  Tsvetkov}, {and} \bibinfo{person}{Graham Neubig}.}
  \bibinfo{year}{2020}\natexlab{a}.
\newblock \showarticletitle{Balancing Training for Multilingual Neural Machine
  Translation}. In \bibinfo{booktitle}{\emph{Proceedings of the 58th Annual
  Meeting of the Association for Computational Linguistics}}.
  \bibinfo{pages}{8526--8537}.
\newblock


\bibitem[\protect\citeauthoryear{Wang, Xia, Zhao, Bian, Qin, Liu, and Liu}{Wang
  et~al\mbox{.}}{2018b}]%
        {wang2018dual}
\bibfield{author}{\bibinfo{person}{Yijun Wang}, \bibinfo{person}{Yingce Xia},
  \bibinfo{person}{Li Zhao}, \bibinfo{person}{Jiang Bian}, \bibinfo{person}{Tao
  Qin}, \bibinfo{person}{Guiquan Liu}, {and} \bibinfo{person}{Tie-Yan Liu}.}
  \bibinfo{year}{2018}\natexlab{b}.
\newblock \showarticletitle{Dual transfer learning for neural machine
  translation with marginal distribution regularization}. In
  \bibinfo{booktitle}{\emph{Proceedings of the AAAI Conference on Artificial
  Intelligence}}, Vol.~\bibinfo{volume}{32}.
\newblock


\bibitem[\protect\citeauthoryear{Wang, Zhai, and Hassan}{Wang
  et~al\mbox{.}}{2020b}]%
        {wang2020multi}
\bibfield{author}{\bibinfo{person}{Yiren Wang}, \bibinfo{person}{ChengXiang
  Zhai}, {and} \bibinfo{person}{Hany Hassan}.}
  \bibinfo{year}{2020}\natexlab{b}.
\newblock \showarticletitle{Multi-task Learning for Multilingual Neural Machine
  Translation}. In \bibinfo{booktitle}{\emph{Proceedings of the 2020 Conference
  on Empirical Methods in Natural Language Processing}}.
  \bibinfo{pages}{1022--1034}.
\newblock


\bibitem[\protect\citeauthoryear{Wang, Zhang, Zhai, Xu, and Zong}{Wang
  et~al\mbox{.}}{2018c}]%
        {wang-etal-2018-three}
\bibfield{author}{\bibinfo{person}{Yining Wang}, \bibinfo{person}{Jiajun
  Zhang}, \bibinfo{person}{Feifei Zhai}, \bibinfo{person}{Jingfang Xu}, {and}
  \bibinfo{person}{Chengqing Zong}.} \bibinfo{year}{2018}\natexlab{c}.
\newblock \showarticletitle{Three Strategies to Improve One-to-Many
  Multilingual Translation}. In \bibinfo{booktitle}{\emph{Proceedings of the
  2018 Conference on Empirical Methods in NLP}}. \bibinfo{pages}{2955--2960}.
\newblock


\bibitem[\protect\citeauthoryear{Wang, Zhang, Zhou, Liu, and Zong}{Wang
  et~al\mbox{.}}{2019c}]%
        {wang2019synchronously}
\bibfield{author}{\bibinfo{person}{Yining Wang}, \bibinfo{person}{Jiajun
  Zhang}, \bibinfo{person}{Long Zhou}, \bibinfo{person}{Yuchen Liu}, {and}
  \bibinfo{person}{Chengqing Zong}.} \bibinfo{year}{2019}\natexlab{c}.
\newblock \showarticletitle{Synchronously Generating Two Languages with
  Interactive Decoding}. In \bibinfo{booktitle}{\emph{Proceedings of the 2019
  Conference on Empirical Methods in Natural Language Processing and the 9th
  International Joint Conference on Natural Language Processing
  (EMNLP-IJCNLP)}}. \bibinfo{pages}{3350--3355}.
\newblock


\bibitem[\protect\citeauthoryear{Wang, Zhou, Zhang, Zhai, Xu, and Zong}{Wang
  et~al\mbox{.}}{2019d}]%
        {wang2019compact}
\bibfield{author}{\bibinfo{person}{Yining Wang}, \bibinfo{person}{Long Zhou},
  \bibinfo{person}{Jiajun Zhang}, \bibinfo{person}{Feifei Zhai},
  \bibinfo{person}{Jingfang Xu}, {and} \bibinfo{person}{Chengqing Zong}.}
  \bibinfo{year}{2019}\natexlab{d}.
\newblock \showarticletitle{A Compact and Language-Sensitive Multilingual
  Translation Method}. In \bibinfo{booktitle}{\emph{Proceedings of the 57th
  Annual Meeting of the Association for Computational Linguistics}}.
  \bibinfo{pages}{1213--1223}.
\newblock


\bibitem[\protect\citeauthoryear{Wu, Wang, and Wang}{Wu et~al\mbox{.}}{2019}]%
        {wu-etal-2019-extract}
\bibfield{author}{\bibinfo{person}{Jiawei Wu}, \bibinfo{person}{Xin Wang},
  {and} \bibinfo{person}{William~Yang Wang}.} \bibinfo{year}{2019}\natexlab{}.
\newblock \showarticletitle{Extract and Edit: An Alternative to
  Back-Translation for Unsupervised Neural Machine Translation}. In
  \bibinfo{booktitle}{\emph{Proceedings of the 2019 Conference of the North
  {A}merican Chapter of the Association for Computational Linguistics: Human
  Language Technologies}}. \bibinfo{pages}{1173--1183}.
\newblock


\bibitem[\protect\citeauthoryear{Wu, Tian, Qin, Lai, and Liu}{Wu
  et~al\mbox{.}}{2018}]%
        {wu2018study}
\bibfield{author}{\bibinfo{person}{Lijun Wu}, \bibinfo{person}{Fei Tian},
  \bibinfo{person}{Tao Qin}, \bibinfo{person}{Jianhuang Lai}, {and}
  \bibinfo{person}{Tie-Yan Liu}.} \bibinfo{year}{2018}\natexlab{}.
\newblock \showarticletitle{A Study of Reinforcement Learning for Neural
  Machine Translation}. In \bibinfo{booktitle}{\emph{Proceedings of the 2018
  Conference on Empirical Methods in NLP}}. \bibinfo{pages}{3612--3621}.
\newblock


\bibitem[\protect\citeauthoryear{Xia, Kong, Anastasopoulos, and Neubig}{Xia
  et~al\mbox{.}}{2019}]%
        {xia-etal-2019-generalized}
\bibfield{author}{\bibinfo{person}{Mengzhou Xia}, \bibinfo{person}{Xiang Kong},
  \bibinfo{person}{Antonios Anastasopoulos}, {and} \bibinfo{person}{Graham
  Neubig}.} \bibinfo{year}{2019}\natexlab{}.
\newblock \showarticletitle{Generalized Data Augmentation for Low-Resource
  Translation}. In \bibinfo{booktitle}{\emph{Proceedings of the 57th Annual
  Meeting of the ACL}}. \bibinfo{pages}{5786--5796}.
\newblock


\bibitem[\protect\citeauthoryear{Xu, Li, Xu, Li, Li, Xiao, and Zhu}{Xu
  et~al\mbox{.}}{2019}]%
        {xu2019analysis}
\bibfield{author}{\bibinfo{person}{Nuo Xu}, \bibinfo{person}{Yinqiao Li},
  \bibinfo{person}{Chen Xu}, \bibinfo{person}{Yanyang Li}, \bibinfo{person}{Bei
  Li}, \bibinfo{person}{Tong Xiao}, {and} \bibinfo{person}{Jingbo Zhu}.}
  \bibinfo{year}{2019}\natexlab{}.
\newblock \showarticletitle{Analysis of Back-translation Methods for
  Low-Resource Neural Machine Translation}. In \bibinfo{booktitle}{\emph{CCF
  International Conference on Natural Language Processing and Chinese
  Computing}}. Springer, \bibinfo{pages}{466--475}.
\newblock


\bibitem[\protect\citeauthoryear{Xu, Niu, and Carpuat}{Xu
  et~al\mbox{.}}{2020}]%
        {xu2020dual}
\bibfield{author}{\bibinfo{person}{Weijia Xu}, \bibinfo{person}{Xing Niu},
  {and} \bibinfo{person}{Marine Carpuat}.} \bibinfo{year}{2020}\natexlab{}.
\newblock \showarticletitle{Dual Reconstruction: a Unifying Objective for
  Semi-Supervised Neural Machine Translation}. In
  \bibinfo{booktitle}{\emph{Findings of the Association for Computational
  Linguistics: EMNLP 2020}}. \bibinfo{pages}{2006--2020}.
\newblock


\bibitem[\protect\citeauthoryear{Yang, Wang, and Chu}{Yang
  et~al\mbox{.}}{2020}]%
        {yang2020survey}
\bibfield{author}{\bibinfo{person}{Shuoheng Yang}, \bibinfo{person}{Yuxin
  Wang}, {and} \bibinfo{person}{Xiaowen Chu}.} \bibinfo{year}{2020}\natexlab{}.
\newblock \showarticletitle{A Survey of Deep Learning Techniques for Neural
  Machine Translation}.
\newblock \bibinfo{journal}{\emph{arXiv preprint arXiv:2002.07526}}
  (\bibinfo{year}{2020}).
\newblock


\bibitem[\protect\citeauthoryear{Yang, {\'{A}}brego, Yuan, Guo, Shen, Cer,
  Sung, Strope, and Kurzweil}{Yang et~al\mbox{.}}{2019}]%
        {yang2019improving}
\bibfield{author}{\bibinfo{person}{Yinfei Yang},
  \bibinfo{person}{Gustavo~Hern{\'{a}}ndez {\'{A}}brego},
  \bibinfo{person}{Steve Yuan}, \bibinfo{person}{Mandy Guo},
  \bibinfo{person}{Qinlan Shen}, \bibinfo{person}{Daniel Cer},
  \bibinfo{person}{Yun{-}Hsuan Sung}, \bibinfo{person}{Brian Strope}, {and}
  \bibinfo{person}{Ray Kurzweil}.} \bibinfo{year}{2019}\natexlab{}.
\newblock \showarticletitle{Improving Multilingual Sentence Embedding using
  Bi-directional Dual Encoder with Additive Margin Softmax}. In
  \bibinfo{booktitle}{\emph{Proceedings of the Twenty-Eighth International
  Joint Conference on Artificial Intelligence, {IJCAI} 2019, Macao, China,
  August 10-16, 2019}}, \bibfield{editor}{\bibinfo{person}{Sarit Kraus}} (Ed.).
  \bibinfo{pages}{5370--5378}.
\newblock


\bibitem[\protect\citeauthoryear{Yang, Chen, Wang, and Xu}{Yang
  et~al\mbox{.}}{2018a}]%
        {yang2017improving}
\bibfield{author}{\bibinfo{person}{Zhen Yang}, \bibinfo{person}{Wei Chen},
  \bibinfo{person}{Feng Wang}, {and} \bibinfo{person}{Bo Xu}.}
  \bibinfo{year}{2018}\natexlab{a}.
\newblock \showarticletitle{Improving Neural Machine Translation with
  Conditional Sequence Generative Adversarial Nets}. In
  \bibinfo{booktitle}{\emph{Proceedings of the 2018 Conference of the North
  {A}merican Chapter of the Association for Computational Linguistics: Human
  Language Technologies, Volume 1 (Long Papers)}}. \bibinfo{pages}{1346--1355}.
\newblock


\bibitem[\protect\citeauthoryear{Yang, Chen, Wang, and Xu}{Yang
  et~al\mbox{.}}{2018b}]%
        {yang2018unsupervised}
\bibfield{author}{\bibinfo{person}{Zhen Yang}, \bibinfo{person}{Wei Chen},
  \bibinfo{person}{Feng Wang}, {and} \bibinfo{person}{Bo Xu}.}
  \bibinfo{year}{2018}\natexlab{b}.
\newblock \showarticletitle{Unsupervised Domain Adaptation for Neural Machine
  Translation}. In \bibinfo{booktitle}{\emph{2018 24th International Conference
  on Pattern Recognition (ICPR)}}. IEEE, \bibinfo{pages}{338--343}.
\newblock


\bibitem[\protect\citeauthoryear{Yang, Chen, Wang, and Xu}{Yang
  et~al\mbox{.}}{2018c}]%
        {yang-etal-2018-unsupervised}
\bibfield{author}{\bibinfo{person}{Zhen Yang}, \bibinfo{person}{Wei Chen},
  \bibinfo{person}{Feng Wang}, {and} \bibinfo{person}{Bo Xu}.}
  \bibinfo{year}{2018}\natexlab{c}.
\newblock \showarticletitle{Unsupervised Neural Machine Translation with Weight
  Sharing}. In \bibinfo{booktitle}{\emph{Proceedings of the 56th Annual Meeting
  of the Association for Computational Linguistics}}. \bibinfo{pages}{46--55}.
\newblock


\bibitem[\protect\citeauthoryear{Zaremoodi, Buntine, and Haffari}{Zaremoodi
  et~al\mbox{.}}{2018}]%
        {zaremoodi-etal-2018-adaptive}
\bibfield{author}{\bibinfo{person}{Poorya Zaremoodi}, \bibinfo{person}{Wray
  Buntine}, {and} \bibinfo{person}{Gholamreza Haffari}.}
  \bibinfo{year}{2018}\natexlab{}.
\newblock \showarticletitle{Adaptive Knowledge Sharing in Multi-Task Learning:
  Improving Low-Resource Neural Machine Translation}. In
  \bibinfo{booktitle}{\emph{Proceedings of the 56th Annual Meeting of the
  Association for Computational Linguistics (Volume 2: Short Papers)}}.
  \bibinfo{pages}{656--661}.
\newblock


\bibitem[\protect\citeauthoryear{Zhang, Nagesh, and Knight}{Zhang
  et~al\mbox{.}}{2020a}]%
        {zhang2020parallel}
\bibfield{author}{\bibinfo{person}{Boliang Zhang}, \bibinfo{person}{Ajay
  Nagesh}, {and} \bibinfo{person}{Kevin Knight}.}
  \bibinfo{year}{2020}\natexlab{a}.
\newblock \showarticletitle{Parallel Corpus Filtering via Pre-trained Language
  Models}. In \bibinfo{booktitle}{\emph{Proceedings of the 58th Annual Meeting
  of the Association for Computational Linguistics}}.
  \bibinfo{pages}{8545--8554}.
\newblock


\bibitem[\protect\citeauthoryear{Zhang, Williams, Titov, and Sennrich}{Zhang
  et~al\mbox{.}}{2020b}]%
        {zhang2020improving}
\bibfield{author}{\bibinfo{person}{Biao Zhang}, \bibinfo{person}{Philip
  Williams}, \bibinfo{person}{Ivan Titov}, {and} \bibinfo{person}{Rico
  Sennrich}.} \bibinfo{year}{2020}\natexlab{b}.
\newblock \showarticletitle{Improving Massively Multilingual Neural Machine
  Translation and Zero-Shot Translation}. In
  \bibinfo{booktitle}{\emph{Proceedings of the 58th Annual Meeting of the
  ACL}}. \bibinfo{pages}{1628--1639}.
\newblock


\bibitem[\protect\citeauthoryear{Zhang and Zong}{Zhang and Zong}{2016}]%
        {zhang-zong-2016-exploiting}
\bibfield{author}{\bibinfo{person}{Jiajun Zhang} {and}
  \bibinfo{person}{Chengqing Zong}.} \bibinfo{year}{2016}\natexlab{}.
\newblock \showarticletitle{Exploiting Source-side Monolingual Data in Neural
  Machine Translation}. In \bibinfo{booktitle}{\emph{Proceedings of the 2016
  Conference on Empirical Methods in Natural Language Processing}}.
  \bibinfo{pages}{1535--1545}.
\newblock


\bibitem[\protect\citeauthoryear{Zhang and Zong}{Zhang and Zong}{2020}]%
        {zhang2020neurala}
\bibfield{author}{\bibinfo{person}{Jiajun Zhang} {and}
  \bibinfo{person}{Chengqing Zong}.} \bibinfo{year}{2020}\natexlab{}.
\newblock \showarticletitle{Neural Machine Translation: Challenges, Progress
  and Future}.
\newblock \bibinfo{journal}{\emph{arXiv preprint arXiv:2004.05809}}
  (\bibinfo{year}{2020}).
\newblock


\bibitem[\protect\citeauthoryear{Zhang, Liu, Li, Zhou, and Chen}{Zhang
  et~al\mbox{.}}{2018}]%
        {zhang2018joint}
\bibfield{author}{\bibinfo{person}{Zhirui Zhang}, \bibinfo{person}{Shujie Liu},
  \bibinfo{person}{Mu Li}, \bibinfo{person}{Ming Zhou}, {and}
  \bibinfo{person}{Enhong Chen}.} \bibinfo{year}{2018}\natexlab{}.
\newblock \showarticletitle{Joint Training for Neural Machine Translation
  Models with Monolingual Data}. In \bibinfo{booktitle}{\emph{Proceedings of
  the Thirty-Second {AAAI} Conference on Artificial Intelligence, (AAAI-18),
  the 30th innovative Applications of Artificial Intelligence (IAAI-18), and
  the 8th {AAAI} Symposium on Educational Advances in Artificial Intelligence
  (EAAI-18)}}. \bibinfo{pages}{555--562}.
\newblock


\bibitem[\protect\citeauthoryear{Zhao, Hu, and Risteski}{Zhao
  et~al\mbox{.}}{2020}]%
        {zhao2020learning}
\bibfield{author}{\bibinfo{person}{Han Zhao}, \bibinfo{person}{Junjie Hu},
  {and} \bibinfo{person}{Andrej Risteski}.} \bibinfo{year}{2020}\natexlab{}.
\newblock \showarticletitle{On Learning Language-Invariant Representations for
  Universal Machine Translation}. In \bibinfo{booktitle}{\emph{Proceedings of
  the 37th International Conference on Machine Learning, {ICML} 2020, 13-18
  July 2020, Virtual Event}} \emph{(\bibinfo{series}{Proceedings of Machine
  Learning Research})}, Vol.~\bibinfo{volume}{119}.
  \bibinfo{pages}{11352--11364}.
\newblock


\bibitem[\protect\citeauthoryear{Zheng, Cheng, and Liu}{Zheng
  et~al\mbox{.}}{2017}]%
        {zheng2017maximum}
\bibfield{author}{\bibinfo{person}{Hao Zheng}, \bibinfo{person}{Yong Cheng},
  {and} \bibinfo{person}{Yang Liu}.} \bibinfo{year}{2017}\natexlab{}.
\newblock \showarticletitle{Maximum Expected Likelihood Estimation for
  Zero-resource Neural Machine Translation}. In
  \bibinfo{booktitle}{\emph{Proceedings of the 26th International Joint
  Conference on AI}}, \bibfield{editor}{\bibinfo{person}{Carles Sierra}} (Ed.).
  \bibinfo{pages}{4251--4257}.
\newblock


\bibitem[\protect\citeauthoryear{Zheng, Zhou, Huang, Li, Dai, and Chen}{Zheng
  et~al\mbox{.}}{2020a}]%
        {Zheng2020Mirror-Generative}
\bibfield{author}{\bibinfo{person}{Zaixiang Zheng}, \bibinfo{person}{Hao Zhou},
  \bibinfo{person}{Shujian Huang}, \bibinfo{person}{Lei Li},
  \bibinfo{person}{Xin{-}Yu Dai}, {and} \bibinfo{person}{Jiajun Chen}.}
  \bibinfo{year}{2020}\natexlab{a}.
\newblock \showarticletitle{Mirror-Generative Neural Machine Translation}. In
  \bibinfo{booktitle}{\emph{8th International Conference on Learning
  Representations, 2020}}.
\newblock


\bibitem[\protect\citeauthoryear{Zheng, Zhou, Huang, Li, Dai, and Chen}{Zheng
  et~al\mbox{.}}{2020b}]%
        {zheng2019mirror}
\bibfield{author}{\bibinfo{person}{Zaixiang Zheng}, \bibinfo{person}{Hao Zhou},
  \bibinfo{person}{Shujian Huang}, \bibinfo{person}{Lei Li},
  \bibinfo{person}{Xin{-}Yu Dai}, {and} \bibinfo{person}{Jiajun Chen}.}
  \bibinfo{year}{2020}\natexlab{b}.
\newblock \showarticletitle{Mirror-Generative Neural Machine Translation}. In
  \bibinfo{booktitle}{\emph{8th International Conference on Learning
  Representations}}.
\newblock


\bibitem[\protect\citeauthoryear{Zhu, Yu, Cheng, and Luo}{Zhu
  et~al\mbox{.}}{2020}]%
        {zhu2020language}
\bibfield{author}{\bibinfo{person}{Changfeng Zhu}, \bibinfo{person}{Heng Yu},
  \bibinfo{person}{Shanbo Cheng}, {and} \bibinfo{person}{Weihua Luo}.}
  \bibinfo{year}{2020}\natexlab{}.
\newblock \showarticletitle{Language-aware Interlingua for Multilingual Neural
  Machine Translation}. In \bibinfo{booktitle}{\emph{Proceedings of the 58th
  Annual Meeting of the ACL}}. \bibinfo{pages}{1650--1655}.
\newblock


\bibitem[\protect\citeauthoryear{Zhu, Xia, Wu, He, Qin, Zhou, Li, and Liu}{Zhu
  et~al\mbox{.}}{2019}]%
        {zhu2019incorporating}
\bibfield{author}{\bibinfo{person}{Jinhua Zhu}, \bibinfo{person}{Yingce Xia},
  \bibinfo{person}{Lijun Wu}, \bibinfo{person}{Di He}, \bibinfo{person}{Tao
  Qin}, \bibinfo{person}{Wengang Zhou}, \bibinfo{person}{Houqiang Li}, {and}
  \bibinfo{person}{Tieyan Liu}.} \bibinfo{year}{2019}\natexlab{}.
\newblock \showarticletitle{Incorporating BERT into Neural Machine
  Translation}. In \bibinfo{booktitle}{\emph{International Conference on
  Learning Representations}}.
\newblock


\bibitem[\protect\citeauthoryear{Zoph and Knight}{Zoph and Knight}{2016}]%
        {zoph-knight-2016-multi}
\bibfield{author}{\bibinfo{person}{Barret Zoph} {and} \bibinfo{person}{Kevin
  Knight}.} \bibinfo{year}{2016}\natexlab{}.
\newblock \showarticletitle{Multi-Source Neural Translation}. In
  \bibinfo{booktitle}{\emph{Proceedings of the 2016 Conference of the North
  {A}merican Chapter of the Association for Computational Linguistics: Human
  Language Technologies}}. \bibinfo{pages}{30--34}.
\newblock


\bibitem[\protect\citeauthoryear{Zoph, Yuret, May, and Knight}{Zoph
  et~al\mbox{.}}{2016}]%
        {zoph2016transfer}
\bibfield{author}{\bibinfo{person}{Barret Zoph}, \bibinfo{person}{Deniz Yuret},
  \bibinfo{person}{Jonathan May}, {and} \bibinfo{person}{Kevin Knight}.}
  \bibinfo{year}{2016}\natexlab{}.
\newblock \showarticletitle{Transfer Learning for Low-Resource Neural Machine
  Translation}. In \bibinfo{booktitle}{\emph{Proceedings of the 2016 Conference
  on Empirical Methods in NLP}}. \bibinfo{pages}{1568--1575}.
\newblock


\end{thebibliography}
	
\end{document}